%% file: main.tex
\documentclass[Afour,sageh,times]{sagej}

\usepackage{url}
\usepackage{booktabs,tabularx,pifont}
\usepackage{multirow}
\usepackage{graphicx}
\usepackage{rotating}
\usepackage{makecell}
\usepackage{xcolor}
\usepackage{appendix}
\usepackage[dvipsnames]{xcolor}
\usepackage{endnotes}
\usepackage{todonotes} 
\let\footnote=\endnote

\PassOptionsToPackage{colorlinks,bookmarksopen,bookmarksnumbered,citecolor=red,urlcolor=red}{hyperref}

\usepackage[utf8]{inputenc}
\DeclareUnicodeCharacter{0301}{\'{} }

\usepackage[most]{tcolorbox}
\newtcolorbox{mybox}[1][]{
  enhanced, colback=gray!5, colframe=black, boxrule=0.6pt, arc=2pt,
  left=6pt, right=6pt, top=4pt, bottom=4pt,
  fonttitle=\bfseries, coltitle=black, colbacktitle=gray!15,
  title={~}, #1
}

\usepackage[font=footnotesize]{caption}

\usepackage{problems}
\usepackage{common}
\usepackage{notation}
\usepackage{balance}

\newcolumntype{Y}{>{\raggedright\arraybackslash}X}
\newcolumntype{L}[1]{>{\raggedright\arraybackslash}p{#1}}
\newcolumntype{M}{>{$}X<{$}}

\newcommand{\xmark}{\textcolor{red!75!black}{\ding{55}}}%

\begin{document}


\title{SPARSER: Sparse Variable Projection by Exploiting Separable
       Structure in Robotic Perception}

\author{Nikolas R. Sanderson$^{1,2}$, Andrew Fishberg$^3$,
  Haoyu Han$^4$,
   Heng Yang$^4$, \\ Jonathan P. How$^3$,
  Hanumant Singh$^2$,
  Michael Everett$^2$, and
  Alan Papalia$^1$
}

\affiliation{\affilnum{1}Department of Naval Architecture and Marine Engineering, University of Michigan, Ann Arbor, MI, USA,
\affilnum{2}Institute of Experiential Robotics, Northeastern University, Boston, MA, USA,
\affilnum{3}Aerospace Controls Lab, Massachusetts Institute of Technology, Cambridge, MA, USA,\affilnum{4}Computational Robotics Group, Harvard University, Cambridge, MA, USA}

\corrauth{Nikolas R. Sanderson, Institute of Experiential Robotics at Northeastern University and Department of Naval Architecture and Marine Engineering at the University of Michigan. Alan Papalia, Assistant Professor, Department of Naval Architecture and Marine Engineering at the University of Michigan.}

\email{sanderson.n@northeastern.edu \& apapalia@umich.edu}

\input{sec/1_abstract}


\maketitle


\input{sec/2_intro}

\input{sec/3_related_work}
\input{sec/4a_preliminary}
\input{sec/4b_nonlinear}

\input{sec/5_experiments}

\input{sec/6_conclusion}

\begin{acks}
This work was supported by the Northeastern University Institute for
Experiential Robotics Postdoctoral Fellowship, Army Research Lab awards
W911NF-24-2-006 and W911NF-24-2-0017, the Office of Naval Research
grant N000142512322.
\end{acks}

\section*{Author's Note}
This is an extended version of worked presented at the 2026 International Conference of Robotics and Automation (ICRA) ~\cite{papalia2025sparsevariableprojectionrobotic}.
\theendnotes

\input{sec/9_appendix}

\bibliography{bib/references}
\bibliographystyle{SageH}

\end{document}

%% file: sec/1_abstract.tex
\begin{abstract}
Robotic perception often requires solving large nonlinear least-squares (NLS)
problems. While sparsity has been well exploited to scale solvers, a
complementary and underexploited structure is \emph{separability} -- where some
variables (e.g., visual landmarks) appear linearly in the residuals and admit a
closed-form solution once the remaining variables (e.g., poses) are fixed.
Variable projection (VarPro) methods exploit this structure by analytically
eliminating the linear variables, yielding a reduced problem with favorable
properties. However, VarPro has seen limited use in robotic perception, largely
because of gauge symmetries (e.g., cost invariance to global shifts and
rotations), which are common in perception and induce specific computational
challenges for standard VarPro approaches. We present SPARSER
(\textbf{S}parsity \textbf{P}reserving \textbf{A}nalytic \textbf{R}eduction for
\textbf{S}eparable \textbf{E}stimation in \textbf{R}obotics), a VarPro scheme
designed for problems with gauge symmetries that jointly exploits separability
and sparsity. In a one-time preprocessing step, our method constructs a
\emph{matrix-free Schur complement operator} that efficiently evaluates costs,
gradients, and Hessian-vector products of the reduced problem and readily
integrates with standard iterative NLS solvers. We characterize the applicable
problem class and identify special -- yet common -- cases admitting analytical
simplifications and further computational savings. Notably, the method extends
to robust cost functions handled via iteratively reweighted least squares
(IRLS): the symbolic preprocessing is computed once and reused across
reweightings, so each IRLS iteration requires only a numerical refactorization.
On 42 benchmark datasets spanning pose-graph optimization, range-aided SLAM,
sensor network localization, and structure from motion, SPARSER achieves the
fastest runtime on 41 of them on CPU and 35 on GPU, and is the only method to
solve all tested. On 24 real-world multi-robot SLAM datasets with naturally
outlier-corrupted loop closures, our robust variant is the only method to solve
all tested and the fastest on each, running $1.1\times$ to $3.1\times$ faster than
GTSAM's graduated non-convexity (GNC) solver on every dataset tested. We release open-source C++ code and all datasets.
\end{abstract}

%% file: sec/2_intro.tex
\section{Introduction}
\label{sec:introduction}

Robotic perception often requires solving large-scale optimization problems, typically
posed as nonlinear least-squares (NLS) problems with up to millions of variables
\citep{agarwal2010Bundle,ebadi2023present,kunze2018artificial,tranzatto2022cerberus}.
Such NLS formulations underpin key tasks such as simultaneous localization and
mapping (SLAM) \citep{cadena2017past}, structure from motion (SfM)
\citep{schonberger2016structure}, and sensor network localization (SNL)
\citep{mao2007wireless}.
In such tasks, a solver's performance directly impacts the
reliability and scale of robotic deployments
\citep{ebadi2023present,kunze2018artificial,tranzatto2022cerberus}.

While important, it is difficult to efficiently solve these problems because:
(i) they are large, often involving \(10^6\!-\!10^7\) variables \citep{agarwal2010Bundle};
(ii) applications typically require solving problems to high numerical accuracy
\citep{triggs2000Bundle}; and
(iii) problems often exhibit ill-conditioning that reduces the convergence rate of
standard methods \citep{golub2013matrix}.
To address these challenges, state-of-the-art systems
\citep{agarwal2022ceres,kummerle2011G2o,dellaert2012factor} exploit problem
sparsity to handle scale \citep{dellaert2017factor}, adopt second-order methods
(e.g., Levenberg--Marquardt) to ensure solution accuracy
\citep{nocedal2006numerical}, and employ numerically stable linear solvers to
mitigate ill-conditioning \citep{golub2013matrix}.
Despite these techniques, scaling limitations remain, motivating approaches to
further exploit problem structure.

\input{fig/title_fig.tex}

One path to improved efficiency is to exploit \emph{separable structure}
\citep{golub2003Separable}: once one set of variables is held fixed, the
remaining variables can be solved for in closed form. In pose-graph
optimization, for example, fixing the rotations leaves a linear least-squares
problem in the translations (see the example in \cref{sec:prelim:separable-structure}).
Variable projection (VarPro) \citep{golub2003Separable} uses this to eliminate
those variables entirely, leaving a reduced problem that is smaller and
converges faster \citep{ruhe1980algorithms}.

Yet in many perception problems, gauge symmetries in the cost (e.g., globally
rotating the solution does not change the cost) introduce rank deficiencies in
the cost matrix. This rank deficiency led prior approaches
\citep{khosoussi2016Sparse} to adopt workarounds that limited efficiency gains.

\textbf{This paper.}
We show that a broad class of robotic perception problems (e.g.,
\cite{rosen2019SESync,papalia2024Certifiably,han2025Building,halsted22arxiv})
admits a VarPro scheme that simultaneously exploits sparsity and separability.
Our approach, SPARSER, avoids challenges due to gauge symmetries and can be
applied as a one-time step before optimization begins.
We characterize this problem class by simple conditions on the cost and
constraints that can be checked a priori:
\begin{enumerate}[(i)]
      \item there must be a set of unconstrained variables -- these are the
            variables to be eliminated;
      \item the cost residuals must be linear functions; and
\item the cost-residual Jacobian block associated with the eliminated variables
      must have a specific graph-theoretic structure to admit the efficient
      closed-form construction in
      \cref{sec:prelim:efficient-schur-complement-products}.
\end{enumerate}
%
Our approach produces a matrix-free Schur complement
\citep{zhang2006schur}, i.e., an operator that can produce the reduced problem's
costs and gradients without forming the Schur complement (a large, dense
matrix).  This operator can be easily integrated into
standard iterative solvers, preserving the scalability of modern systems.
Finally, we discuss how the method extends when the ideal conditions are only
partially satisfied (e.g., when some residuals are nonlinear).

Across a range of synthetic and real-world benchmarks in SLAM, SNL, and SfM,
including a scalability sweep and 24 outlier-corrupted multi-robot datasets, our
method is the fastest on 41 of 42 standard benchmarks on CPU, 35 on GPU, and all
24 robust datasets, and is the only method to solve every dataset tested.

Our contributions are as follows:
\begin{enumerate}
      \item SPARSER a novel sparsity-preserving VarPro scheme to accelerate a broad class
            of perception problems;
      \item a precise characterization of the common structure under which the operator reaches full, sparsity-preserving efficiency;
\item an amending of SPARSER to robust cost functions via IRLS, together with a demonstration of its computational advantages;
      \item an open-source C++ implementation and all datasets used in the experiments, for reproducibility.\footnote{Code: \href{https://github.com/UMich-RobotExploration/variable-projection}{github.com/UMich-RobotExploration/variable-projection}}
\end{enumerate}

The remainder of this paper is organized as follows.
Section~\ref{sec:related-work} surveys related work on accelerating solvers for robotic perception with a focus on variable
projection and situates our contribution among
matrix-free Schur complement methods. Section~\ref{sec:problem-formulation}
formalizes the class of nonlinear least-squares problems we consider and
reduces the linear-residual case to a constrained quadratic cost problem.
Section~\ref{sec:prelim:separable-structure} reviews how separable
structure in a quadratic cost induces a closed-form elimination of the
unconstrained variables via the Schur complement, and identifies the
computational challenges that motivate our approach.
Section~\ref{sec:prelim:focus-schur-complement-products} presents our main
technical contribution: an exact, matrix-free reformulation of the
Schur complement product, graph-theoretic conditions under which the
reformulation preserves sparsity, and a probabilistic interpretation of
the resulting separable structure. Section~\ref{sec:robust-loss} extends
the framework to robust costs via iteratively reweighted least squares,
showing that the symbolic preprocessing is reusable across outer
iterations. Section~\ref{sec:cora-marg:computational-experiments}
evaluates our method on pose-graph optimization, range-aided SLAM,
sensor network localization, and structure-from-motion benchmarks against
state-of-the-art baselines. Finally, Section~\ref{sec:conclusion} summarizes our contributions and discusses
directions for future work.

\subsection{Extension of Conference Work}
This paper is an extended version of the conference paper presented at the International Conference on Robotics and Automation (ICRA) 2026~\citep{papalia2025sparsevariableprojectionrobotic}. The conference version introduced the core matrix-free Schur complement operator and the graph-theoretic sparsity conditions required for its efficient construction. The present work substantially extends that contribution in three directions:

\begin{enumerate}
\item \textbf{Extension to robust cost functions.}
We identify robust cost functions as a favorable special case of the nonlinear residuals handled by SPARSER and show how their structure can be exploited through iteratively reweighted least squares (IRLS). In particular, the symbolic preprocessing can be reused across IRLS outer iterations, requiring only a numerical refactorization after each reweighting (Section~\ref{sec:robust-loss}).

\item \textbf{Linear-algebraic and probabilistic interpretations.}
We develop complementary interpretations of the structure exploited by SPARSER, connecting its linear-algebraic formulation to conditional inference in Gaussian graphical models and clarifying the relationship between variable projection, conditional structure, and sparsity under marginalization (Section~\ref{sec:prelim:probabilistic-interpretation}).

\item \textbf{Substantially expanded empirical evaluation.}
We extend the experimental evaluation to characterize both the applicability and computational behavior of SPARSER across several regimes. This includes: (i) controlled synthetic parameter sweeps that characterize how problem structure and scale affect computational performance; (ii) the large-scale \emph{outfinite} dataset, demonstrating applicability to a problem for which the comparison methods are unable to produce a solution; (iii) real-world multi-robot datasets with naturally occurring outliers, used to evaluate the robust-cost extension and its behavior under outlier contamination; and (iv) a GPU implementation demonstrating that the computational advantages of the proposed matrix-free construction extend beyond CPU-based implementations and can be effectively exploited on hardware with parallelization capabilities (Section~\ref{sec:cora-marg:computational-experiments}).

\end{enumerate}

%% file: fig/title_fig.tex
\begin{figure}[t]
    \centering
    \setlength{\fboxsep}{2.5pt}   
    \setlength{\fboxrule}{0pt}  

    \centering
    \includegraphics[width=0.99\linewidth]{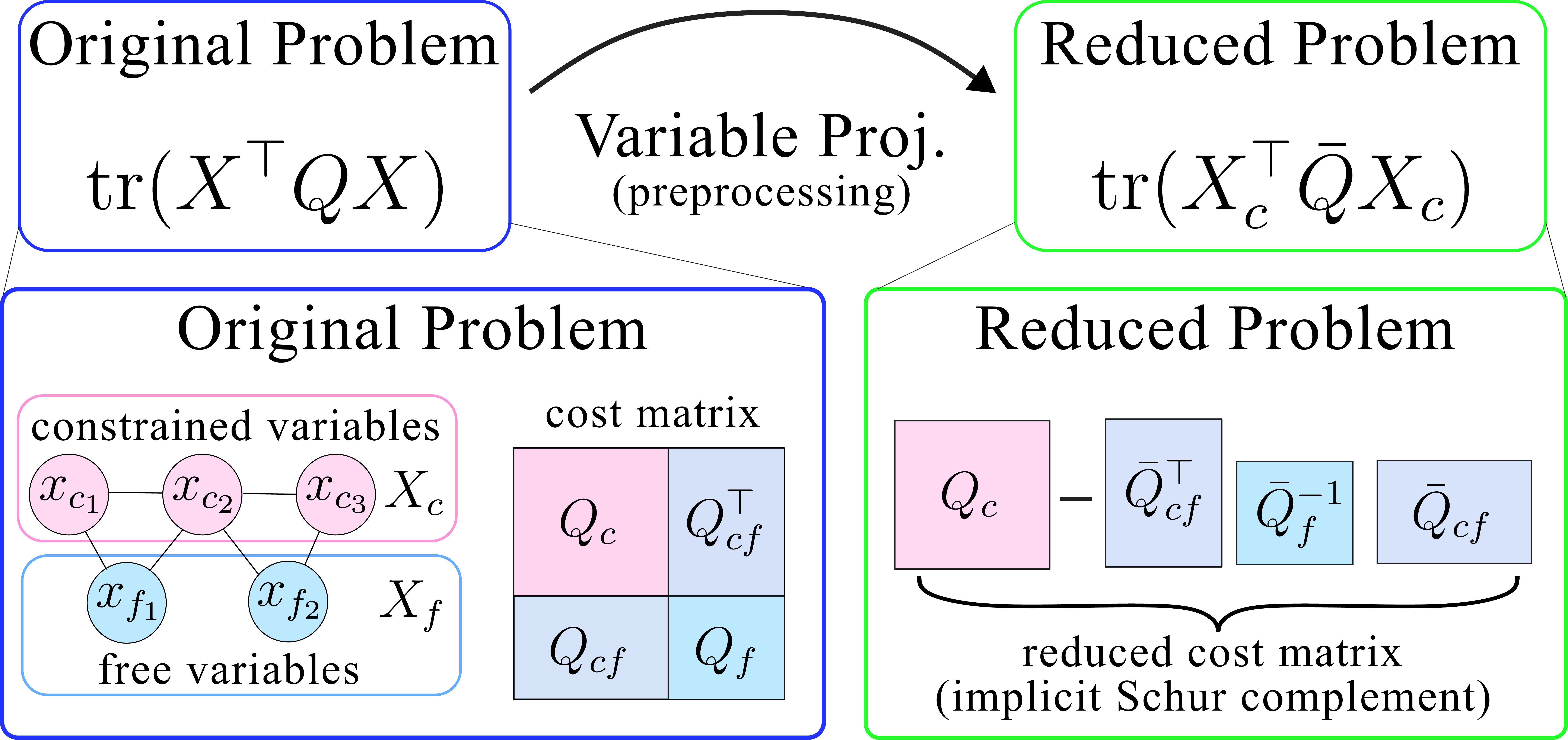}
    \fbox{}
    \vspace{-3mm}
    \hrule
    \fbox{}
    \includegraphics[width=0.99\linewidth]{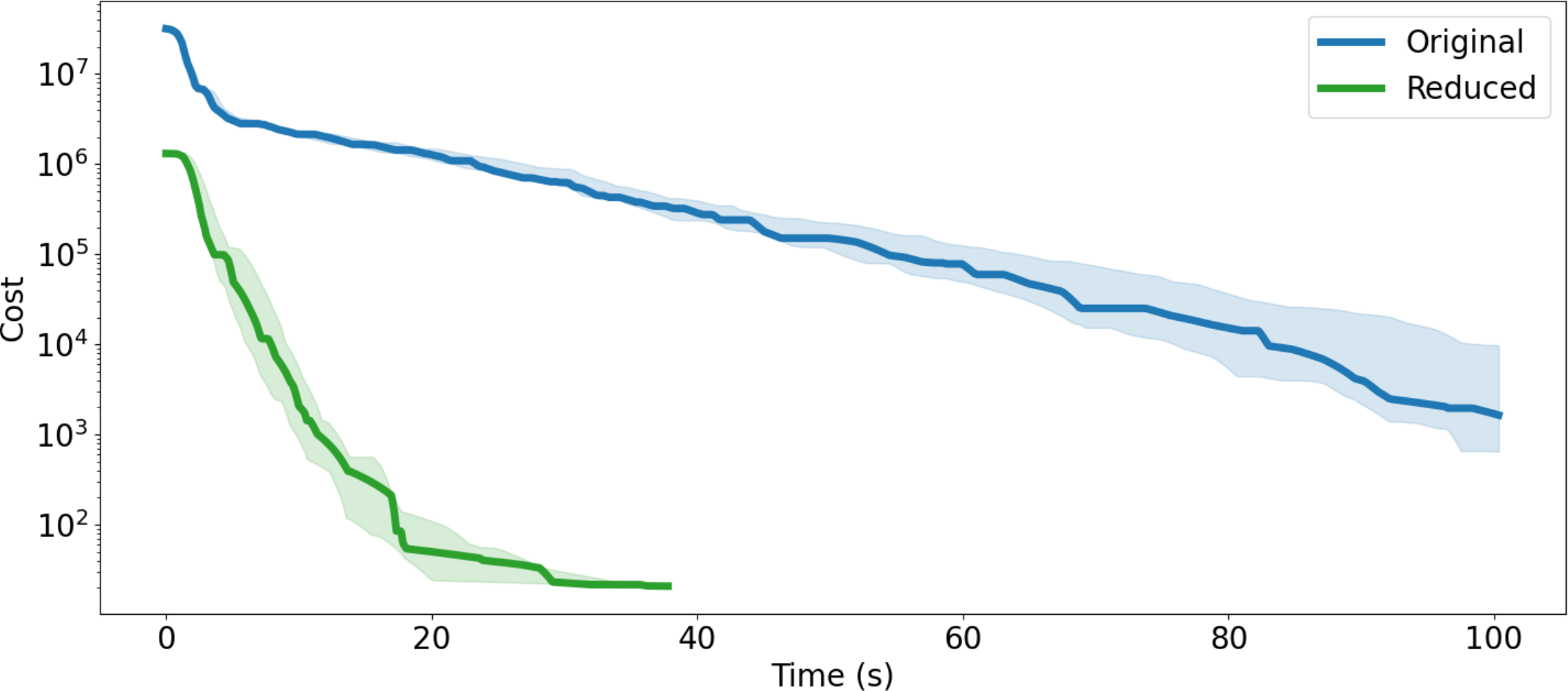}

    \caption{
        \textbf{(Top) method overview:}
        Our approach exploits \emph{separability} in optimization problem
        to perform variable projection and analytically eliminate a subset of
        variables (reducing problem size and improving conditioning),
        while preserving the efficiency of the original problem's sparsity
        structure.  Our approach can be applied as a one-time preprocessing step
        before passing the problem to a standard iterative solver.
        %
        \textbf{(Bottom) runtime improvement:}
        Comparison of cost vs.\ time for our reduced problem (green) and the
        original problem (blue) on a real-world structure from motion dataset
        (BAL-1934).  The variable projection step allows for substantial
        improvements in solver convergence and overall runtime as our method is the only one able to reach the global optimum on CPU (see~\cref{tab:pgo_wide_gtsam_compare}) and the fastest on GPU (see~\cref{tab:gpu_results}).
    }
    \vspace{-5mm}
\end{figure}

%% file: sec/3_related_work.tex

\section{Related Work}
\label{sec:related-work}
This review covers methods that accelerate robotic perception by reducing the
cost of the underlying optimization. We first survey a broad range of techniques
for improving solver efficiency, which situates our work within the wider
literature. We then turn to variable projection, the line of work that our
contribution directly targets and extends.

\subsection{Accelerating Robotic Perception Solvers}
\label{subsec:general-acceleration}

Efforts to accelerate the nonlinear least-squares problems underlying robotic
perception are largely complementary, and can be organized into three
categories.

\textbf{Structure-exploiting solvers.}
These methods exploit inherent structure in the problem for computational gain. The most widely exploited structure is sparsity. Measurement graphs arising in SLAM and bundle adjustment are typically highly sparse because each factor depends on only a small subset of variables. Consequently, widely used optimization frameworks in robotics and vision are designed around this structure, relying heavily on sparse linear algebra for computational efficiency~\citep{dellaert2012factor,kummerle2011G2o,agarwal2022ceres}.
Incremental smoothing exploits the same sparsity structure over time rather than only within a single batch solve. Since each new measurement typically involves only a small set of recently introduced variables, it modifies only a localized portion of the elimination structure, allowing the factorization to be updated incrementally rather than recomputed from scratch~\citep{kaess2008isam,kaess2012isam2}. These methods have also been extended to handle outliers robustly~\cite{Mcgann2023riSAM}. A related principle appears in the certifiable optimization literature, where chordal sparsity in semidefinite relaxations is exploited to decompose a single large positive-semidefinite constraint into a collection of smaller constraints over overlapping cliques of the measurement graph~\citep{dumbgen2025exploiting,subramanian2026exploitingchordalsparsityglobally}. Beyond exploiting sparsity in the optimization problem itself, a second line of work reduces implementation overhead by shifting computation from runtime to compile time. By representing residuals symbolically, these methods generate flattened, branch-free implementations of both the residual and its analytic Jacobian, avoiding dynamic allocations and enabling aggressive compiler optimization~\citep{Martiros-RSS-22}.

\textbf{Initialization.}
Rather than reducing the cost of each iteration, initialization methods accelerate optimization by reducing the number of iterations required to converge. Examples include decoupling the problem and initializing rotations separately~\citep{carlone2015Initialization} (chordal initialization) or solving a convex relaxation of the original problem~\citep{papalia2023score,goudar2024optimal}. More recently, Fast-Sync~\citep{holmes2026fastsync} generalized chordal initialization to synchronization over arbitrary matrix Lie groups. Good initialization can also improve solution quality in these nonconvex problems by reducing the likelihood of convergence to poor local minima.

\textbf{Parallel and distributed computation.}
Rather than reducing the total computational workload, these methods distribute it across compute units that operate concurrently. On a single machine, specialized solvers exploit multi-core CPUs or GPUs~\citep{gopinath2026graphitegpuacceleratedmixedprecisiongraph,ren2022megba}. The symbolic approach of~\cite{Martiros-RSS-22} has also been extended to this setting, where fixing the problem structure at compile time enables branch-free kernels and static memory layouts well suited to GPU execution~\citep{martens2026casparcudaacceleratorsymbolic}. When computation is distributed across multiple robots, communication constraints introduce additional challenges. Existing approaches therefore focus on restricting communication to shared variables and relaxing synchronization so that agents need not proceed in lockstep~\citep{tian2020asynchronous,mcgann2024asynchronous,fan2023daba}, or on decompositions that explicitly trade communication cost against convergence rate~\citep{fan2023majorization,sonawalla2026overlappingdomaindecompositiondistributed}. Incremental smoothing has likewise been extended to the distributed setting~\citep{mcgann_imesa_2024} and, more recently, combined with robust estimation to provide incremental, distributed, and outlier-robust inference within a single framework~\citep{mcgann2026rimesa}.

\subsection{Variable Projection}\label{subsec:variable-projection}
VarPro originated in numerical analysis as a technique for separable
nonlinear least-squares problems \citep{golub1976differentiation}, where a
subset of variables appears linearly in the residuals and can be
eliminated in closed form conditioned on the remaining variables. The
resulting reduced problem has fewer variables and typically improved
conditioning, yielding faster convergence of local methods
\citep{ruhe1980algorithms}. For readers seeking broader background on
VarPro, see \cite{golub2003Separable} for algorithms and applications and
\cite{ruhe1980algorithms} for convergence analysis.

\textbf{Schur complement.}
The most common VarPro approach in practice is the \emph{Schur complement
trick}: eliminate the linear variables via block elimination of the normal
equations, leaving a reduced problem in the remaining variables
\citep{triggs2000Bundle,zhang2006schur}. Schur-based methods can broadly be divided into explicit and implicit formulations. \emph{Explicit/direct Schur} methods fully form the reduced
matrix and optimize over the (dense) reduced problem.
This can be effective in SfM \citep{han2025Building,woodford2020Large} or in cases of camera calibration~\citep{Wise_2026} which are instances when the reduced problems are relatively
small.
%
%
\emph{Implicit/matrix-free} methods do not form the reduced matrix; they
apply it as an operator via a series of sparse matrix operations. This
better exploits problem sparsity and pairs naturally with iterative linear
solvers~\citep{saad2003iterative}. Matrix-free
Schur has been used in SfM
\citep{agarwal2010Bundle,weber2023Power,weber2024Power} and in SLAM
\citep{rosen2019SESync,holmes2023landmark}. For SLAM, \citep{khosoussi2016Sparse}
gives a sparsity-preserving scheme that does not use the Schur complement
but is iteration-equivalent to the Schur-based approach of
\cite{barham1972algorithm}.

\textbf{Projection/QR.}
Instead of modifying normal equations, projection methods \emph{remove the effect
of the linear variables} by projecting the residuals onto a subspace that implies
the optimal assignment of the linear variables (obtained by thin QR
factorization). From this the residual only depends on the remaining variables
and optimization proceeds on that reduced problem.  This
has been used in appearance modeling \citep{matthews2004Active} and computer
vision applications of matrix factorization
\citep{hong2017Revisiting,okatani2011Efficient,hong2016Projective, Demmel2021square}.
While numerically stable, the projection step can be costly and
the projected residuals generally have denser Jacobians than the original
problem, making sparsity exploitation more difficult.

\subsection{Placement of our work}
Our work is situated within a large body of literature on matrix-free Schur
complement methods~\citep{golub2003Separable,ruhe1980algorithms}, discussed
in~\cref{subsec:variable-projection}.
Our contribution is to show that (a)~a broad class of perception problems
(PGO, RA-SLAM, SNL, and SfM) admits a sparse, matrix-free Schur complement
operator despite the gauge symmetries that have limited prior approaches, and
(b)~this operator can be constructed efficiently by removing a single row and
column from specific problem matrices and performing a sparse Cholesky
factorization of the resulting matrix.
To our knowledge, only~\cite{rosen2019SESync} and~\cite{khosoussi2016Sparse} consider
problems with gauge symmetries while seeking to preserve sparsity.
We generalize~\cite{rosen2019SESync} to a broader class of problems within a
framework that makes explicit both the single requirement for the operator to
be applicable and the common structural conditions under which it can be
constructed with full sparsity-preserving efficiency.
Relative to~\cite{khosoussi2016Sparse}, which handles general separable
residuals, we consider a more specific residual class. This narrower scope is
deliberate, as it exposes additional structure that we exploit for greater
computational efficiency.
%


%% file: sec/4a_preliminary.tex
\def\matAf{\matA_{\text{f}}}
\def\matAc{\matA_{\text{c}}}
\def\matAfT{\matA_{\text{f}}^{\top}}
\def\matAcT{\matA_{\text{c}}^{\top}}

\section{Problem Formulation}
\label{sec:problem-formulation}

Many robotic perception problems estimate variables (e.g., robot poses, landmark
locations) from noisy measurements (e.g., odometry, visual observations, range
measurements) by minimizing a sum of squared residuals, where each residual
measures the discrepancy between a predicted and actual measurement.
We formalize this as a nonlinear least squares (NLS) problem, where the quantities
to estimate are represented as matrices.
\input{tab/common_residuals.tex}

\begin{problem}Nonlinear Least Squares (NLS) Estimation
\label{prob:nlls}
\begin{align}
    & \begin{alignedat}{2}
        & \min_{\matX = [\matXc; \matXf]} \quad &  & \quad \sum_{i=1}^{m} \|r_i(\matX) \|^2_{\matOmega_i} \\
        & \st                           &  & \quad \matXc \in \Xconstraint \subseteq \R^{n_c \times d}       \\
        &                               &  & \quad \matXf \in \R^{n_f \times d}
    \end{alignedat}\\
     & \matX \in \R^{n \times d} \triangleq \begin{bmatrix}
                            \matXc \\
                            \matXf
                        \end{bmatrix} \in \begin{bmatrix}
                                              \Xconstraint \subseteq \R^{n_c \times d} \\
                                              \R^{n_f \times d}
                                          \end{bmatrix},
\end{align}
where
$\matXc$ are variables constrained to the domain $\Xconstraint$,
$\matXf$ are unconstrained variables,
$r_i : \R^{n \times d} \to \R^{k_i \times d}$ is the residual function associated with the $i$-th measurement,
$\matOmega_i \in \PosDef^{k_i}$ is the positive definite concentration matrix
indicating the precision of the $i$-th measurement, and
$\|r_i(X)\|^2_{\matOmega_i} = \tr \left( r_i(X)^\top \matOmega_i r_i(X) \right)$ is a weighted squared Frobenius norm.
\end{problem}

This matrix-valued formulation generalizes the standard vector-valued NLS
formulation, enabling us to address problems
involving constrained variables that are difficult to express in vector form
(e.g., rotations satisfying $R^\top R = I$, which defines $\mathrm{O}(d)$ and
is necessary for $R \in \mathrm{SO}(d)$
\citep{rosen2019SESync,papalia2024Certifiably}).
We focus on a subclass of \cref{prob:nlls} where the residuals
are linear functions of the variables, enabling us to exploit
separable structure via a one-time preprocessing step.

\textbf{The case of linear residuals.}
\def\inhomogeneousFootnote{\footnote{We focus on the homogeneous
        case here for both simplicity and because there are no known inhomogeneous
        residuals that exhibit the graph structure we exploit in this work. However,
        the derivation for inhomogeneous residuals is straightforward, and we provide
        it in \cref{sec:appendix:quadratic-derivation}.}}
When residuals $r_i(X)$ are linear, the NLS problem
(\cref{prob:nlls}) reduces to a \emph{quadratic} cost.
We focus on \emph{homogeneous} linear residuals, i.e.,
$r_i(X) = A_i X$ for some matrix $A_i \in \R^{k_i \times n}$,
as (i) it simplifies notation, (ii)
many common perception residuals are homogeneous (see
\cref{tab:quadratic-factors}), and (iii) inhomogeneous residuals can be
made homogeneous through variable augmentation
\citep{luo2010semidefinite}.

By vertically stacking the Jacobians of the residuals $A_i$ and block-diagonally
arranging the concentration matrices $\matOmega_i$, we can rewrite the cost as
\begin{align}
     & \sum_i \big\| r_i(\matX)\big\|_{\matOmega_{i}}^2 \,=\, \,\big\|
    \matA\, \matX \big\|_{\matOmega}^2 \,=\, \tr \left(
    \matX^\top \matA^\top  \matOmega \matA
    \matX \right),                                                                             \\
     & \matA \triangleq \begin{bmatrix}\matA_1^\top \cdots \matA_m^\top \end{bmatrix}^\top ,\; \\
     & \matOmega \triangleq \text{blkdiag}(\matOmega_1, \ldots, \matOmega_m),
\end{align}
where the $\text{blkdiag}(\cdot)$ operator constructs a block-diagonal matrix
from its arguments.

We can reformulate the NLS problem (\cref{prob:nlls}) as
follows, where $\matQ \triangleq \matA^\top \matOmega \matA$:
\begin{problem}Constrained Quadratic Cost Problem
\label{prob:general-qcqp}
\begin{equation}
    \begin{alignedat}{2}
        & \min_{X = [\matXc;\, \matXf]} \quad && \tr(X^\top Q X)                            \\
        & \st                                 && \matXc \in \Xconstraint \subseteq \R^{n_c \times d} \\
        &                                     && \matXf \in \R^{n_f \times d}
    \end{alignedat}
\end{equation}
\end{problem}

Importantly, \cref{prob:general-qcqp} possesses a globally quadratic cost (i.e.,
is quadratic without linearization). This enables VarPro as a one-time preprocessing
step, as shown next.

\section{Separable Structure \& Variable Projection}
\label{sec:prelim:separable-structure}

Separable structure in an optimization problem refers to
the situation where, if certain variables are held fixed, the remaining
variables can be efficiently solved in closed-form. This structure partitions
the variables into two sets: those that are more difficult to optimize over
(e.g., due to constraints) and those that are easily optimized.

Variable projection (VarPro) methods
\citep{golub2003Separable} exploit separable structure by
iteratively optimizing over the `difficult' variables while implicitly
considering the `easy' variables at their optimal values conditioned on the
current iterate of the `difficult' variables.
VarPro effectively reduces the dimensionality of the problem to just the
`difficult' variables, and has been both theoretically and empirically shown to
improve convergence rates of optimization
\citep{ruhe1980algorithms}.

\begin{mybox}[title={Example: why eliminate the translations in pose-graph optimization?}]
In pose-graph optimization, the `difficult' variables $\matXc$ are the
rotations $R_i$, which must satisfy $R_i^\top R_i = I$, and the `easy'
variables $\matXf$ are the unconstrained translations $t_i$.
Each relative-pose measurement contributes the residual
$t_j - t_i - R_i\tilde{t}_{ij}$ (\cref{tab:quadratic-factors}), which is
linear in the translations.
For any fixed rotations, the best translations therefore solve a linear
least-squares problem with the closed-form solution
$\matXf^\star(\matXc)$ derived below (\cref{eq:optimal-unconstrained-variables}), and VarPro
replaces the joint problem with one over the rotations alone
(\cref{prob:reduced-qcqp}).

\textbf{Why this helps.} Rotations and translations are tightly coupled:
slightly rotating one pose swings every translation after it along the
trajectory, like a lever arm. A solver working on the joint problem must
discover this coordinated motion over many iterations. In the reduced problem
the translations jump to their best values at every step, so the solver only
has to find the rotations. Because the cost is quadratic, the map
$\matXf^\star(\matXc)$ is linear and the same at every iterate, so the
elimination can be computed once, before optimization begins.
\end{mybox}

The problems we consider exhibit separable
structure that is particularly well suited for VarPro.  Because
the `easy' variables are unconstrained and the cost is quadratic, a
closed-form elimination of the unconstrained variables can be performed once and
holds at every iteration. This
requirement excludes certain robotics perception problems composed entirely of constrained variables,
such as pure rotation averaging~\citep{hartley2013rotation,dellaert2020shonan}, but it makes it a natural fit for SLAM, SfM, and SNL where there exist many unconstrained variables such as translations and landmarks which can be eliminated.

We now show how those unconstrained variables in a quadratic cost induce separable
structure and how the Schur complement can eliminate these
variables.

\textbf{Step 1: Solve for the unconstrained variables.}
With the variable ordering $X = [\matXc^\top \mid \matXf^\top]^\top$, we can
partition the matrix $\matQ$ as
\begin{equation}
    \label{eq:Q-partition}
    \matQ = \begin{bmatrix}
        \matQcc & \matQcf \\
        \matQfc & \matQff
    \end{bmatrix}.
\end{equation}
For a fixed $\matXc$, we set the gradient of the cost
with respect to the unconstrained variables $\matXf$ to zero and solve for the
optimal value $\matXfstar$ as a function of $\matXc$:
\begin{align}
     & \frac{\partial}{\partial \matXf} \tr(X^\top Q X) = 2 \matQff \matXf + 2 \matQfc \matXc = 0, \\
     & \matXfstar = - \matQff^{\dagger} \matQfc \matXc, \label{eq:optimal-unconstrained-variables}
\end{align}
where $\matQff^{\dagger}$ is the Moore-Penrose pseudoinverse of $\matQff$.
\begin{remark}
This exact solution of $\matXfstar$ requires that $\matQff \matXf = -\matQfc \matXc$ is
consistent (solvable for all $\matXc$), which holds iff
$\operatorname{range}(\matQfc) \subseteq \operatorname{range}(\matQff)$.
From \cref{eq:Q-expanded-jacobians}, $\matQff = \matAfT \matOmega \matAf$
and $\matQfc = \matAfT \matOmega \matAc$. Since $\matOmega \succ 0$, it follows that
$\operatorname{range}(\matQff) = \operatorname{range}(\matAfT)$ and
that $\operatorname{range}(\matQfc) \subseteq \operatorname{range}(\matAfT)$, thus
implying $\operatorname{range}(\matQfc) \subseteq \operatorname{range}(\matQff)$.
Additionally, since $\matQff$ may be rank-deficient and the system may have
multiple solutions, the pseudoinverse in
\cref{eq:optimal-unconstrained-variables} returns the minimum-norm solution
$\matXfstar$.
\end{remark}

\textbf{Step 2: Eliminate them from the cost.}
Plugging this optimal value $\matXfstar$ into the cost, we obtain a cost
that depends only on the constrained variables $\matXc$:
\begin{equation}
    \label{eq:reduced-cost}
    \tr(\matXc^\top \matQmarg \matXc)  =
    \tr\left(\matXc^\top \left(\matQcc - \matQcf \matQff^{\dagger} \matQfc\right) \matXc\right),
\end{equation}
where $\matQmarg \triangleq \matQcc - \matQcf \matQff^{\dagger} \matQfc$
is commonly called the \emph{Schur complement} \citep{zhang2006schur} of
$\matQff$ in $\matQ$.

\textbf{Step 3: Form the reduced problem.}
Minimizing the reduced cost over the constraint set gives a problem in which
the unconstrained variables no longer appear:
\begin{problem}[\emph{Reduced} Constrained Quadratic Cost Problem]
\label{prob:reduced-qcqp}
\begin{equation}
    \min_{\matXc \in \Xconstraint} \quad \tr(\matXc^\top \matQmarg \matXc)
\end{equation}
\end{problem}

\textbf{Computational challenges in the reduced cost.}
%
Steps~1--3 yield a reduced problem (\cref{prob:reduced-qcqp}) in the
constrained variables $\matXc$ alone. This reduction both shrinks the problem
and reduces the number of iterations needed to converge
\citep{ruhe1980algorithms}. However, it also introduces two obstacles. First,
forming the Schur complement $\matQmarg$ requires a pseudoinverse, which is
costly and numerically fragile. Second, $\matQmarg$ is typically dense even
when the original matrix $\matQ$ is sparse, making storage and operations
expensive. Either issue can erase the benefits of the reduction, since a
smaller problem that needs fewer iterations is no faster overall if forming
$\matQmarg$ is expensive or each iteration requires dense products with it. We
avoid these challenges by leveraging iterative methods that do not require
explicitly forming $\matQmarg$.

\begin{mybox}[title={Proposed approach: implicit (matrix-free) Schur via iterative methods}]
Rather than explicitly forming the dense Schur complement $\matQmarg$,
we solve the reduced problem (\cref{prob:reduced-qcqp}) using iterative
methods \citep{saad2003iterative}, which only require computing
matrix-vector products $\matQmarg\,\matXc$.
Our key contribution is an efficient, matrix-free routine for computing
these products without ever forming $\matQmarg$ explicitly or computing
a pseudoinverse.
Specifically, we derive an exactly equivalent reformulation of $\matQmarg$
that can be applied via sparse matrix operations and triangular solves
with Cholesky factors of the original problem data.
\end{mybox}

\section{Matrix-Free Schur Complement Products}
\label{sec:prelim:focus-schur-complement-products}

This section identifies the specific term that creates computational
challenges in computing Schur complement products $\matQmarg \matXc$, namely a
pseudoinverse that creates a large, dense matrix. We then show how
least-squares structure admits a reformulation that
replaces the pseudoinverse with a positive definite matrix inverse, which can be
efficiently represented via Cholesky factorizations. This reformulation
allows us to compute products with $\matQmarg$ via a series of sparse matrix
products and triangular solves.

We revisit the quadratic cost matrix $\matQ = \matA^\top {\Omega} \matA$.  By
ordering the variables as $X = [\matXc^\top \mid \matXf^\top]^\top$, the
Jacobian matrix becomes $\matA = [\matAc \mid \matAf ]$, where $\matAc$ and
$\matAf$ are the stacked Jacobians of the residuals with respect to the
constrained and unconstrained variables, respectively. This then reframes
$\matQ$ and the Schur complement $\matQmarg$ as:
\begin{align}
    \label{eq:Q-expanded-jacobians}
     & \matQ = \begin{bmatrix}
                   \matQcc                  & \matAcT \matOmega \matAf \\
                   \matAfT \matOmega \matAc & \matAfT \matOmega \matAf
               \end{bmatrix}, \\
     & \matQmarg = \matQcc -
    \matAcT \matOmega \matAf
    (\matAfT \matOmega \matAf)^{\dagger}
    \matAfT \matOmega \matAc.
\end{align}

\def\matQtwo{\matQ_{2}}
Importantly, $\matQ$ and the constituent matrices
$\matQcc,\matAc,\matAf,\matOmega$ naturally inherit the sparsity of the
graphical structure of the underlying problem.
%
We can write products with the Schur complement $\matQmarg \matXc$ as:
\begin{align}
    \label{eq:schur-product-dense-pseudoinverse}
     & \underbrace{\matQmarg}_{\text{dense}} \matXc =
    \underbrace{\matQcc}_{\text{sparse}} \matXc
    -
    \underbrace{
        \Big(
        \underbrace{\matAcT \matOmega \matAf}_{\text{sparse}}
        \underbrace{
            (\matAfT \matOmega \matAf)^{\dagger}
        }_{\text{dense}}
        \underbrace{\matAfT \matOmega \matAc}_{\text{sparse}}
        \Big)
    }_{\matQtwo \text{ (dense)}}
    \matXc,                \\
     & \matQtwo \triangleq
    \matAcT \matOmega \matAf
    (\matAfT \matOmega \matAf)^{\dagger}
    \matAfT \matOmega \matAc,
\end{align}
where we have highlighted the sparsity of each term
assuming the original problem is sparse.

\begin{mybox}[title=Key challenge: dense pseudoinverse in Schur complement products.]
    \cref{eq:schur-product-dense-pseudoinverse} emphasizes that the pseudoinverse
$(\matAfT \matOmega \matAf)^{\dagger}$ creates a dense matrix that is the
computational bottleneck in computing products with the Schur complement.
If $(\matAfT \matOmega \matAf)$ were full rank and positive definite,
as in many VarPro applications \citep{golub2003Separable},
the pseudoinverse becomes an inverse. Since $(\matAfT \matOmega
    \matAf)$ inherits the sparsity of the residuals, this inverse could be
efficiently computed via sparse Cholesky factorization and
applied via sparse triangular solves. However, in many robotic perception
problems the matrix is rank-deficient due to inherent symmetries.
\end{mybox}

\subsection{Exact Reformulation of $\matQtwo$ via CR Decomposition}
\label{sec:prelim:graphical-structure}

Our reformulation relies on the $CR$ decomposition of a matrix
\citep{strang2022Three,strang2024Elimination}, which factorizes a matrix by its
column and row spaces and can be computed e.g., via rank-revealing QR
decomposition \citep{golub2013matrix}.
We first preform a CR factorization of $\matAf$ as
\begin{equation}
    \matAf = \matC \matR,
\end{equation}
where the columns of $\matC$ form a basis for the column space of $\matAf$ and
the rows of $\matR$ form a basis for the row space of $\matAf$.  The matrix
$\matC$ has full column rank and the matrix $\matR$ has full row rank.
Using this decomposition, we can rewrite the inner portion of $\matQtwo$ as
\begin{equation}
    \label{eq:Q2-reformulation-start}
    \matAf (\matAfT \matOmega \matAf)^{\dagger} \matAfT
    = (\matC \matR) (\matR^\top \matC^\top \matOmega \matC \matR)^{\dagger} (\matC \matR)^\top .
\end{equation}

\def\matM{\mat{M}}

Let $\matM \triangleq \matC^\top \matOmega \matC$. Since $\matOmega \succ 0$
and $\matC$ has full column rank, $\matM$ is positive definite and therefore
invertible. Because $\matR$ has full row rank, the following identity holds
\begin{equation}
    \label{eq:RTMR-pinv}
    (\matR^\top \matM \matR)^{\dagger} =
    \matR^{\dagger} \matM^{-1} (\matR^{\dagger})^\top .
\end{equation}
Substituting \cref{eq:RTMR-pinv} into \cref{eq:Q2-reformulation-start} and
recognizing that $\matR \matR^{\dagger} = I$ (since $\matR$ has full row rank)
yields
\begin{align}
    \matAf (\matAfT \matOmega \matAf)^{\dagger} \matAfT
     & = \matC \matR \left(\matR^{\dagger} \matM^{-1} (\matR^{\dagger})^\top \right) \matR^\top \matC^\top \\
     & = \matC (\matR \matR^{\dagger}) \matM^{-1} (\matR \matR^{\dagger})^\top \matC^\top \\
     & = \matC \matM^{-1} \matC^\top \\
     & = \matC L^{-\top} L^{-1} \matC^\top,
    \label{eq:Q2-final}
\end{align}
where the last step replaces the inverse via Cholesky factorization $\matM = LL^\top$.

Plugging this reformulation into $\matQtwo$ in
\cref{eq:schur-product-dense-pseudoinverse} obtains
\begin{equation}
    \label{eq:Qmarg-product-final}
    \begin{aligned}
        \matQmarg \matXc
         & =
        \matQcc \matXc -
        \left( \matAcT \matOmega \matC \right)
        L^{-\top} L^{-1}
        \left(\matC^\top \matOmega \matAc\right)
        \matXc \\
         & =
        \matQcc \matXc - \left(B L^{-\top} L^{-1} B^\top\right) \matXc,
    \end{aligned}
\end{equation}
where $B \triangleq \matAcT \matOmega \matC$.
This can be performed as a series of matrix products and forward- and
back-substitution with the Cholesky factors $L$ and $L^{\top}$ as described in
\cref{alg:schur-complement-product}.

\input{alg/schur_complement_product.tex}

\textbf{When is this reformulation valid?}
The reformulation in \cref{eq:Qmarg-product-final} is exact and only
depends on the least-squares structure (which induces the block
structuring of $\matQ$ in \cref{eq:Q-partition}). This could be applied at each
iteration of an iterative solver that linearizes the problem,
though this incurs the cost of recomputing $\matC$ and the Cholesky
factorization. Because our problems have linear
residuals, we perform this reformulation once as preprocessing.

\textbf{When is the Schur product in \cref{eq:Qmarg-product-final} efficient?}
Efficiency is driven by the sparsity of
$\matC$, which determines the density of both $B=\matAcT \matOmega \matC$
and the Cholesky factors $L^{-\top} L^{-1}$.
As $\matC$ is a linearly independent basis for the column space of $\matAf$,
there are many ways to compute it with sparsity in mind \citep{coleman1987null}.
These approaches typically require either non-trivial
computation (e.g., factorization) or \emph{a priori} knowledge of the rank of
$\matAf$ (e.g., to determine stopping in greedy algorithms).

\subsection{Leveraging Graph Structure for Efficient Schur Products}
\label{sec:prelim:efficient-schur-complement-products}

\begin{figure*}[t]
  \centering
  \includegraphics[width=0.8\textwidth]{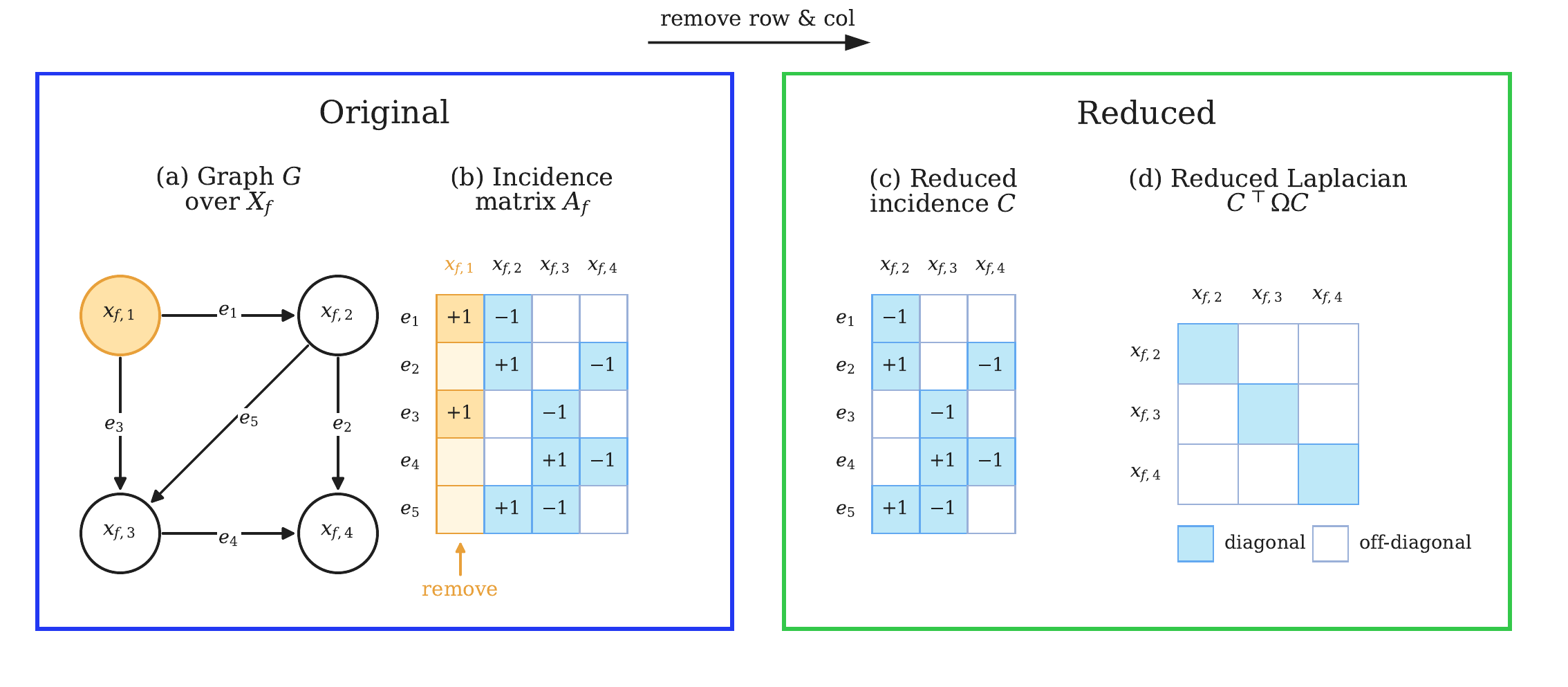}
\caption{\textbf{Graph-theoretic construction of the reduced incidence $C$ and Laplacian.}
  The incidence matrix $A_f$ has a one-dimensional null space (the gauge of relative measurements); dropping one column of $A_f$, and the corresponding row and column of the Laplacian, removes it.
  \textbf{(a)} Toy graph $G$ over $X_f$, with $x_{f,1}$ (orange) chosen as the anchor.
  \textbf{(b)} In $A_f$, each row is a $+1$/$-1$ pair. Removing the shaded column of $x_{f,1}$ leaves a full-rank basis.
  \textbf{(c)} Reduced incidence $C$.
  \textbf{(d)} $C^\top \Omega C$ deletes the same row and column from $A_f^\top \Omega A_f$. Unlike $A_f^\top \Omega A_f$ (rank-deficient, needs a pseudoinverse), $C^\top \Omega C$ is sparse and positive definite, admitting the Cholesky factorization used in Algorithm~\ref{alg:schur-complement-product}.}
  \label{fig:graph-structure}
\end{figure*}

We now establish specific, yet common, graph-theoretic conditions on the
Jacobian $\matAf$ under which the implicit Schur complement product in
\cref{eq:Qmarg-product-final} retains the sparsity of the original problem.
The two conditions are that the graph over the unconstrained variables is
connected and that each residual involves those variables only through a
pairwise difference. Under these conditions, $\matAf$ is the incidence matrix
of this graph and $\matAfT \matOmega \matAf$ is its weighted graph Laplacian.
Standard results from spectral graph theory then give $\matC$ and
$\matC^\top \matOmega \matC$ in closed form as the \emph{reduced} incidence
matrix and \emph{reduced} graph Laplacian. These are obtained by deleting one
column of $\matAf$ and the corresponding row and column of the Laplacian
(\cref{fig:graph-structure}), making the reformulation in
\cref{eq:Qmarg-product-final} efficient.

Consider a graph $G$ where each unconstrained variable is a node and any two
variables that appear together in a residual are connected by an edge.
In every well-posed problem we demonstrate (that is, one with a unique solution
up to gauge symmetry), $G$ is connected. This is expected, as a disconnected
graph introduces gauge freedoms beyond the global one, which breaks uniqueness.

For residuals based on relative differences between pairs of unconstrained
variables (i.e., $r_i(\matX) = g(\matXc) + (\matX_{fi} - \matX_{fj})$ for some
function $g(\matXc)$), the Jacobian $\matAf$ is the \emph{incidence matrix} of
$G$. Each row of $\matAf$ corresponds to an edge in $G$ and each column to a
variable (node). For a row corresponding to an edge between nodes $i$ and $j$,
the $k$-th entry is
\begin{equation}
    \matA_{\text{f}, (i,j)} (k) = \begin{cases}
        1,  & k = i            \\
        -1, & k = j            \\
        0,  & \text{otherwise}
    \end{cases}.
\end{equation}

\begin{remark}
The rank-deficiency of $\matAf$ is because
the graphs inherently capture relative information. This can be seen
from the fact that the all-ones vector spans the null space of $\matAf$;
adding a constant offset to all variables does not affect relative
differences.
\end{remark}

We leverage one key property of incidence matrices for connected graphs.
Removing any one column of the incidence matrix yields a linearly independent
basis for its column space \citep{chung1997spectral}. The resulting reduced
incidence matrix is therefore a natural choice for $\matC$ in
\cref{eq:Qmarg-product-final}.

The same deletion carries over to the Laplacian. The weighted graph Laplacian
$\matAfT \matOmega \matAf$ is positive semidefinite with a single zero
eigenvalue \citep{chung1997spectral}. When $\matC$ is a reduced incidence
matrix, the product $\matC^\top \matOmega \matC$ is the reduced graph
Laplacian, which is positive definite \citep{chung1997spectral}. It is formed
by removing from the Laplacian the row and column corresponding to the column
removed from $\matAf$ to form $\matC$.

Because both the reduced incidence matrix and the reduced graph Laplacian are
formed by deleting a column of $\matAf$ and a row and column of
$\matAfT \matOmega \matAf$, they are easily constructed and inherit the
sparsity of the original matrices. With this choice of $\matC$, the only
non-negligible computation required to form the operator for the Schur
complement product (as in \cref{alg:schur-complement-product}) is a sparse
Cholesky factorization of the reduced graph Laplacian.
\input{tab/dense_vs_sparse_marg}

In \cref{fig:schur-matrix-free}, we visualize the sparsity patterns of the
matrices involved in the Schur complement product $\matQmarg \matXc$
for a specific problem instance. In contrast to forming
the dense Schur complement $\matQmarg$, the reformulation in
\cref{eq:Qmarg-product-final} retains the sparsity of the original problem.

If $\matAf$ is not incidence-like (i.e., $\matXf$ does not appear in residuals
solely through pairwise differences of its entries),
$\matC$ must be computed using basis-construction techniques
\citep{coleman1987null}, which may incur additional computational cost and yield
a denser $\matC$; in either case the reformulation in
\cref{eq:Qmarg-product-final} remains exact.

\subsection{Probabilistic Interpretation of Separable Structure}
\label{sec:prelim:probabilistic-interpretation}

The development above takes an optimization-facing, linear-algebraic view of the
implicit Schur complement. The same objects and operations also admit a
probabilistic reading, which we develop here to build intuition and to connect the
construction to broader themes in estimation.

\textbf{Gaussian graphical models.}
Quadratic costs of the form $\tr(X^\top \matQ X)$ typically arise as the
maximum-likelihood objective for linear-Gaussian models~\citep{dellaert2017factor}, so
$\tfrac{1}{2}\tr(X^\top \matQ X)$ can be read as the negative log-density of a Gaussian
with precision matrix $\matQ$ (up to an additive constant). Under this reading, the
sparsity pattern of $\matQ$ is exactly the conditional-independence structure among the
variables, with $\matQ_{ij} = 0$ if and only if $X_i$ and $X_j$ are conditionally
independent given all other variables~\citep{koller2009probabilistic}. This is the same
structure encoded by the factor graph familiar from robotic
perception~\citep{dellaert2017factor}, where two variables are adjacent precisely when
they share a factor. The graph over the unconstrained variables that drives our
efficient construction (\cref{sec:prelim:efficient-schur-complement-products}) is the
conditional-independence graph this structure induces on $\matXf$ alone, i.e., the sparsity pattern of the block $\matQff$.

\textbf{Variable projection as conditioning.}
The joint distribution over $X = [\matXc; \matXf]$ is not Gaussian in general, because
the constrained variables are manifold-valued and their measurements (e.g., relative
rotations) are modeled by distributions such as the Langevin~\citep{cis/1241018500, rosen2019SESync}
rather than a Gaussian. What holds exactly is the weaker
\emph{conditional-linear-Gaussian} property~\citep{khosoussi2016Sparse}. With $\matXc$
fixed, every residual is linear in $\matXf$, so $p(\matXf \mid \matXc)$ is Gaussian, and
when $\matQff$ is nonsingular it has mean
$\mathbb{E}[\matXf \mid \matXc] = -\matQff^{-1}\matQfc\matXc$ and covariance
$\mathrm{Cov}(\matXf \mid \matXc) = \matQff^{-1}$~\citep{barfoot2017state}. This
conditional is exactly the closed-form elimination of
\cref{eq:optimal-unconstrained-variables}. The optimal $\matXfstar$ is the conditional
mean, and because that mean is linear in $\matXc$ while its covariance is independent of
$\matXc$, the unconstrained block can be eliminated once and reused at every $\matXc$.

The pseudoinverse in $\matXfstar$ is the probabilistic counterpart of the same gauge
symmetry. When the gauge of remark 2 renders $\matQff$ rank-deficient, the
conditional $p(\matXf \mid \matXc)$ becomes improper. Its precision has a zero eigenvalue along the gauge, the
variance there is unbounded, and the conditional mean is no longer unique. The
minimum-norm representative $\matXfstar = -\matQff^{\dagger}\matQfc\matXc$ is the
conditional MAP estimate after gauge fixing, with $\matQff^{\dagger}$ the covariance
on the observable, gauge-orthogonal subspace \citep{930934,8354808}. The
degeneracy is harmless, since the flat direction is exactly the unobservable gauge,
so choosing a representative discards no information.

\textbf{The Schur complement as marginalization.}
Where conditioning fixes $\matXf$, marginalization integrates it out. The resulting
marginal over $\matXc$ has information matrix equal to the Schur complement
$\matQmarg = \matQcc - \matQcf\matQff^{\dagger}\matQfc$~\citep{barfoot2017state},
\begin{equation}
    p(\matXc) = \int p(\matXc, \matXf)\,d\matXf
    \;\propto\; \exp\!\left(-\tfrac{1}{2}\tr(\matXc^\top \matQmarg \matXc)\right),
    \label{eq:marginal-Xc}
\end{equation}
and this is the object our matrix-free routine applies without ever forming it. As with
conditioning, two caveats keep the reading exact. Because $\matXc$ is manifold-valued,
\cref{eq:marginal-Xc} is really the marginal negative log-density over $\matXc$ (its
quadratic part), and when $\matQff$ is singular the integral diverges along the $\matXf$
gauge and must be read after gauge fixing, that is, on the
observable subspace. That gauge fixing is exactly the reduced-incidence construction of
\cref{sec:prelim:efficient-schur-complement-products}, with $\matQff^{\dagger}$ its
minimum-norm surrogate, so the algebraic Schur complement stays exact whether or not
$\matQff$ is invertible.

\textbf{Marginalization densifies.}
As established, forming $\matQmarg$ marginalizes out the unconstrained variables,
typically the large and well-connected set of translations or landmarks. Marginalizing
a variable couples all of its former neighbors~\citep{6898876},
since dependencies once routed through it become direct edges, and the sparse joint $\matQ$
therefore gives way to a generally dense marginal $\matQmarg$~\citep{koller2009probabilistic}. Because
the eliminated set touches nearly every constrained variable, this fill-in is severe.
It is the probabilistic form of the densification shown in
\cref{eq:schur-product-dense-pseudoinverse} and \cref{fig:schur-matrix-free}, and
precisely what our matrix-free operator is designed to avoid.

%% file: tab/common_residuals.tex
\begin{table}[t]
    \renewcommand{\arraystretch}{1.5}
    \vspace{0.5em}
    \caption{Example linear residuals in state-of-the-art formulations.
        In these problems the variables are: $R_i$ (rotation matrices),
        $t_i$ (translation vectors), $u_{ij}$ (unit bearing vectors),
        $RS_i$ (scaled rotation matrices), and $\alpha$ (an unknown scale).
        Rotations are constrained by $R_i^\top R_i = I$, which is necessary
        for membership in both $\mathrm{O}(d)$ and $\mathrm{SO}(d)$
        ($\mathrm{SO}(d)$ additionally requires $\det R_i = 1$).
        The quantities $\tilde{R}_{ij}, \tilde{t}_{ij}, \tilde{u}_{ij},
        \tilde{R}_{B_{ij}}, \tilde{t}_{B_{ij}}$ are noisy measurements of
        the same objects and $\tilde{d}_{ij}$ are noisy distance measurements.
    }
    \label{tab:quadratic-factors}
    \centering
    \begin{tabularx}{\linewidth}{@{}>{\raggedright\arraybackslash}X l@{}}
        \toprule
        \textbf{Measurement Type} & \textbf{Residual $r_{ij}(\matX)$} \\
        \midrule
        Relative rotation
            & $R_j - R_i\,\tilde{R}_{ij}$ \\
        Relative translation
            & $t_j - t_i - R_i\,\tilde{t}_{ij}$ \\
        Scale-free relative translation
            & $t_j - t_i - RS_i\,\tilde{t}_{ij}$ \\
        Range
            & $t_j - t_i - u_{ij}\,\tilde{d}_{ij}$ \\
        \bottomrule
    \end{tabularx}

    \vspace{0.25em}
    \parbox{\linewidth}{\footnotesize
        Relative rotation and translation~\citep{rosen2019SESync};
        scale-free relative translation~\citep{han2025Building};
        range~\citep{papalia2024Certifiably,halsted22arxiv}.
    }
    \vspace{-1.5em}
\end{table}

%% file: alg/schur_complement_product.tex
\begin{algorithm}[t]
    \caption{Matrix-Free Schur Complement Products}
    \label{alg:schur-complement-product}
    \textbf{Input:} Matrices $\matQcc$, $\matAc$, $\matAf$, $\matOmega$, and $\matXc$. \\
    \textbf{Output:} Product $\matQmarg \matXc$

    \textbf{Precomputation} (once, given $\matQcc$, $\matAc$, $\matAf$, $\matOmega$):
    \begin{algorithmic}[1]
        \State $\matC \gets \text{CR} (\matAf)$ \Comment{CR decomposition (\cref{sec:prelim:focus-schur-complement-products})}
        \State $L \gets \text{Cholesky}(\matC^\top \matOmega \matC)$ \Comment{sparse, $LL^\top = \matC^\top \matOmega \matC$}
        \State $B \gets \matAcT \matOmega \matC$
    \end{algorithmic}
    \textbf{Online computation} (per new $\matXc$):
    \begin{algorithmic}[1]
        \State $Y \gets B^\top \matXc$ \Comment{sparse matrix product}
        \State $Z \gets L^{-\top} (L^{-1} Y)$ \Comment{forward, then back substitution}
        \State \Return $\matQmarg \matXc = \matQcc \matXc - B Z$
    \end{algorithmic}
\end{algorithm}

%% file: tab/dense_vs_sparse_marg.tex
\setlength{\fboxsep}{0pt}%
\setlength{\fboxrule}{0.5pt}%
\newlength{\cholsize}
\setlength{\cholsize}{0.35cm}

\newcommand{\MatrixImg}[2]{%
    \fbox{\adjustbox{valign=c}{%
            \includegraphics[height=\dimexpr #1\cholsize\relax]{#2}%
        }}%
}

\newcommand{\QMainSparse}{\MatrixImg{4}{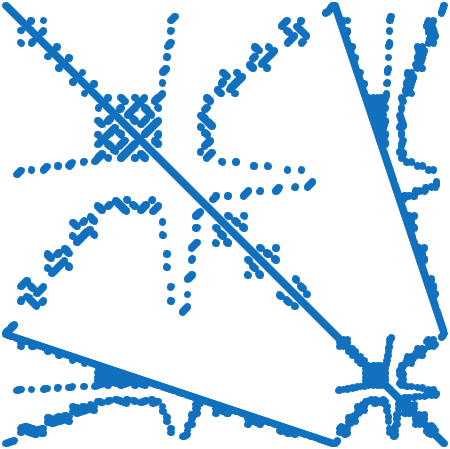}}
\newcommand{\QOneSparse}{\MatrixImg{3}{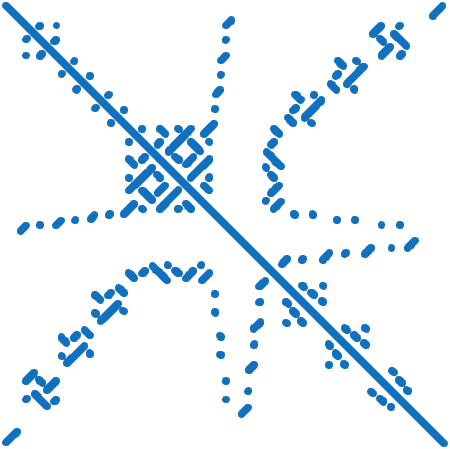}}
\newcommand{\QTwoDense}{\MatrixImg{3}{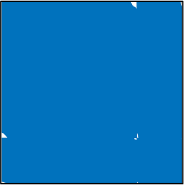}}
\newcommand{\BSparse}{\MatrixImg{3}{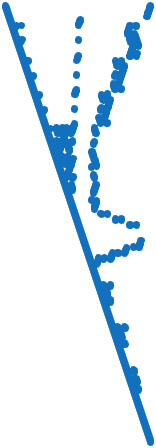}}
\newcommand{\LSparse}{\MatrixImg{1}{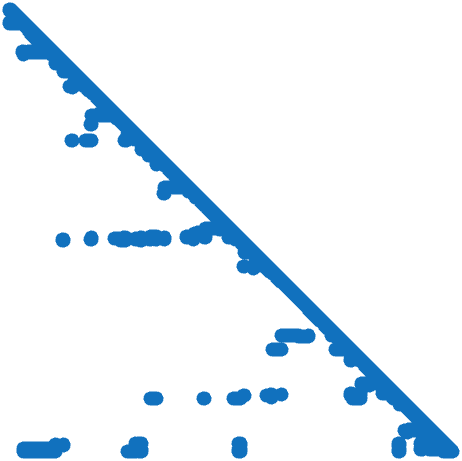}}

\def\SparseOp{\left( \BSparse \underbrace{ \left(\LSparse\right)^{\shortminus 1} \left(\LSparse\right)^{\shortminus\top} }_{\text{Cholesky Factors}} \BSparse^\top \right)}
\newcommand{\wrapWords}[2]{\makecell[l]{#1 \\ #2}}

\def\SparseFullOp{\QOneSparse  - \SparseOp }
\def\DenseFullOp{\QTwoDense  }

\renewcommand{\arraystretch}{1}
\begin{figure}[t]
    \vspace{0.5em}
    \centering
    \begin{tabularx}{\linewidth}{L{1.7 cm} M}
                                                                          & \textbf{Schur Complement Products} \\
        \midrule
        \wrapWords{\textbf{Original}}{\textbf{Matrix} $\mathbf{(\matQ)}$} & \QMainSparse                               \\
        \midrule
        \wrapWords{\textbf{Explicit}}{\small{(Dense)}}                    & \DenseFullOp                               \\ \addlinespace[4pt]
        \wrapWords{\textbf{Implicit}}{\small{(Sparse)}}                   & \SparseFullOp                              \\
        \bottomrule
    \end{tabularx}
    \caption{
        \textbf{Matrix-Free Schur Complement Products:}
        Here we demonstrate the difference between explicitly forming the Schur
        complement $\matQmarg$ as a dense matrix versus performing a series of
        sparse operations to implicitly compute the product $\matQmarg \matXc$
        without forming the matrix.
        The top row shows the sparsity pattern of the original matrix $\matQ$.
        The bottom two rows show the sparsity patterns of the dense (explicit)
        and sparse (implicit) Schur complement approaches with the data
        from the `Garage' dataset \citep{carlone2015Initialization}.
    }
    \label{fig:schur-matrix-free}
    \vspace{-1.5em}
\end{figure}

%% file: sec/4b_nonlinear.tex
\section{Robust Loss: A Special Case of Nonlinear Residuals}
\label{sec:robust-loss}

The framework developed in \cref{sec:prelim:focus-schur-complement-products} and specifically in \cref{sec:prelim:graphical-structure} relies on residuals that are linear functions of the variables. This linearity yields a globally quadratic cost, which in turn allows the matrix-free Schur complement operator to be constructed once, in the preprocessing step.
In practice, however, many robotic perception problems use \emph{robust} cost
functions to limit the influence of outlier measurements (e.g., spurious loop
closures in SLAM~\citep{sunderhauf2012Switchable,sh-ch3-outlier}, mismatched feature
correspondences in SfM~\citep{yang2020Graduated}, or multipath measurements in ranging~\citep{jiang2024Robust, olson2006Robust}. Robust costs are nonlinear
functions of the residuals, and at first glance appear to invalidate the
structural assumptions underlying \cref{eq:Qmarg-product-final}.

In this section we show that robust costs are in fact a particularly favorable
case of nonlinear residuals. When handled via \emph{iteratively reweighted least
squares} (IRLS) \citep{green1984iteratively,sh-ch3-outlier}, each outer iteration
is a weighted linear least-squares problem in which reweighting acts only by
rescaling each measurement's contribution through a positive scalar weight. As a
result, the Jacobian $\matAf$ and the variable partition $X = [X_c; X_f]$ are
\emph{identical} across outer iterations, while the cost matrix $Q$ retains its
sparsity pattern and only its nonzero entries change. Consequently, the bulk of
our preprocessing (\cref{alg:schur-complement-product}) is reusable across IRLS
iterations, and each outer iteration requires only a numerical refactorization of the
reduced Laplacian $M^{(k)} \triangleq C^\top \tilde\Omega^{(k)} C$ rather than a full
reformulation.

\subsection{Robust Cost via IRLS}
\label{sec:robust-loss:irls}

We replace the quadratic cost in \cref{prob:nlls} with a robust cost
\begin{equation}
    \label{eq:robust-cost}
    \min_{\matX = [\matXc; \matXf]} \;
    \sum_{i=1}^m \rho\!\left(\| r_i(\matX)\|^2_{\matOmega_i}\right),
\end{equation}
where $\rho : \R_{\geq 0} \to \R$ is a robust kernel: a differentiable,
nondecreasing, concave function with $\rho(0)=0$ (e.g., Huber, Cauchy, or
Geman--McClure (GM)~\citep{sh-ch3-outlier})\footnote{Due to this property, a common robust kernel, Truncated Least Squares, is not applicable to our method.}, and $\rho'$ denotes its derivative.
Each such kernel grows more slowly
than the quadratic penalty $\|r_i\|^2_{\matOmega_i}$, so residuals that are large
relative to the measurement noise (typically outliers) contribute far less to
the total cost than they would under ordinary least squares.

IRLS minimizes \cref{eq:robust-cost}
by solving a sequence of \emph{outer iterations}, each of which is an ordinary
weighted least-squares problem; under standard assumptions the resulting iterates
converge to a stationary point of the robust cost. At outer iteration $k$, given
the current iterate $\matX^{(k)}$, we first hold the residuals fixed and assign
each measurement a scalar weight~\citep{black1996unification}
\begin{equation}
    \label{eq:irls-weights}
    w_i^{(k)}
    = \frac{\rho'\!\left(\|r_i(\matX^{(k)})\|^2_{\matOmega_i}\right)}
           {\|r_i(\matX^{(k)})\|^2_{\matOmega_i}},
\end{equation}
and then obtain the next iterate by solving the reweighted quadratic subproblem
\begin{equation}
\label{eq:irls-subproblem}
\begin{aligned}
    \matX^{(k+1)}
    &\in \argmin_{\matX} \; \sum_{i=1}^m w_i^{(k)} \|r_i(\matX)\|^2_{\matOmega_i} \\
    &= \argmin_{\matX} \; \tr\!\left(\matX^\top \matA^\top
        \widetilde{\matOmega}^{(k)} \matA\,\matX\right),
\end{aligned}
\end{equation}
where the reweighted concentration matrix is
\begin{equation}
    \label{eq:reweighted-omega}
\widetilde{\matOmega}^{(k)} \triangleq
\text{blkdiag}\!\left(w_1^{(k)} \matOmega_1,\,\ldots,\, w_m^{(k)} \matOmega_m\right).
\end{equation}
These two steps (reweight, then resolve) are repeated until the iterate
(equivalently, the weights) converges. Solving a single subproblem
\cref{eq:irls-subproblem} is itself an iterative process: we refer to the
reweight-and-resolve steps indexed by $k$ as the \emph{outer} iterations, and to
the iterations of the solver used within a fixed subproblem as the \emph{inner}
iterations (or \emph{inner solves}). This inner/outer distinction is what the
amortized-cost discussion at the end of \cref{sec:robust-loss:preserved-structure}.

Each IRLS subproblem (\cref{eq:irls-subproblem}) is an instance of
\cref{prob:general-qcqp} in which \emph{only} the diagonal weight matrix
$\widetilde{\matOmega}^{(k)}$ depends on the iterate. The Jacobians $\matAc$
and $\matAf$, the variable partition $\matX = [\matXc; \matXf]$, and the
constraint set $\Xconstraint$ are all unchanged.

\subsection{Reusable Preprocessing under IRLS}
\label{sec:robust-loss:preserved-structure}

The reformulation of $\matQmarg$ in \cref{eq:Qmarg-product-final} is determined
by three quantities: the column-space basis $\matC$ of $\matAf$, the matrix
$\matM \triangleq \matC^\top \matOmega \matC$ (and its Cholesky factor $L$),
and the cross term $B = \matAcT \matOmega \matC$. Reweighting affects these
quantities differently:
\begin{enumerate}
    \item 

\textbf{$\matC$ is unchanged.} The column-space basis depends only on the
graph topology of $\matAf$, which is determined by the residual structure and
not by any iterate-dependent quantity. When $\matAf$ is incidence-like
(\cref{sec:prelim:efficient-schur-complement-products}), $\matC$ is the reduced
incidence matrix; in general, it is whatever basis was constructed during
preprocessing. In either case, $\matC$ is computed once and reused across all
IRLS iterations.

\item
\textbf{The sparsity pattern of $\matM^{(k)} \triangleq \matC^\top
\widetilde{\matOmega}^{(k)} \matC$ is unchanged.} Because
$\widetilde{\matOmega}^{(k)}$ is diagonal and strictly positive (so long as the
robust kernel $\rho$ produces nonnegative weights, as is standard), reweighting
rescales the entries of $\matM$ but creates no new fill and destroys no
existing nonzeros. Consequently, the symbolic Cholesky factorization of
$\matM^{(k)}$, including any fill-reducing ordering and the sparsity pattern of
$L^{(k)}$, depends only on the graph structure and is computed
once~\citep{doi:10.1137/1.9780898718881}.

\item
\textbf{Numerical values must be recomputed.} The numerical Cholesky factor
$L^{(k)}$, the cross term $B^{(k)} = \matAcT \widetilde{\matOmega}^{(k)} \matC$,
and the constrained block $\matQcc^{(k)} = \matAcT \widetilde{\matOmega}^{(k)} \matAc$
all depend on $\widetilde{\matOmega}^{(k)}$ and must be recomputed at each
outer iteration. Crucially, because their sparsity patterns are fixed, these
recomputations are performed in-place and are inexpensive relative to the full
preprocessing cost.\footnote{This is the standard setting for libraries that separate
symbolic and numerical Cholesky phases (e.g., CHOLMOD~\citep{chen2008algorithm}),
and is precisely the case in which they yield the largest amortized speedups.}
\end{enumerate}
The net effect is that
\cref{alg:schur-complement-product} splits cleanly into a one-time
\emph{symbolic} preprocessing step (computing $\matC$ and the symbolic
factorization of $\matC^\top \matOmega \matC$) and a per-outer-iteration
\emph{numerical} factorization, summarized in \cref{alg:irls-schur-complement-product}.

\input{alg/irls}



%% file: alg/irls.tex
\begin{algorithm}[t]
\caption{Matrix-Free Schur Complement Products with Robust Loss (IRLS)}
\label{alg:irls-schur-complement-product}
\begin{algorithmic}[1]
\Statex \textbf{Input:} Matrices $\matAc, \matAf, \matOmega$; robust kernel $\rho$.
\Statex \textbf{Output:} Operator $\matXc \mapsto \matQmarg^{(k)} \matXc$ at each IRLS iteration $k$.
\Statex
\Statex \textit{Symbolic preprocessing} (once, given $\matAc, \matAf, \matOmega$):
\State $\matC \leftarrow \mathrm{CR}(\matAf)$ \Comment{e.g., reduced incidence matrix}
\State $\mathcal{P} \leftarrow \mathrm{SymbolicCholesky}(\matC^\top \matOmega\, \matC)$ \Comment{Sparsity \& ordering}
\Statex
\Statex \textit{Per outer IRLS iteration $k$} (given current iterate $\matX^{(k)}$):
\State $w_i^{(k)} \leftarrow \rho'\!\big(\|r_i(\matX^{(k)})\|^2_{\matOmega_i}\big) \big/ \|r_i(\matX^{(k)})\|^2_{\matOmega_i}$ \Comment{IRLS weights, \cref{eq:irls-weights}}
\State $\widetilde{\matOmega}^{(k)} \leftarrow \text{blkdiag}(w_1^{(k)} \matOmega_1, \ldots, w_m^{(k)} \matOmega_m)$
\State $L^{(k)} \leftarrow \mathrm{NumericCholesky}(\matC^\top \widetilde{\matOmega}^{(k)} \matC,\; \mathcal{P})$ \Comment{Reuses symbolic factorization}
\State $B^{(k)} \leftarrow \matAcT \widetilde{\matOmega}^{(k)} \matC$ \Comment{Sparsity pattern fixed}
\State $\matQcc^{(k)} \leftarrow \matAcT \widetilde{\matOmega}^{(k)} \matAc$ \Comment{Sparsity pattern fixed}
\Statex
\Statex \textit{Per inner solve} (per new $\matXc$):
\State $Y \leftarrow (B^{(k)})^\top \matXc$
\State $Z \gets \big(L^{(k)}\big)^{-\top}\big(\big(L^{(k)}\big)^{-1} Y\big)$ \Comment{forward, then back substitution}
\State \Return $\matQmarg^{(k)} \matXc = \matQcc^{(k)} \matXc - B^{(k)} Z$
\end{algorithmic}
\end{algorithm}

%% file: sec/5_experiments.tex
\section{Experiments}\label{sec:cora-marg:computational-experiments}
We evaluated SPARSER through three complementary studies, each
designed to isolate a different aspect of its behavior to ensure a comprehensive assessment of its capabilities.
\begin{itemize}
    \item A controlled \emph{scalability sweep} on synthetic 3D PGO problems
to isolate how performance correlates to the total problem size.
Our method's runtime advantage over all baselines grows consistently
with problem size, reflecting both fewer iterations and a lower
per-iteration cost.
    \item A \emph{standard-benchmark evaluation} across PGO, RA-SLAM, SNL, and
SfM tests whether the synthetic trends carry over to established
datasets with realistic noise conditions.
These trends from the previous experiment hold, where our method achieves the fastest
runtime on 41 of the 42 CPU datasets and 35 of the 42 GPU datasets.
    \item Finally, a \emph{robust optimization} evaluation on CosmoBench and
Nebula multi-robot SLAM data tests the IRLS extension of
\cref{sec:robust-loss} under naturally outlier-corrupted loop closures.
This is the setting where we see if relaxing the core requirement of linear residuals intelligently still enables performance boosts.
The gains persist, with our robust variant achieving the fastest runtime
on all 24 datasets. 
\end{itemize}
All experiments use residuals from \cref{tab:quadratic-factors}, where
our methodology fully exploits the problem structure: cost residuals
are homogeneous linear functions (\cref{sec:problem-formulation}) and
the Jacobian block $\matAf$ corresponding to unconstrained variables is
a directed incidence matrix
(\cref{sec:prelim:efficient-schur-complement-products}).

\subsection{Implementation and Baselines}

\textbf{Our approach.}
We implement our CPU approach within a Riemannian trust-region (RTR)
framework~\citep{absil2007trust} using preconditioned truncated conjugate
gradients (pTCG)~\cite[Ch. 6.5]{boumal2023introduction} for trust-region
subproblems. We use the matrix-free Schur complement product algorithm
(\cref{sec:prelim:efficient-schur-complement-products}) to efficiently compute
matrix products, with Riemannian gradients and Hessian-vector products derived
via standard conversions from their Euclidean counterparts
\citep{boumal2023introduction}.
We precondition pTCG with the \emph{regularized Cholesky preconditioner}
\cite[Sec. VI.A]{papalia2024Certifiably} of $\matQ$: $P = L^{\shortminus \top}
L^{\shortminus 1} = (\matQ + \mu I)^{\shortminus 1}$, where $\mu$ is chosen so
the condition number of $P$ is below $10^6$. The preconditioner is stored as a
sparse Cholesky factor $L L^{\top} = (\matQ + \mu I)$ computed once at
initialization. Since the reduced variable $\matXc$ does not match the
preconditioner dimensions, we bottom-pad with zeros ($\matX = [\matXc; 0]$).
With respect to $\matXc$, this is equivalent to preconditioning with the Schur
complement of $(\matQ + \mu I)$, making it a principled preconditioner. We implement the GPU solver following the GPU Riemannian trust-region scheme
of~\citep{han2025Building}, running the same algorithm and the same
preconditioner as above and differing only in that the optimization occurs only on the GPU.
\begin{figure}[t]
    \centering
    \includegraphics[width=0.30\textwidth]{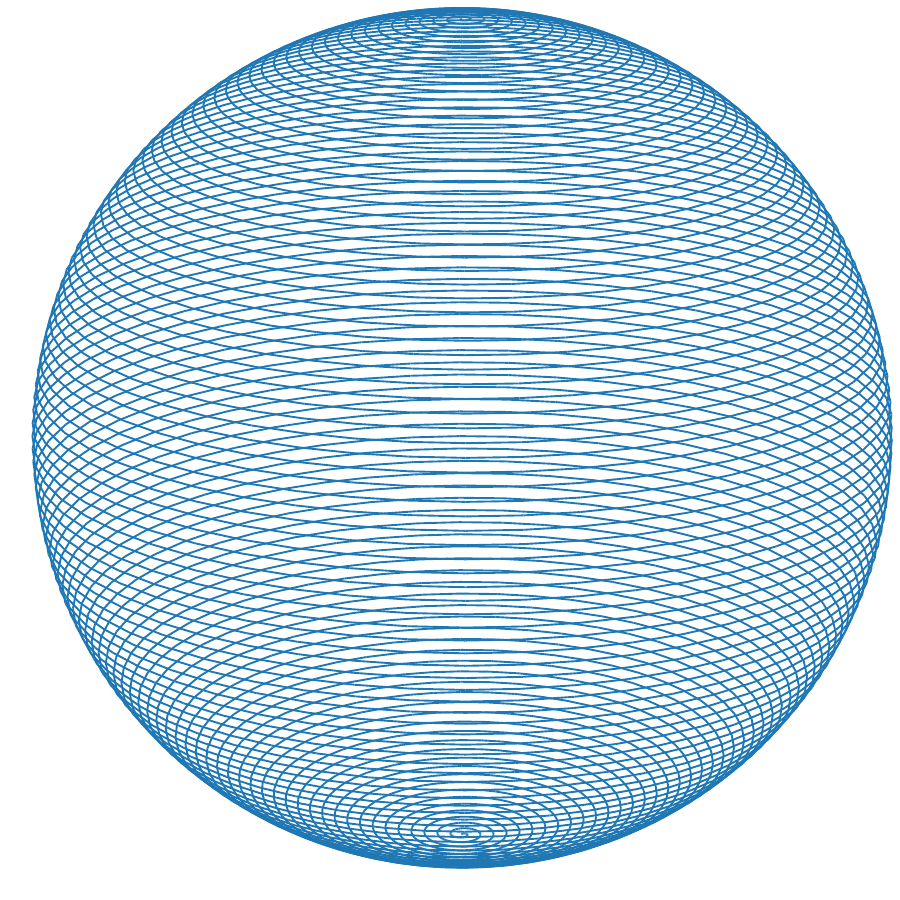}
    \caption{\textbf{Synthetically Generated Sphere of 5k variables.} An example of the dataset we generate for our scalability sweep with the 5k variables at their ground truth postions.}
    \label{fig:robust_converge}
\end{figure}

\textbf{Baselines.}
We benchmark against four baselines: (i) \emph{Original}, which directly
optimizes the full problem without variable elimination; (ii) \emph{Original +
VarPro}~\citep{khosoussi2016Sparse}, which optimizes the same problem as
\emph{Original} but uses variable projection to update unconstrained variables
in closed form at each iteration; (iii) \emph{Dense} which is the dense Schur complement~\citep{zhang2006schur} directly computing and instantiating the matrix; and (iv) we compare against \emph{GTSAM}~\citep{dellaert2012factor}, a state-of-the-art solver using direct factorization-based linear solvers\footnote{GTSAM also has the capability to use iterative solvers to solve the subproblems, analogous to our implementation.}. The GTSAM baseline serves two purposes. First, it provides an external reference point for runtime as GTSAM is a mature, highly optimized library, both in its implementation and in its algorithmic choices, and these system-level optimizations yield speedups that may not admit a one-to-one algorithmic comparison. Second, it contrasts two distinct strategies for the inner subproblem because GTSAM employs a direct solve for the inner problem unlike our iterative pTCG. Direct linear solvers are inherently robust to the conditioning issues that necessitate preconditioning in our PCG-based inner solve, which allows libraries like GTSAM to achieve fast, reliable solves without a preconditioning step. However, direct solvers cannot exploit the sparse structure of our Schur complement. The Levenberg-Marquardt is configured with a relative error tolerances of $10^{\shortminus 7}$ to ensure no early termination;
all other LM hyperparameters are kept at the defaults.
\begin{table}[t]
    \centering
    \caption{Precompute timing as the problem grows. We show our methods required precompute versus  the runtime it takes to form the dense Schur-complement on the synthetic 3D
    PGO benchmark. Our formulation of the problem keeps the precompute computational cost low.The alternative with forming the Dense Schur quickly becomes expensive.}
    \label{tab:precompute_vs_dense}
    \begin{tabular}{r r r r}
        \toprule
        Variables & Ours (ms) & Dense Schur (ms) & Speedup \\
        \midrule
        1k  &  1.50 &   28.3 &  19$\times$ \\
        2k  &  3.27 &   64.5 &  20$\times$ \\
        3k  &  5.45 &  184.7 &  34$\times$ \\
        4k  &  8.76 &  436.7 &  50$\times$ \\
        5k  & 10.16 &  569.5 &  56$\times$ \\
        6k  &  9.88 &  722.6 &  73$\times$ \\
        7k  & 11.09 & 1159.5 & 105$\times$ \\
        8k  & 13.87 & 1456.1 & 105$\times$ \\
        9k  & 23.39 & 2098.0 &  90$\times$ \\
        10k & 16.34 & 2613.1 & 160$\times$ \\
        \bottomrule
    \end{tabular}
\end{table}

The first three baselines (\emph{Original}, \emph{Original + VarPro}, and \emph{Dense}) are implemented as alternative options within the same RTR framework as our approach; \emph{Dense} is evaluated only with the GPU solver, as the size of these matrices makes a CPU solve infeasible. Our GTSAM implementations use custom manifolds and residuals from \cite{xu2026certifiableestimationfactorgraphs} to ensure identical problem formulations. For the robust optimization experiment, \emph{Original}, \emph{Original + VarPro}, \emph{Dense}, and Ours are wrapped in our own IRLS-GNC loop (\cref{alg:irls-schur-complement-product}), while GTSAM uses its native graduated non-convexity (GNC) implementation~\citep{yang2020Graduated}. These methods follow the same procedures except in the solver type used. All robust solves use the Geman--McClure kernel with squared inlier threshold $c^{2} = 25$, initial multiplier $\mu_{0} = 64$, and per-iteration shrink factor $1.4$. These values were selected and tuned empirically to give reliable convergence and fast runtime across all methods.

\begin{figure*}[t]
    \centering
    \begin{subfigure}[t]{0.495\textwidth}
        \centering
        \includegraphics[trim={0 0 {0.5} 0}, clip, width=\textwidth]{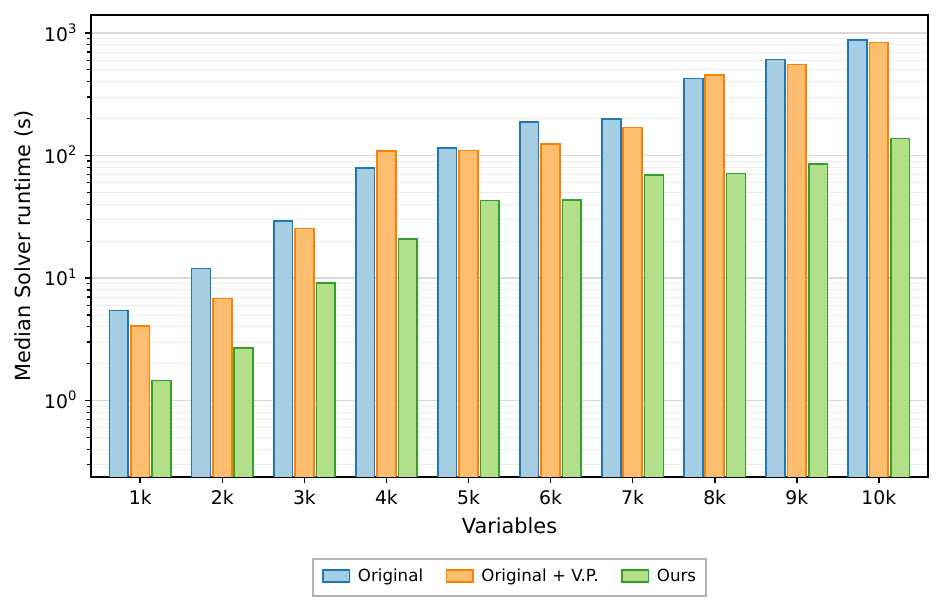}
        \caption{Wall-clock runtime until convergence on CPU.}
        \label{fig:sweep_runtime_cpu}
    \end{subfigure}
    \hfill
    \begin{subfigure}[t]{0.495\textwidth}
        \centering
        \includegraphics[width=\textwidth]{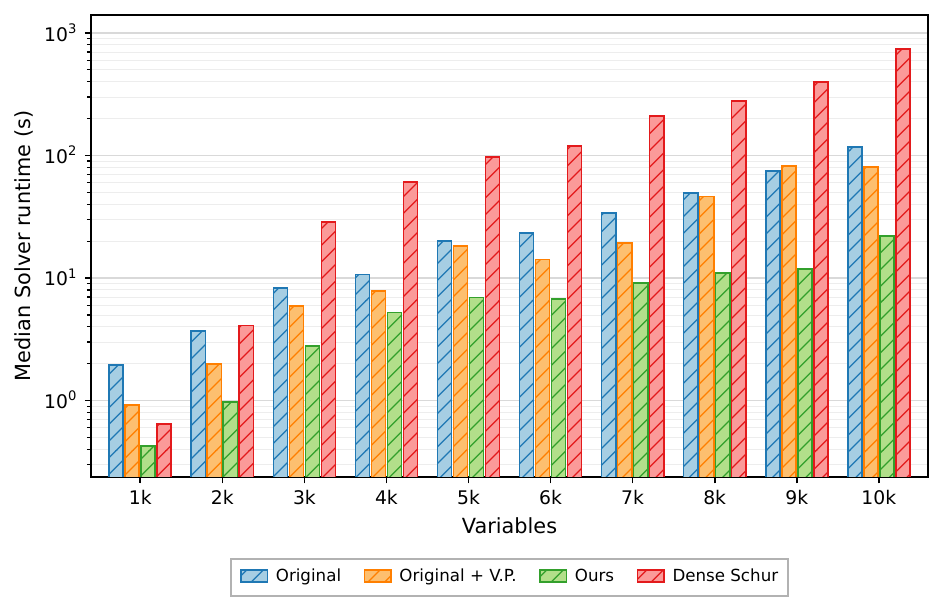}
        \caption{Wall-clock runtime until convergence on GPU.}
        \label{fig:sweep_runtime_gpu}
    \end{subfigure}
    \vspace{1em}
    \begin{subfigure}[t]{0.495\textwidth}
        \centering
        \includegraphics[width=\textwidth]{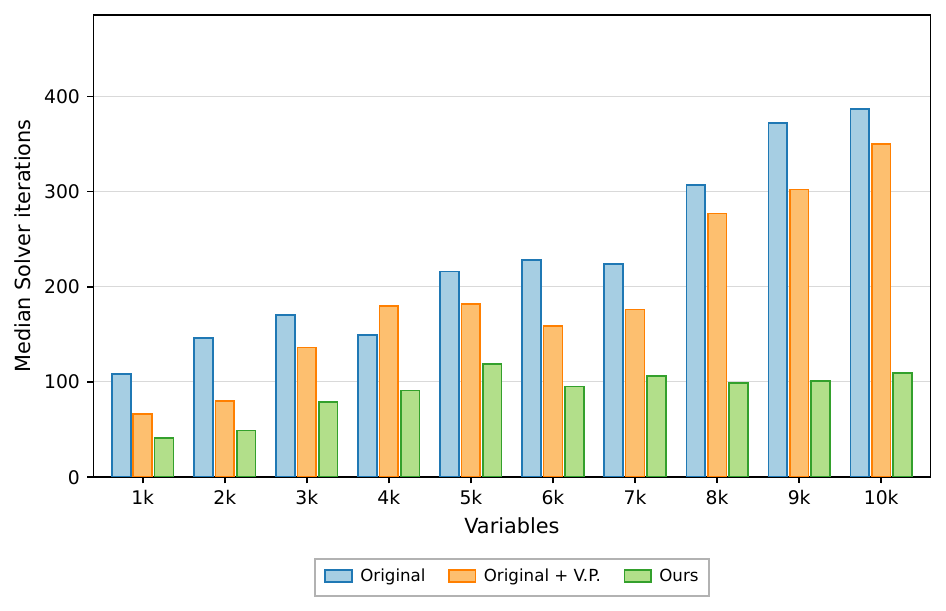}
        \caption{Iterations until convergence on CPU.}
        \label{fig:sweep_iterations_cpu}
    \end{subfigure}
    \hfill
    \begin{subfigure}[t]{0.495\textwidth}
        \centering
        \includegraphics[width=\textwidth]{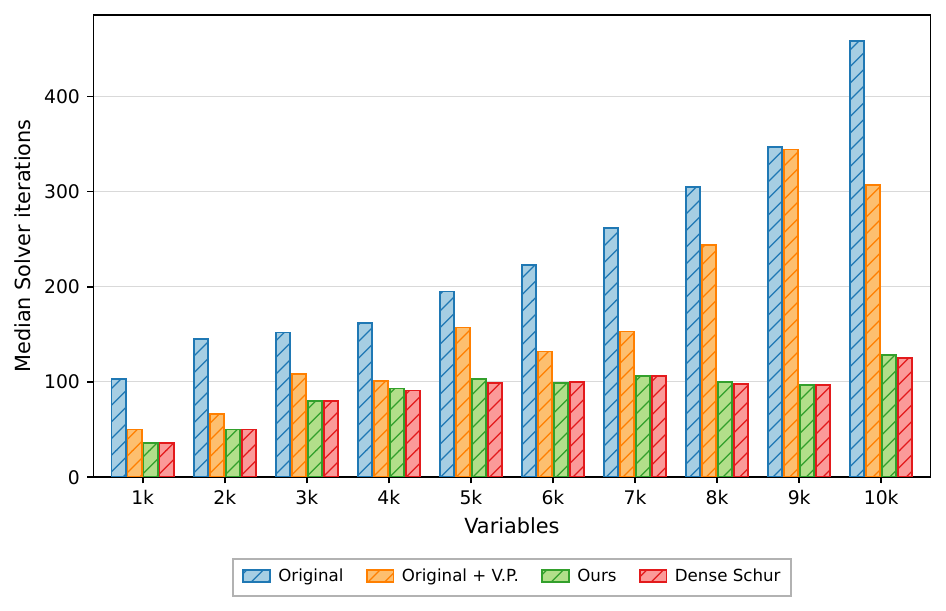}
        \caption{Iterations until convergence on GPU}
        \label{fig:sweep_iterations_gpu}
    \end{subfigure}
    \vspace{1em}
    \begin{subfigure}[t]{0.495\textwidth}
        \centering
        \includegraphics[width=\textwidth]{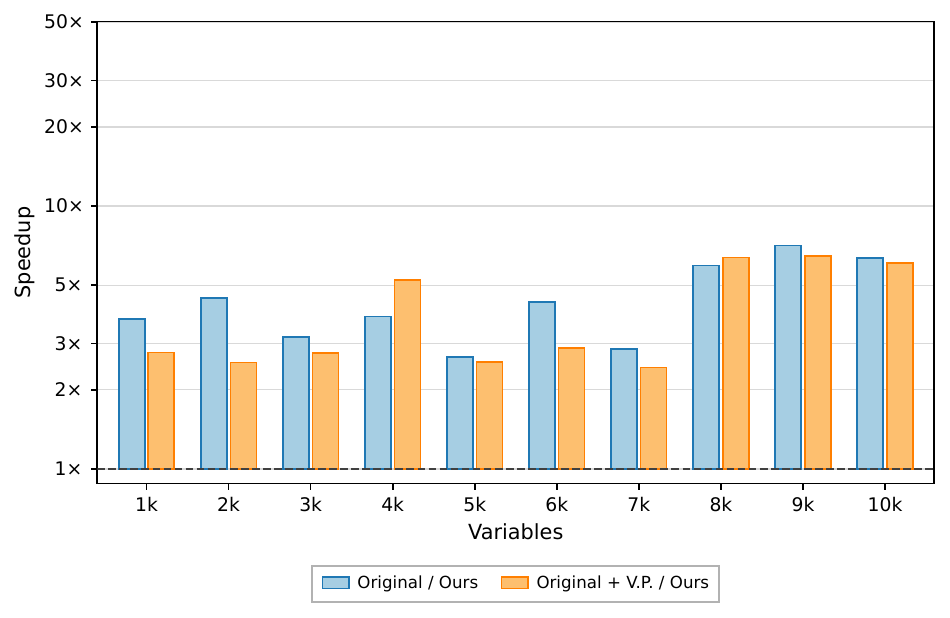}
        \caption{Our method's speedup compared to baselines on CPU.}
        \label{fig:sweep_speedup_cpu}
    \end{subfigure}
    \hfill
    \begin{subfigure}[t]{0.495\textwidth}
        \centering
        \includegraphics[width=\textwidth]{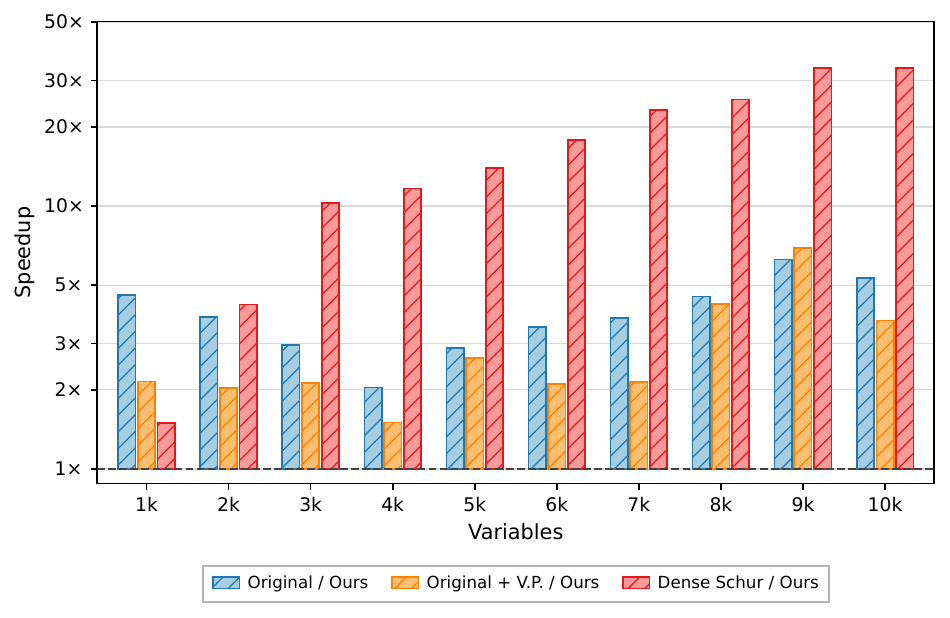}
        \caption{Our method's speedup compared to baselines on GPU.}
        \label{fig:sweep_speedup_gpu}
    \end{subfigure}
    \vspace{1em}
    \caption{\textbf{Scalability of the solvers on the synthetic 3D PGO
benchmark.} Median solver performance as the problem grows from 1k to
10k variables (a 1:1 constrained-to-unconstrained ratio), with medians
taken over 25 random initializations per problem size. The left column
(a, c, e) reports CPU results comparing \textit{Original},
\textit{Original + VarPro}, and \textit{Ours}; the right column
(b, d, f) reports GPU results, additionally including the
\textit{Dense Schur} baseline, which is evaluated only on the GPU
because forming the dense complement is infeasible on CPU at this
scale. \textbf{Top row (a, b):} median wall-clock runtime (log scale).
\textbf{Middle row (c, d):} median solver iterations. \textbf{Bottom row (e, f):} speedup of \textit{Ours} relative to each
baseline, computed as $\text{Baseline}/\text{Ours}$, so values above
$1\times$ favor \textit{Ours}. Every ratio exceeds $1\times$ at these ten
problem sizes on both CPU and GPU, and on GPU the speedup over
\textit{Dense Schur} grows from about $1.5\times$ at 1k variables to over
$30\times$ at 10k.}
    \label{fig:spheresweep}
\end{figure*}

\textbf{Hardware and software.}
Experiments were conducted on a laptop running Ubuntu 22.04 with an
Intel Core Ultra 9 275HX (24 cores), 64\,GB of RAM, and an NVIDIA GeForce RTX 5090
Laptop GPU with 24\,GB of VRAM. We utilize the CUDA 12.6 toolkit for our GPU implementation. All implementations are available as
open-source C++ libraries along with data and scripts to reproduce the
experiments.

\begin{figure*}[htbp]
    \centering
    \begin{subfigure}[t]{0.48\textwidth}
        \centering
        \includegraphics[width=\linewidth]{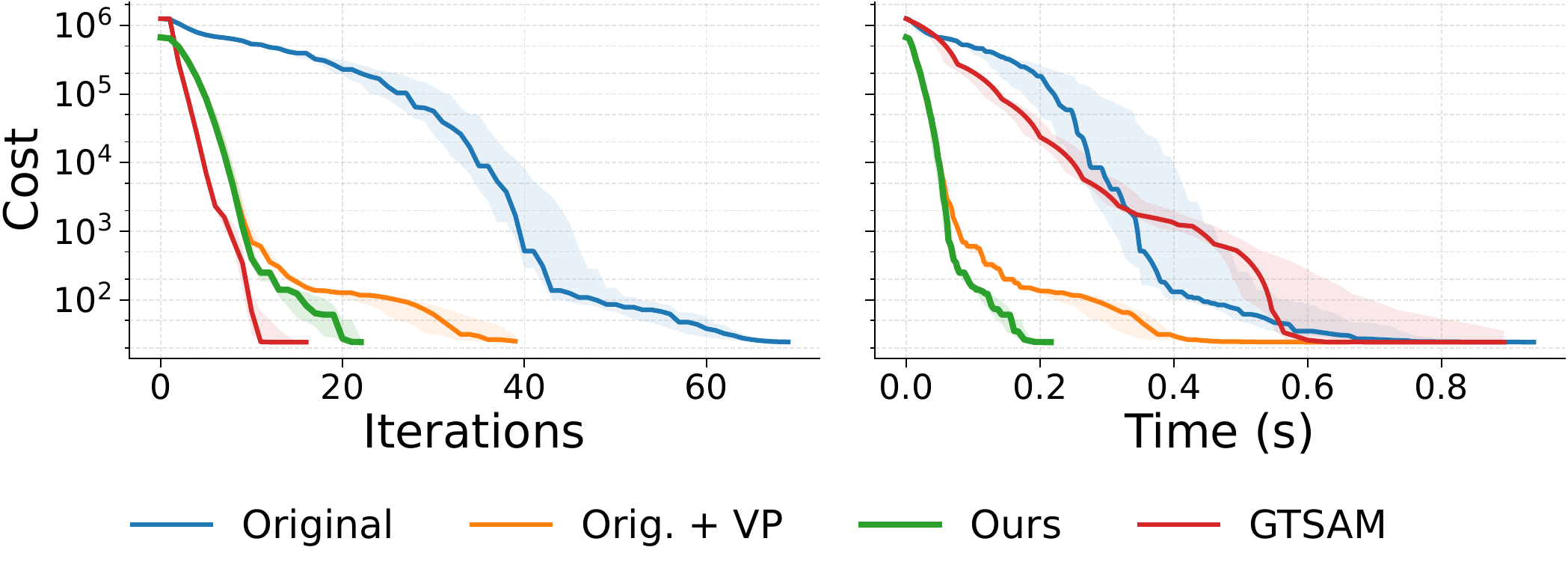}
        \caption{Pose-graph optimization (Intel)}
        \label{fig:sub1}
    \end{subfigure}
    \hfill
    \begin{subfigure}[t]{0.48\textwidth}
        \centering
        \includegraphics[width=\linewidth]{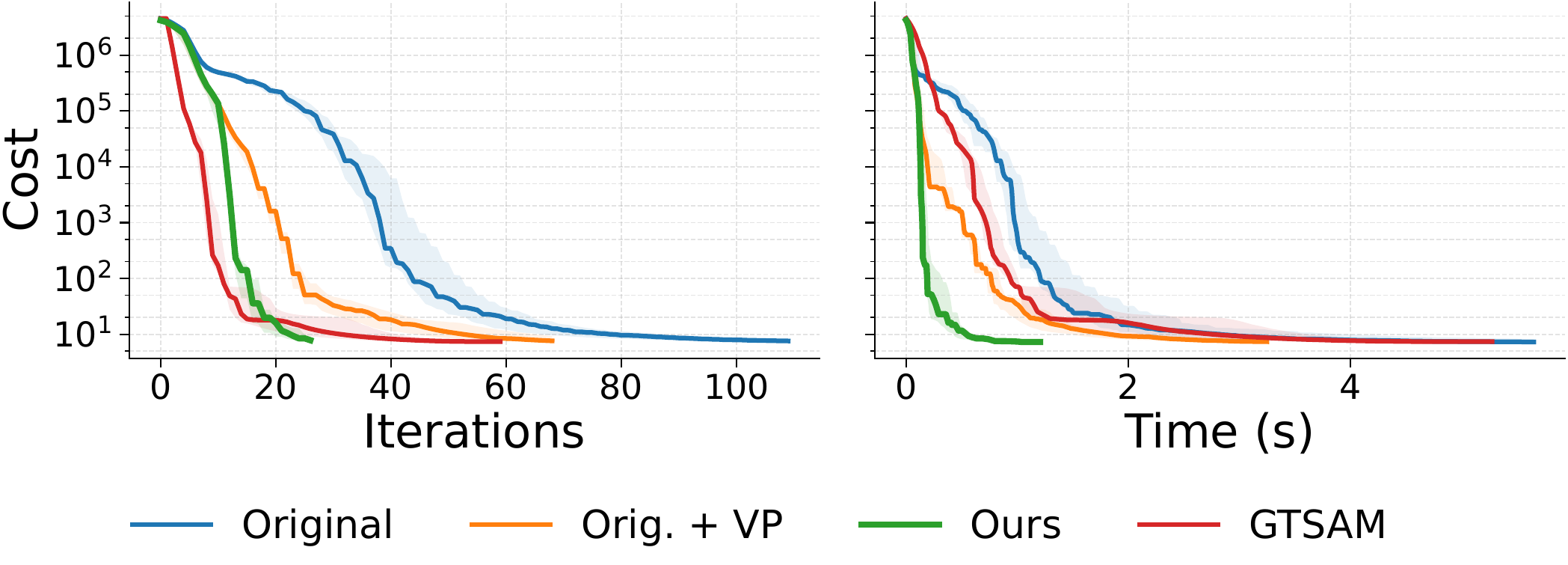}
        \caption{Range-aided SLAM (Single Drone)}
        \label{fig:sub2}
    \end{subfigure}
    \vskip\baselineskip
    \begin{subfigure}[t]{0.48\textwidth}
        \centering
        \includegraphics[width=\linewidth]{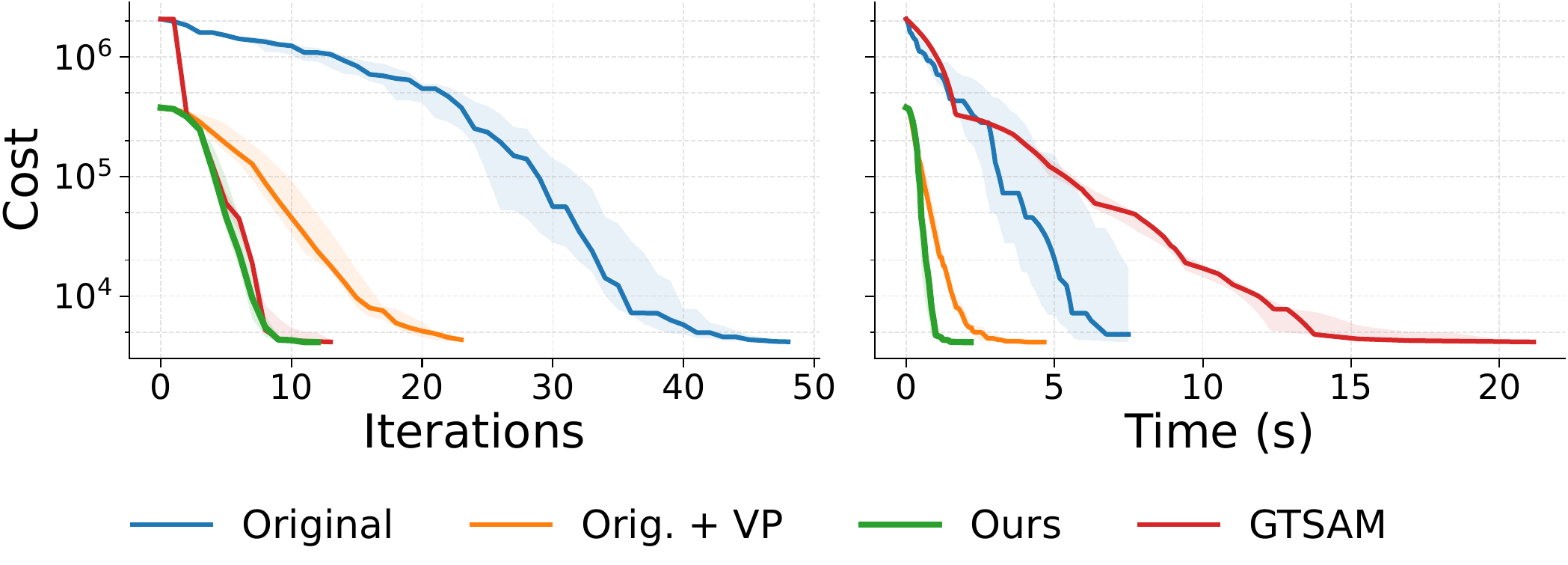}
        \caption{Structure from motion (Mip-NeRF Garden)}
        \label{fig:sub3}
    \end{subfigure}
    \hfill
    \begin{subfigure}[t]{0.48\textwidth}
        \centering
        \includegraphics[width=\linewidth]{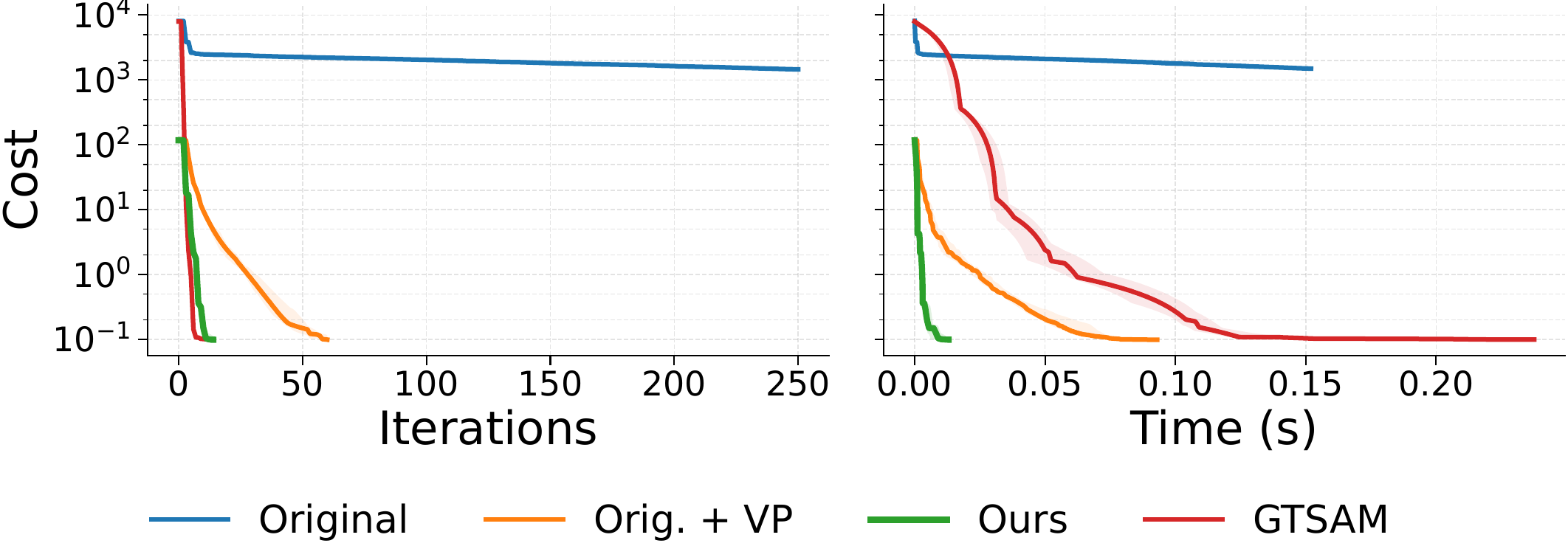}
        \caption{Sensor network localization (MIT)}
        \label{fig:sub4}
    \end{subfigure}
    \caption{\textbf{Convergence behavior on select problems.} Each panel pair
    shows (left) cost vs. iterations and (right) cost vs. time for our method
    and the baseline methodologies on representative datasets chosen from
    (a) pose-graph optimization (Intel), (b) range-aided SLAM (Single Drone),
    (c) structure from motion (Mip-NeRF Garden), and (d) sensor network
    localization (MIT). Solid lines show the median over five random
    initializations, and shaded regions show the range of different convergence paths. In (d), \textit{Original} does not converge, and its line ends
    where the run reached the iteration limit. In all four cases shown,
    SPARSER (\textit{Ours}) is the most efficient in wall-clock runtime.}
    \label{fig:main}
\end{figure*}

\input{tab/tab_experiments}

\subsection{Experiments}

We evaluate our method through three complementary experiments. The first is a
controlled \emph{scalability sweep} on synthetically generated 3D dataset, isolating
how performance depends on total problem size. The second is a \emph{standard-benchmark
evaluation} across PGO, RA-SLAM, SNL, and SfM on well-established datasets
without outliers.\footnote{Landmarks measurements are present in MRCLAM and outfinite RA-SLAM datasets. While not explicitly noted through the conventional definitions of PGO and RA-SLAM the problem formulations do include this.} The third is a \emph{robust optimization} evaluation on
CosmoBench~\citep{mcgann2025cosmobenchbenchmarkcollaborativeslam} and nebula~\citep{chang2022lamp}, real-world
multi-robot SLAM datasets with naturally outlier-corrupted loop closures.

\textbf{PGO scalability sweep.}
We synthetically generate spherical PGO problems using the
\texttt{create\_sphere} generator of g2o~\citep{kummerle2011G2o}, the same
generator behind the standard \emph{sphere2500} dataset. The trajectory
spirals from the south pole of a sphere of radius $100\,$m to the north pole in
circumferential laps of $50$ poses, so problem size is set purely by the number
of laps. The constrained variables are the pose rotations and the unconstrained
variables are the pose translations, a constrained-to-unconstrained ratio of
$1{:}1$. Measurements consist of sequential odometry between consecutive poses,
together with loop closures between consecutive laps: each pose on lap $f-1$ is
connected to the three adjacent poses on lap $f$, clamped at the lap
boundaries, giving a loop closure rate of $\approx\!2.9$ per pose (at $10$k
poses, $29{,}403$ loop closures against $9{,}999$ odometry edges). We use the
g2o default noise model, with translational noise $\sigma_t = 0.01\,$m per axis
and rotational noise $\sigma_R = 0.005\,$rad per axis sampled in g2o's
quaternion parameterization. The sweep varies the number of poses
in $\{1\text{k}, 2\text{k},3\text{k},4\text{k},5\text{k},6\text{k},7\text{k},8\text{k},9\text{k}, 10\text{k}\}$. We use a rank-$5$
Burer--Monteiro relaxation to ensure a benign landscape~\citep{mcrae2024benign},
which for PGO has been shown to empirically obtain globally optimal
solutions~\citep{rosen2019SESync}. We evaluate four formulations (\emph{Original}, \emph{Original + VarPro}, \emph{Dense}, and Ours). On CPU, we evaluate \emph{Original}, \emph{Original + VarPro}, and Ours; on GPU, we evaluate all four formulations. We report medians over $25$ random initializations per scenario, with each run executed to convergence without wall-clock or iteration limits.

\textbf{Standard benchmark datasets.}
The PGO datasets are from~\cite{carlone2015Initialization}. The RA-SLAM
datasets except \emph{outfinite} are from~\cite{papalia2024Certifiably}. We collected the \emph{outfinite} dataset because large-scale, multi-agent, range-aided datasets remain limited compared with existing benchmarks~\citep{leung2011utias, djugash2009plaza, yu2023fusingodometryuwbranging}. The dataset comprises three agents traversing approximately 6~km and generating roughly 13.5k poses. Six fixed landmarks provide approximately 7.6k range measurements. The scale of \emph{outfinite}, together with its multi-agent and range-aided structure, presents a significant challenge for existing solvers and provides a demanding benchmark for demonstrating the capabilities of our proposed method.  SNL datasets are generated
synthetically from the PGO datasets by converting all poses to points and all
measurements to range measurements with identical noise levels. SfM datasets
are generated according to~\cite[Sec. IV]{han2025Building} with scale set to $1$. All dataset names
match original sources.
We use problem formulations from \cite{rosen2019SESync} (PGO),
\cite{papalia2024Certifiably} (RA-SLAM), \cite{halsted22arxiv} (SNL), and
\cite{han2025Building} (SfM). Like the PGO sweep, all these  formulations have previously empirically been found to possess benign
optimization landscapes~\citep{mcrae2024benign,criscitiello2025sensor}\footnote{We also verified that each dataset, from our random initializations, the solver could reach the globally optimal cost value.},
meaning local optimization methods can reliably reach global minima from
random initializations and thus we use a rank-$5$ Burer--Monteiro relaxation. Since all solvers obtain the global minimum, our
analysis avoids complications from local minima.
For each dataset we generate $5$ random initializations and run each method
on all $5$ trials, with solvers running until convergence or $300$ seconds
elapsed.

\textbf{Robust optimization on real-world SLAM data with outliers.}
CosmoBench~\citep{mcgann2025cosmobenchbenchmarkcollaborativeslam} is a
real-world multi-robot SLAM benchmark with pose-graph data across the \texttt{kth},
\texttt{ntu}, \texttt{kittredge\_loop}, and \texttt{main\_campus} scenes,
yielding 20 datasets. These datasets contain outliers that have naturally occurred from real-world collection. We additionally include
the four Nebula~\citep{chang2022lamp} multi-robot datasets, which follow
the same data conventions. All solvers are initialized from the same odometry chain starting at the origin,
with noise added by zero-mean Gaussian on the odometry measurements as described
below, and solved at the ambient problem dimension (3D), which provides a reasonable starting point within the basin of
attraction of an accurate trajectory and ensures a fair runtime comparison
across methods.

To handle outliers, we wrap each solver in iteratively reweighted least
squares (IRLS) with the Geman--McClure (GM) kernel, embedded in a graduated
non-convexity (GNC) schedule. Unlike the
outlier-free setting, certifiably correct and globally optimal solutions
under outliers remain an
open problem~\citep{xu2026implementingrobustmestimatorscertifiable}. GNC mitigates this by starting from a convex approximation of the GM cost and
gradually increasing its nonconvexity until the original GM cost is recovered,
making IRLS substantially less sensitive to initialization. For our method and the \emph{Original} /
\emph{Original + VarPro} baselines, this procedure corresponds to
\cref{alg:irls-schur-complement-product}, while the GTSAM baseline uses its
native GNC implementation. We run 5 trials per dataset for each method initialized from odometry with given noise of $0.5^\circ$ standard deviation on each axis-angle rotation component and $0.02\,$m on each translation component, applied independently to every sequential odometry measurement before composition, so that the perturbations accumulate along each robot's trajectory.

We report trajectory accuracy via the absolute trajectory error (ATE),
\begin{equation}
    \mathrm{ATE}_{\mathrm{mean}}
    = \frac{1}{n} \sum_{i=1}^{n} \lVert \mathrm{trans}(F_i) \rVert,
    \label{eq:ate}
\end{equation}
where $F_i = Q_i^{-1}\, S\, P_i$ is the per-frame pose error after global
alignment~\citep{Umeyama1991least} of the estimated trajectory
$\{P_i\}$ to the outlier-free ground truth $\{Q_i\}$. We additionally
report convergence cost as wall-clock time to convergence and total solver
iterations.

\textbf{Convergence criterion and failure conventions.}
For the sweep and the standard-benchmark experiments, a solver is considered
converged if it reaches within $1\%$ of the globally optimal minimum cost. For the robust optimization experiment, convergence is defined
by the GNC schedule itself, with a solver considered converged when the GNC
continuation parameter is fully annealed and the inner least-squares
subproblem has reached its tolerance. To ensure our robust experiment is a fair comparison of runtime we only report if the formulation can achieve with 5\% of the best ATE result.  All reported times, iteration counts, and ATE values in
\cref{fig:sweep_runtime_cpu,fig:sweep_runtime_gpu,fig:sweep_iterations_cpu,fig:sweep_iterations_gpu,fig:main,fig:robust_paetro,tab:pgo_wide_gtsam_compare,tab:gpu_results,tab:cosmobench_irls_gnc}
are medians across trials. Solver failures due to non-convergence (e.g. not
reaching the global optimum/ATE result within the time or iteration limit) and memory exhaustion
are indicated by (\mredx) and (\xmark), respectively. In all instances where
a solver failed in a trial with multiple initializations, it was found to converge in none of the trials.

\subsection{Results}

\textbf{Scalability sweep.} On the synthetic 3D PGO benchmark
(Fig.~\ref{fig:spheresweep}), the runtime advantage of \textit{Ours}
over all baselines grows consistently with problem scale on both
backends (Fig.~\ref{fig:sweep_runtime_cpu},
Fig.~\ref{fig:sweep_runtime_gpu}). On CPU, \textit{Ours} runs in about
1.5\,s at 1k variables and about 140\,s at 10k, while \textit{Original}
reaches roughly 900\,s at 10k. The margin over \textit{Original} is the
largest, as \textit{Original} performs no variable elimination, while
the margin over \textit{Original + VarPro} is smaller since that
baseline already eliminates the unconstrained variables in closed form
at each iteration and our remaining savings come from reusing that
elimination through the matrix-free operator rather than recomputing it
each step. This shows in the iteration counts
(Fig.~\ref{fig:sweep_iterations_cpu},
Fig.~\ref{fig:sweep_iterations_gpu}), where the median iterations of
\textit{Ours} stay roughly flat near 100 across the sweep while
\textit{Original} climbs from about 108 to nearly 400, so our runtime
gains reflect both a lower per-iteration cost and fewer iterations. The
GPU speedups (Fig.~\ref{fig:sweep_speedup_gpu}) are consistently
largest for \textit{Ours}, reflecting a per-iteration kernel of sparse
matrix products and triangular solves that maps well to GPU hardware.

The \textit{Dense} baseline isolates the value of our reformulation.
\textit{Dense} and \textit{Ours} solve the same reduced problem and
differ only in how the Schur complement product is formed, so they
record nearly identical iteration counts at every size
(Fig.~\ref{fig:sweep_iterations_gpu}). Both therefore inherit the
improved conditioning of the reduced problem over the original, which
is visible in the iteration gap between \textit{Ours} and
\textit{Original}. The two methods diverge only in per-iteration cost.
Because this benchmark holds a 1:1 constrained-to-unconstrained variable ratio,
eliminating the pose translations only halves the problem, so
\textit{Dense} trades a sparse full-size system for only slightly smaller but dense one, whereas \textit{Ours} realizes the same
reduction through a matrix-free operator that preserves the original
sparsity. At the smallest sizes the reduced system is small in absolute
terms and \textit{Dense} is competitive, running close to \textit{Ours}
at 1k and about 1.5$\times$ slower. As the problem grows the dense
factorization dominates and its cost scales far worse than our sparse
operator, so on GPU the speedup of \textit{Ours} over \textit{Dense}
widens from about 1.5$\times$ at 1k to over 30$\times$ at
10k (Fig.~\ref{fig:sweep_speedup_gpu}). Since the iteration counts are
matched, this gap is attributable entirely to sparsity rather than to
any difference in convergence.

Table~\ref{tab:precompute_vs_dense} isolates the one-time preprocessing
our reported runtimes include. Our precompute stays small as the
problem grows, from 1.50\,ms at 1k variables to 16.34\,ms at 10k, while
forming the dense Schur complement grows from 28.3\,ms to 2613.1\,ms
over the same range. The resulting speedup widens with scale, from
19$\times$ at 1k to 160$\times$ at 10k, confirming that the matrix-free
construction remains efficient in preprocessing compared to the \emph{Dense} baseline.

\begin{figure}[t]
    \centering
    \includegraphics[width=0.95\linewidth]{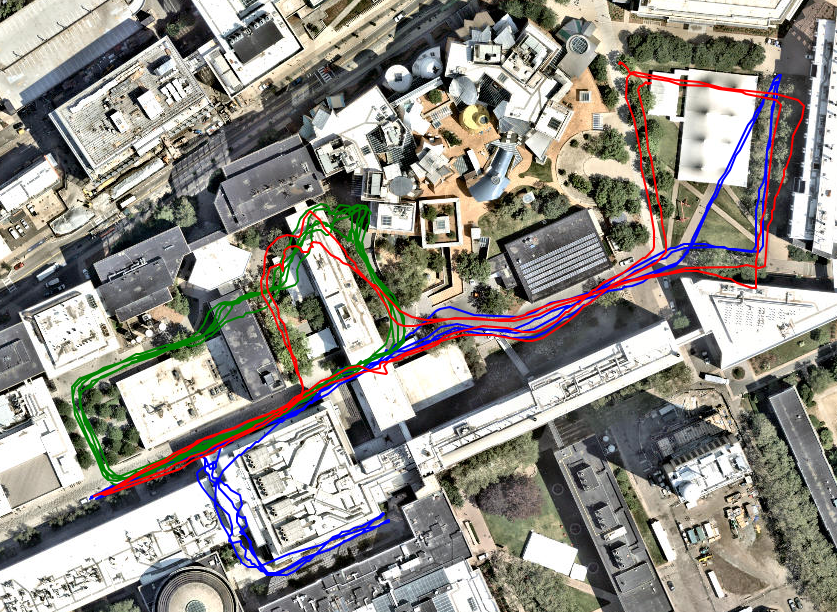}
    \caption{\textbf{The Outfinite dataset.} A large-scale, three-robot cooperative
range-aided SLAM problem collected from sensor
payloads traversing roughly 6\,km along the outdoor corridor on the MIT campus.
The problem comprises ${\sim}13.5$k poses tied by ${\sim}13.5$k odometry and
relative-pose constraints, together with ${\sim}7.6$k UWB range measurements to
six static beacons and between robots. The trajectories shown are the solution
computed by our method, overlaid on satellite imagery. On CPU, every baseline
fails to converge, and our method is the only approach that solves it
(\cref{tab:pgo_wide_gtsam_compare}). On GPU, only \textit{Dense Schur} also
solves it, roughly $70\times$ slower ($197.7$\,s versus $2.80$\,s;
\cref{tab:gpu_results}).}
    \label{fig:outfinite}
\end{figure}

\textbf{Standard benchmark datasets.}
On CPU (\cref{tab:pgo_wide_gtsam_compare}), \emph{Ours} achieves the fastest
runtime on $41$ of the $42$ datasets when compared against \emph{Original},
\emph{Original~+~VarPro}, and \emph{GTSAM}, gaining efficiency through both
lower per-iteration cost and reduced iteration counts. The single exception is
PGO \emph{Garage}, where \emph{GTSAM} is faster by a factor of $2.27$; this
relatively small, well-conditioned problem is well suited to \emph{GTSAM}'s
direct linear solvers. Our advantage is largest on SfM, where the reduced
problem is most favorable, with \emph{Ours} running $43.6\times$ faster than
\emph{Original} and $30.9\times$ faster than \emph{Original~+~VarPro} on
\emph{BAL-392} and comparable multiplicative gains across the Mip-NeRF and TUM
sequences.

Our analysis starts with the \emph{Original} and \emph{Original~+~VarPro} formulations as they share our RTR framework and
differ only in how they treat the unconstrained variables. \emph{Original} carries the full
variable set through the optimization and never eliminates, so it inherits the
worst conditioning and the largest per-iteration system. It consistently
requires the most iterations of any method (e.g., \emph{M3500}, $188$ against our
$31$) and is up to $44\times$ slower than \emph{Ours} across the benchmark. It
also fails outright on the hardest problems, converging on only one of the eight
CPU SNL datasets, where the absence of elimination leaves the problem too
ill-conditioned to solve within the time budget. \emph{Original~+~VarPro}
eliminates the unconstrained variables in closed form, and this conditioning
benefit alone lets it converge on six of those eight SNL datasets while sharply
reducing its iteration counts relative to \emph{Original}. The cost is that it
recomputes the elimination at every iteration rather than reusing it, so despite
iteration counts that are comparable to or only somewhat above \emph{Ours} it
remains up to $31\times$ slower (e.g., \emph{BAL-392}, $140.73$\,s against
$4.56$\,s).

\emph{GTSAM} with its direct solver
produces highly accurate steps and on several problems converges in very few
iterations (as few as $2$ on \emph{BAL-392}), but each step requires an
expensive factorization, so \emph{GTSAM} remains slower overall despite these
low iteration counts. In addition, that same direct factorization must materialize
dense intermediate factors, whose fill-in exhausts the $64$~GB memory budget on
the largest SfM scenes even when the underlying problem is reduced. \emph{GTSAM}
runs out of memory on \emph{BAL-1934}, all three \emph{IMC} scenes
(\emph{Gate}, \emph{Temple}, \emph{Rome}), and all four TUM sequences. Because
\emph{Ours} preserves the sparsity of the original problem and relies on
iterative solves (pTCG), it never forms these dense factors, its memory
footprint stays small, and it solves every dataset in the benchmark.

On GPU (\cref{tab:gpu_results}) we replace \emph{GTSAM} with the \emph{Dense}
Schur baseline, whose behavior most clearly isolates the value of preserving
sparsity. The same ordering among the framework baselines holds, with two
effects worth noting. \emph{Original} and \emph{Original~+~VarPro} both retain
dense subkernels that limit GPU utilization, so their gains from the accelerator
are smaller than ours. The acceleration does, however, bring
\emph{Original~+~VarPro} under the $300$\,s budget on the largest SfM scenes
where it timed out on CPU (\emph{BAL-1934}, \emph{IMC Gate}, \emph{IMC Rome}),
recovering convergence there even as it stays $12$ to $27\times$ slower than
\emph{Ours}. \emph{Dense}, by contrast, solves the same reduced problem as
\emph{Ours} and records essentially the same iteration counts on every dataset,
so it differs only in per-iteration cost, exactly as in the scalability sweep
(\cref{fig:spheresweep}). Its performance therefore tracks the
constrained-to-unconstrained ratio. In SfM the unconstrained landmarks vastly
outnumber the constrained rotations, so the Schur complement is small and the
dense matrix is a fair trade for a reduced problem. On the smaller scenes the
dense product even edges out our sparse operator (e.g., \emph{BAL-93},
$0.09$\,s against $0.11$\,s; \emph{BAL-392}, $0.90$\,s against $1.07$\,s;
\emph{Mip-NeRF Room}, $1.42$\,s against $2.03$\,s), and here the dense trade-off
is the correct one. As the SfM problems grow, the reduced block grows with them
and \emph{Ours} retakes the lead (e.g., \emph{BAL-1934},
\emph{IMC Gate/Temple/Rome}, \emph{TUM Comp-T}). Overall \emph{Ours} is fastest
on $35$ of the $42$ GPU datasets and tied on one more, the exceptions being
RA-SLAM \emph{Plaza2}, a near-tie with \emph{Original~+~VarPro}, and the five
SfM scenes claimed by \emph{Dense}.

The picture inverts on SLAM and SNL, where the constrained and unconstrained
blocks are comparable in number. The reduced
system never shrinks to a small dense block, so forming it explicitly is
expensive and \emph{Dense} is far slower than \emph{Ours}, by $60\times$ on
RA-SLAM \emph{TIERS} ($170.70$\,s against $2.82$\,s), $40\times$ on \emph{Plaza1}
($86.19$\,s against $2.17$\,s), and $32\times$ and $41\times$ on \emph{MR.CLAM6}
and \emph{MR.CLAM7}. On the larger \emph{MR.CLAM2} and \emph{MR.CLAM4} the dense
complement no longer fits in memory, so \emph{Dense} cannot even be run in that
configuration, whereas \emph{Ours} solves both from the same setup. Since the
iteration counts are matched, these gaps are attributable entirely to the dense
representation rather than to any difference in convergence.

Finally, we highlight \emph{Outfinite} (\cref{fig:outfinite}), our
self-collected three-robot dataset and one of the hardest problem in the benchmark. On
CPU no baseline converges and \emph{Ours} is the only solver to succeed
($9.29$\,s, $76$ iterations). On GPU the only other method that solves it is
\emph{Dense}, which is roughly $70\times$ slower ($197.70$\,s against
$2.80$\,s), while \emph{Original}, \emph{Original~+~VarPro}, and \emph{GTSAM}
all fail. By solving problems that leave every competing method either
non-convergent or out of memory, \emph{Ours} extends the scale and difficulty of
the environments in which reliable robotic perception is feasible.

\input{tab/robustcpu}

\begin{figure}[t]
    \centering
    \includegraphics[width=0.47\textwidth]{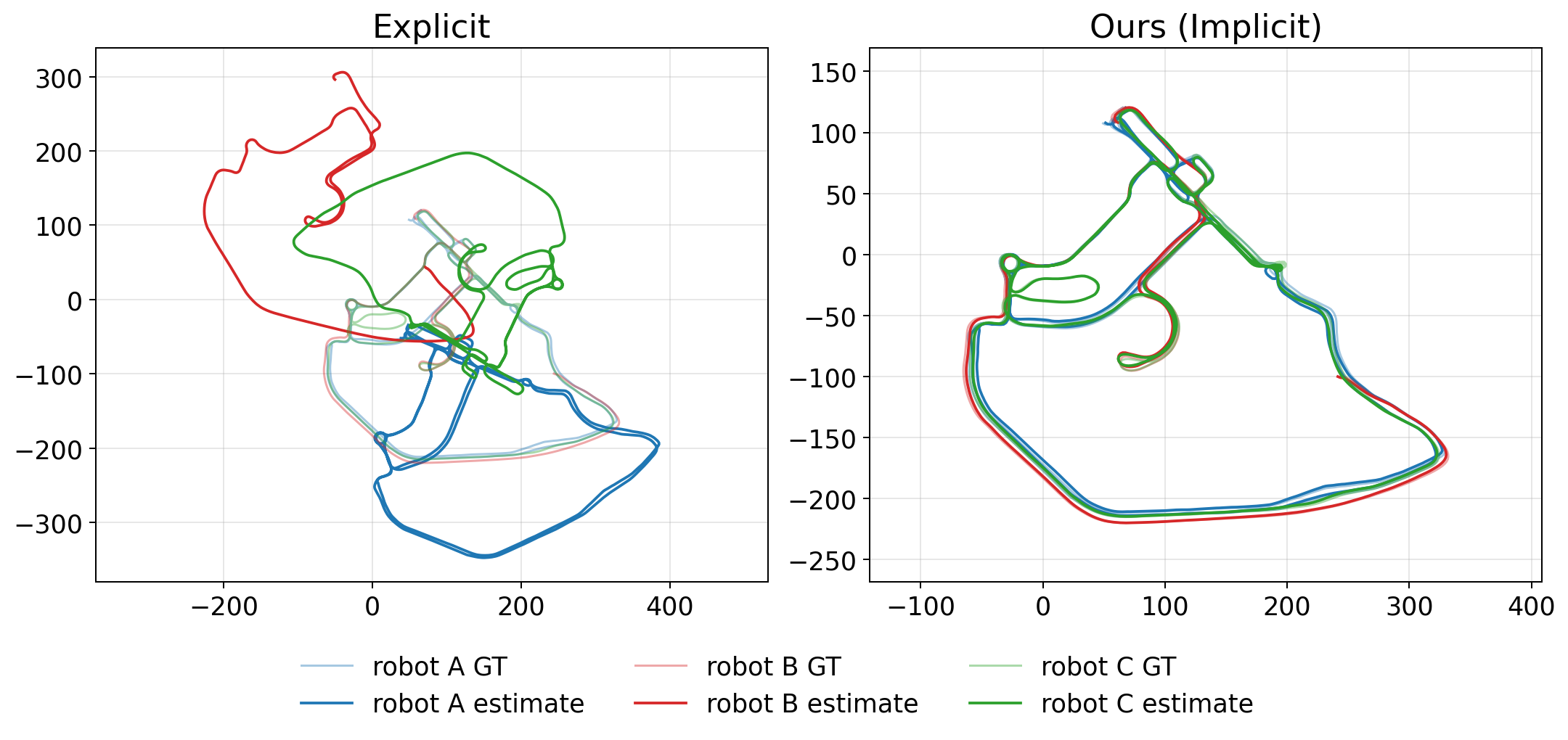}
    \caption{\textbf{Comparison of multi-robot trajectory estimates on the NTU WiFi sequence.} Three robots (A: blue, B: red, C: green) traverse overlapping regions; light-colored lines show ground truth and saturated lines show estimates. Left Original (Explicit) and Right Ours (Implicit).}
    \label{fig:trajectory_plot}
\end{figure}


\begin{figure}[t]
    \centering
    \includegraphics[width=0.47\textwidth]{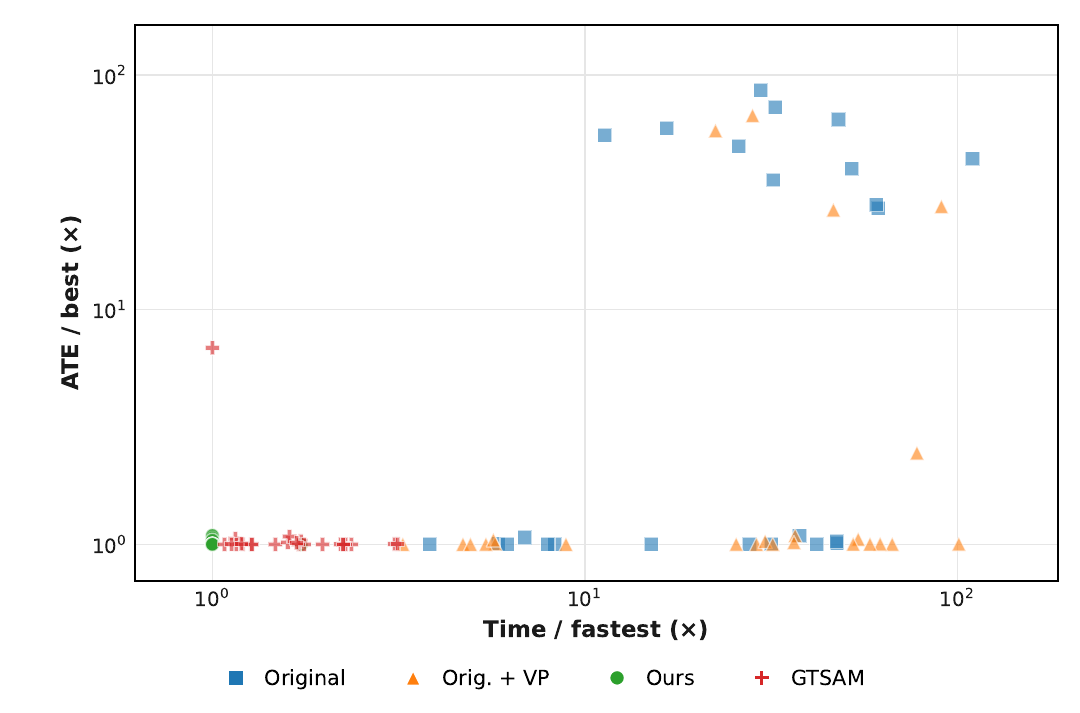}
    \caption{\textbf{Accuracy and timing on the robust CosmoBench and Nebula
benchmark.} Each point is one solver on one dataset, plotting trajectory error
against runtime after both are normalized per dataset. The $y$-axis is ATE
relative to the best ATE achieved on that dataset and the $x$-axis is runtime
relative to the fastest solver on that dataset, both on a log scale. The
bottom-left corner is ideal, denoting the most accurate
and fastest solve. \emph{Ours} clusters at this corner across the benchmark,
being the fastest solver on every dataset and at or near the best accuracy.
\emph{GTSAM-GNC} often matches our accuracy along the bottom edge but is
shifted right, running $1.1\times$ to $3.1\times$ slower, while \emph{Original}
and \emph{Original~+~VarPro} trail on both axes, spreading rightward in runtime
and upward into heavy ATE tails on the harder datasets. The concentration of our
points in the bottom-left corner means that SPARSER does not trade accuracy for
speed. On the datasets tested, it is the fastest solver while reaching an ATE at or
near the best achieved by any method.}
    \label{fig:robust_paetro}
\end{figure}

\textbf{Robust optimization on CosmoBench.}
Timing wise (\cref{tab:cosmobench_irls_gnc}), \emph{Ours} records the fastest
runtime on all $24$ datasets across CosmoBench and Nebula. It is faster than
\emph{Original} on every dataset where \emph{Original} converges, by up to
$47.5\times$ on the Wi-Fi \emph{KTH R3-01} scene, faster than
\emph{Original~+~VarPro} wherever that baseline converges, by up to $100.9\times$
on the Wi-Fi \emph{NTU R3-01} scene, and faster than \emph{GTSAM-GNC} on every
dataset, by factors from roughly $1.1\times$ to $3.1\times$. The robust setting
also separates the methods by whether they solve the problem at all.
\emph{Original}, which carries the full un-eliminated system and its poorer
conditioning, fails on $11$ of the $24$ datasets, and
\emph{Original~+~VarPro} on $5$, whereas \emph{GTSAM-GNC} fails on $1$ and
\emph{Ours} succeeds on all $24$. On Nebula \emph{Kentucky Underground},
\emph{Ours} is the only method that succeeds at all, mirroring the
\emph{Outfinite} result in the outlier-free benchmark.

\cref{fig:robust_paetro} combines both axes of this comparison, plotting each
run's accuracy (ATE relative to the best achieved on that dataset) against its
runtime (relative to the fastest on that dataset), so the ideal solver sits at
the bottom-left corner. \emph{Ours} clusters tightly there, at or near the best
ATE and the fastest time on essentially every dataset. \emph{GTSAM-GNC} lies
along the accurate bottom edge but is shifted to the right, matching our accuracy
while running several times slower. \emph{Original} and \emph{Original~+~VarPro}
trail on both axes, spreading rightward in runtime and upward into heavy ATE
tails on the harder datasets. The figure therefore shows our method is the single dominant
method; both accurate and efficient.

The comparison against \emph{GTSAM-GNC} isolates why our runtime advantage holds
even against an accurate solver. Its direct linear solver produces accurate
steps (as discussed from the standard benchmark experiments), and its iteration count is comparable to ours, lower on some datasets and
higher on others (e.g., $95$ against our $101$ on Nebula \emph{Finals}, but $448$
against $123$ on Wi-Fi \emph{NTU R5-00}). Despite this, each of its iterations
carries the cost of a full factorization, so it is slower than \emph{Ours} in
wall-clock on every dataset. Nebula \emph{Finals} makes the point cleanly, where
\emph{GTSAM-GNC} uses fewer iterations than \emph{Ours} yet is $1.7\times$
slower. This effect is amplified by the robust schedule. Each dataset is solved
with graduated non-convexity wrapped around IRLS, so a single robust solve
triggers many inner least-squares solves rather than one, thus a more efficient inner solver produces faster results when in the robust setup with IRLS and GNC. \emph{Original~+~VarPro}, by contrast, recomputes its
elimination at every inner iteration, and this cost compounds under GNC: on the
NTU sequences it requires thousands of iterations against our low hundreds
(e.g., Wi-Fi \emph{NTU R3-01}, $2919$ against $107$) and runs up to $100.9\times$
slower.

Finally, the accuracy that places \emph{Ours} at the bottom of
\cref{fig:robust_paetro} does not come from approximating the problem. Because
our reformulation of the Schur complement product is exact, \emph{Ours}
optimizes the same reduced cost at every IRLS iteration that a direct solver
would, and all methods in the comparison optimize the same robust objective.
Differences in the final estimates therefore arise from each solver's path
through the nonconvex robust cost (e.g., inexact inner solves and the IRLS
weights they produce), not from a modified objective. The heavy ATE tails of
\emph{Original} and \emph{Original~+~VarPro} instead reflect the harder
datasets, on which they settle in worse local minima. The dominant source of
this error is not the shape of each per-robot trajectory but how the
trajectories are aligned across robots in the global frame, as seen in
\cref{fig:trajectory_plot} on the NTU Wi-Fi sequence, where \emph{Original}
fails to recover accurate inter-robot relationships while \emph{Ours} recovers a
consistent multi-robot estimate.

%% file: tab/tab_experiments.tex
\newcommand{\mredx}{\ensuremath{-}}
\newcommand{\goodPct}[1]{
    \textcolor{ForestGreen}{#1}%
}
\newcommand{\badPct}[1]{
    \textcolor{BrickRed}{#1}%
}
\def\best{\bf}
\def\emptyPct{\textemdash}

\newcommand{\dashNA}{\textcolor{gray}{~---\,}}

\newcommand{\pgoMultiRow}{\multirow{8}{*}{\rotatebox[origin=c]{90}{\textbf{PGO}}}}
\newcommand{\raslamMultiRow}{\multirow{8}{*}{\rotatebox[origin=c]{90}{\textbf{RA-SLAM}}}}
\newcommand{\snlMultiRow}{\multirow{8}{*}{\rotatebox[origin=c]{90}{\textbf{SNL}}}}
\newcommand{\sfmMultiRow}{\multirow{17}{*}{\rotatebox[origin=c]{90}{\textbf{SfM}}}}

\newcommand{\speedupCalc}[2]{%
    \begingroup
    \ifstrequal{\detokenize{#1}}{\detokenize{\mredx}}{\dashNA}{%
        \ifstrequal{\detokenize{#1}}{\detokenize{\xmark}}{\dashNA}{%
            \ifstrequal{\detokenize{#2}}{\detokenize{\mredx}}{\dashNA}{%
                \ifstrequal{\detokenize{#2}}{\detokenize{\xmark}}{\dashNA}{%
                    \edef\A{\fpeval{#1}}%
                    \edef\B{\fpeval{#2}}%
                    \ifdim \B pt = 0pt
                        \dashNA
                    \else
                        \edef\ratio{\fpeval{round(\A/\B,2)}}%
                        \ifdim \ratio pt > 1pt
                            \goodPct{\ratio}%
                        \else
                            \badPct{\ratio}%
                        \fi
                    \fi
                }}}}%
    \endgroup
}

\newcommand{\iterCalc}[2]{%
    \begingroup
    \ifstrequal{\detokenize{#1}}{\detokenize{\mredx}}{\dashNA}{%
        \ifstrequal{\detokenize{#1}}{\detokenize{\xmark}}{\dashNA}{%
            \ifstrequal{\detokenize{#2}}{\detokenize{\mredx}}{\dashNA}{%
                \ifstrequal{\detokenize{#2}}{\detokenize{\xmark}}{\dashNA}{%
                    \edef\A{\fpeval{#1}}%
                    \edef\B{\fpeval{#2}}%
                    \ifdim \B pt = 0pt
                        \dashNA
                    \else
                        \fpeval{round(\A/\B,2)}%
                    \fi
                }}}}%
    \endgroup
}

\newif\ifOKBthree
\newif\ifOKCthree
\newif\ifOKBfour
\newif\ifOKCfour
\newif\ifOKDfour
\newif\ifOKBfive
\newif\ifOKCfive
\newif\ifOKDfive
\newif\ifOKEfive


\newif\ifHasLower
\newif\ifSecondCandidate

\newcommand{\checkSecond}[1]{%
    \ifstrequal{#1}{\mredx}{}{%
        \ifstrequal{#1}{\xmark}{}{%
            \edef\T{\fpeval{#1}}%
            \ifdim \T pt < \A pt
                \ifHasLower
                    \ifdim \T pt = \Lower pt
                    \else
                        \SecondCandidatefalse
                    \fi
                \else
                    \edef\Lower{\T}%
                    \HasLowertrue
                \fi
            \fi
        }}%
}

\newcommand{\boldiffour}[4]{%
    \begingroup
    \ifstrequal{#1}{\mredx}{#1\endgroup}{%
        \ifstrequal{#1}{\xmark}{#1\endgroup}{%
            \edef\A{\fpeval{#1}}%
            \ifstrequal{#2}{\mredx}{\OKBfourtrue}{%
                \ifstrequal{#2}{\xmark}{\OKBfourtrue}{%
                    \edef\B{\fpeval{#2}}%
                    \ifdim \A pt > \B pt \OKBfourfalse \else \OKBfourtrue \fi
                }}%
            \ifstrequal{#3}{\mredx}{\OKCfourtrue}{%
                \ifstrequal{#3}{\xmark}{\OKCfourtrue}{%
                    \edef\C{\fpeval{#3}}%
                    \ifdim \A pt > \C pt \OKCfourfalse \else \OKCfourtrue \fi
                }}%
            \ifstrequal{#4}{\mredx}{\OKDfourtrue}{%
                \ifstrequal{#4}{\xmark}{\OKDfourtrue}{%
                    \edef\D{\fpeval{#4}}%
                    \ifdim \A pt > \D pt \OKDfourfalse \else \OKDfourtrue \fi
                }}%
            \HasLowerfalse
            \SecondCandidatetrue
            \checkSecond{#2}%
            \checkSecond{#3}%
            \checkSecond{#4}%
            \ifOKBfour\ifOKCfour\ifOKDfour
                \textbf{\A}%
            \else
                \ifHasLower\ifSecondCandidate\underline{\A}\else\A\fi\else\A\fi
            \fi\else
                \ifHasLower\ifSecondCandidate\underline{\A}\else\A\fi\else\A\fi
            \fi\else
                \ifHasLower\ifSecondCandidate\underline{\A}\else\A\fi\else\A\fi
            \fi
            \endgroup
        }}%
}

\newcommand{\boldiffive}[5]{%
    \begingroup
    \ifstrequal{#1}{\mredx}{#1\endgroup}{%
        \ifstrequal{#1}{\xmark}{#1\endgroup}{%
            \edef\A{\fpeval{#1}}%
            \ifstrequal{#2}{\mredx}{\OKBfivetrue}{%
                \ifstrequal{#2}{\xmark}{\OKBfivetrue}{%
                    \edef\B{\fpeval{#2}}%
                    \ifdim \A pt > \B pt \OKBfivefalse \else \OKBfivetrue \fi
                }}%
            \ifstrequal{#3}{\mredx}{\OKCfivetrue}{%
                \ifstrequal{#3}{\xmark}{\OKCfivetrue}{%
                    \edef\C{\fpeval{#3}}%
                    \ifdim \A pt > \C pt \OKCfivefalse \else \OKCfivetrue \fi
                }}%
            \ifstrequal{#4}{\mredx}{\OKDfivetrue}{%
                \ifstrequal{#4}{\xmark}{\OKDfivetrue}{%
                    \edef\D{\fpeval{#4}}%
                    \ifdim \A pt > \D pt \OKDfivefalse \else \OKDfivetrue \fi
                }}%
            \ifstrequal{#5}{\mredx}{\OKEfivetrue}{%
                \ifstrequal{#5}{\xmark}{\OKEfivetrue}{%
                    \edef\E{\fpeval{#5}}%
                    \ifdim \A pt > \E pt \OKEfivefalse \else \OKEfivetrue \fi
                }}%
            \HasLowerfalse
            \SecondCandidatetrue
            \checkSecond{#2}%
            \checkSecond{#3}%
            \checkSecond{#4}%
            \checkSecond{#5}%
            \ifOKBfive\ifOKCfive\ifOKDfive\ifOKEfive
                \textbf{\A}%
            \else
                \ifHasLower\ifSecondCandidate\underline{\A}\else\A\fi\else\A\fi
            \fi\else
                \ifHasLower\ifSecondCandidate\underline{\A}\else\A\fi\else\A\fi
            \fi\else
                \ifHasLower\ifSecondCandidate\underline{\A}\else\A\fi\else\A\fi
            \fi\else
                \ifHasLower\ifSecondCandidate\underline{\A}\else\A\fi\else\A\fi
            \fi
            \endgroup
        }}%
}

\newcommand{\tableRow}[3]{%
    #1
    \tableRowTimes   #2%
    \tableRowIters   #3%
    \tableRowTimeImp #2%
    \tableRowIterImp #3%
    \\%
}

\newcommand{\tableRowTimes}[5]{%
    & \boldiffour{#1}{#2}{#3}{#4}
    & \boldiffour{#2}{#1}{#3}{#4}
    & \boldiffour{#3}{#1}{#2}{#4}
    & \boldiffour{#4}{#1}{#2}{#3}
}
\newcommand{\tableRowIters}[5]{%
    & #1 & #2 & #3 & #4
}
\newcommand{\tableRowTimeImp}[5]{%
    & \speedupCalc{#2}{#1}\,/\!\speedupCalc{#3}{#1}\,/\!\speedupCalc{#4}{#1}%
}
\newcommand{\tableRowIterImp}[5]{%
    & \iterCalc{#2}{#1}\,/\,\iterCalc{#3}{#1}\,/\,\iterCalc{#4}{#1}%
}

\newcommand{\tableRowGPU}[9]{
    #1
    & \boldiffour{#2}{#3}{#4}{#5}
    & \boldiffour{#3}{#2}{#4}{#5}
    & \boldiffour{#4}{#2}{#3}{#5}
    & \boldiffour{#5}{#2}{#3}{#4}
    & #6 & #7 & #8 & #9
    & \speedupCalc{#3}{#2}\,/\!\speedupCalc{#4}{#2}\,/\!\speedupCalc{#5}{#2}
    & \iterCalc{#7}{#6}\,/\,\iterCalc{#8}{#6}\,/\,\iterCalc{#9}{#6}
    \\%
}
\begingroup
\renewcommand{\arraystretch}{0.9}%
\setlength{\tabcolsep}{3pt}%
\begin{table*}[t]
    \centering
    \caption{ \textbf{CPU Runtime and Iterations Results:}
        The median runtime (in seconds) and the median number of iterations each solver required for the proposed method (Ours) and the Original, Original + VarPro, and GTSAM approaches across all datasets, evaluated from random initializations over 5 trials. The runtime includes our method's required precompute.
        {The runtime improvement factor is defined as $\frac{\text{Baseline}}{\text{Ours}}$.
        Iteration improvement factor is $\frac{\text{Baseline iters}}{\text{Ours iters}}$.
        Values $>1$ indicate our method required less time/iterations than the baseline.
        Each row's fastest solve time is bolded.
        For runtime improvement factors, values $>1$ are green and values $\leq1$ are red.
        We use (\mredx) to indicate a method failed to converge and (\xmark) to indicate a method exceded the 64GB of RAM allocated. \textbf{Bold} is the fastest time while \underline{underline} is the second fastest. \emph{Dense} is omitted because the dense Schur complement is not tractable on CPU; see \cref{tab:gpu_results} for its GPU results.}}
    \label{tab:pgo_wide_gtsam_compare}
    \resizebox{\linewidth}{!}{%
        \begin{tabular}{ll cccc cccc cc}
            \toprule
                            &         & \multicolumn{4}{c}{\textbf{Runtime (s)} $\downarrow$} & \multicolumn{4}{c}{\textbf{Solver Iterations} $\downarrow$} & \multicolumn{2}{c}{\textbf{Improvement Factor} $\uparrow$} \\
            \cmidrule(lr){3-6} \cmidrule(lr){7-10} \cmidrule(lr){11-12}
                            & Dataset & Ours & Original & Orig.+VP & GTSAM
                            & Ours & Original & Orig.+VP & GTSAM
                            & \makecell[c]{Runtime \\{\scriptsize(Orig.\,/\,Orig.+VP\,/\,GTSAM)}}
                            & \makecell[c]{Iterations \\{\scriptsize(Orig.\,/\,Orig.+VP\,/\,GTSAM)}} \\
            \midrule
            \pgoMultiRow
                            & \tableRow{Intel}{{0.08}{0.48}{0.26}{0.35}{1.18}}      {{21}{72}{40}{11}{21}}
                            & \tableRow{Garage}{{8.52}{\mredx}{\mredx}{3.72}{145.33}}      {{250}{\mredx}{\mredx}{63}{261}}
                            & \tableRow{Grid3D}{{3.09}{7.63}{3.81}{\mredx}{64.14}}      {{14}{22}{13}{\mredx}{14}}
                            & \tableRow{MIT}{{0.03}{0.12}{0.07}{0.13}{0.28}}      {{20}{39}{31}{10}{20}}
                            & \tableRow{M3500}{{0.41}{6.62}{4.91}{1.87}{11.21}}      {{31}{188}{120}{21}{31}}
                            & \tableRow{City10000}{{2.26}{13.69}{6.24}{15.57}{96.71}}      {{31}{91}{54}{28}{31}}
                            & \tableRow{Torus}{{0.40}{0.59}{0.53}{18.52}{15.35}}      {{12}{16}{13}{24}{12}}
                            & \tableRow{Sphere}{{0.51}{1.05}{0.95}{9.14}{7.86}}      {{20}{30}{28}{38}{20}}
            \midrule
            \raslamMultiRow
                            & \tableRow{TIERS}{{5.14}{13.14}{9.86}{15.95}{\mredx}}      {{69}{138}{81}{38}{\mredx}}
                            & \tableRow{Single Drone}{{0.23}{1.79}{1.67}{9.66}{14.39}}      {{33}{123}{86}{69}{33}}
                            & \tableRow{Plaza2}{{0.59}{1.18}{0.68}{4.55}{37.79}}      {{35}{74}{37}{40}{35}}
                            & \tableRow{Plaza1}{{4.26}{7.59}{5.92}{12.91}{\mredx}}      {{58}{140}{72}{50}{\mredx}}
                            & \tableRow{MR.CLAM2}{{3.08}{11.27}{9.22}{11.21}{\xmark}}      {{28}{98}{34}{15}{\xmark}}
                            & \tableRow{MR.CLAM4}{{1.80}{6.87}{6.70}{6.61}{\xmark}}      {{24}{62}{28}{11}{\xmark}}
                            & \tableRow{MR.CLAM6}{{1.37}{4.92}{2.76}{5.46}{165.54}}      {{32}{95}{39}{22}{32}}
                            & \tableRow{MR.CLAM7}{{1.32}{5.42}{2.83}{6.61}{184.75}}      {{30}{90}{29}{21}{30}}
                            & \tableRow{Outfinite} {{9.29}{\mredx}{\mredx}{\mredx}{\mredx}} {{76}{\mredx}{\mredx}{\mredx}{\mredx}}
            \midrule
            \snlMultiRow
                            & \tableRow{Intel}{{0.08}{\mredx}{0.54}{\mredx}{1.97}}      {{31}{\mredx}{117}{\mredx}{31}}
                            & \tableRow{Garage}{{10.44}{\mredx}{\mredx}{\mredx}{\mredx}}      {{486}{\mredx}{\mredx}{\mredx}{\mredx}}
                            & \tableRow{Grid3D}{{6.80}{19.27}{9.07}{\mredx}{293.01}}      {{59}{331}{83}{\mredx}{59}}
                            & \tableRow{MIT}{{0.01}{\mredx}{0.05}{0.09}{0.05}}      {{13}{\mredx}{60}{11}{13}}
                            & \tableRow{M3500}{{0.55}{\mredx}{3.23}{\mredx}{25.45}}      {{59}{\mredx}{305}{\mredx}{59}}
                            & \tableRow{City10000}{{17.62}{\mredx}{\mredx}{\mredx}{\mredx}}      {{260}{\mredx}{\mredx}{\mredx}{\mredx}}
                            & \tableRow{Torus}{{1.18}{\mredx}{5.73}{\mredx}{42.73}}      {{41}{\mredx}{123}{\mredx}{41}}
                            & \tableRow{Sphere}{{1.50}{\mredx}{2.34}{\mredx}{42.73}}      {{65}{\mredx}{372}{\mredx}{65}}
            \midrule
            \sfmMultiRow
                            & \tableRow{BAL-93}{{0.14}{2.32}{0.98}{12.69}{0.12}}      {{18}{121}{45}{65}{18}}
                            & \tableRow{BAL-392}{{4.56}{198.97}{140.73}{26.45}{4.36}}      {{20}{303}{165}{2}{21}}
                            & \tableRow{BAL-1934}{{77.15}{\mredx}{\mredx}{\xmark}{128.07}}      {{23}{\mredx}{\mredx}{\xmark}{23}}
                            & \tableRow{IMC Gate}{{52.79}{\mredx}{\mredx}{\xmark}{64.35}}      {{32}{\mredx}{\mredx}{\xmark}{32}}
                            & \tableRow{IMC Temple}{{8.65}{87.29}{40.08}{\xmark}{23.40}}      {{21}{86}{49}{\xmark}{21}}
                            & \tableRow{IMC Rome}{{56.33}{\mredx}{\mredx}{\xmark}{114.66}}      {{36}{\mredx}{\mredx}{\xmark}{36}}
                            & \tableRow{Rep. Office0-100}{{0.17}{1.58}{0.80}{6.35}{0.17}}      {{12}{40}{24}{9}{12}}
                            & \tableRow{Rep. Office1-100}{{0.07}{1.10}{0.54}{2.84}{0.08}}      {{11}{77}{27}{8}{11}}
                            & \tableRow{Rep. Room0-100}{{0.34}{2.42}{1.13}{9.51}{0.31}}      {{14}{39}{20}{9}{14}}
                            & \tableRow{Rep. Room1-100}{{0.24}{3.36}{1.40}{9.02}{0.24}}      {{13}{52}{31}{11}{13}}
                            & \tableRow{Mip-NeRF Garden}{{0.48}{4.16}{2.50}{19.31}{0.69}}      {{11}{46}{24}{12}{11}}
                            & \tableRow{Mip-NeRF Room}{{7.04}{64.49}{29.70}{21.50}{5.53}}      {{21}{156}{52}{6}{21}}
                            & \tableRow{Mip-NeRF Kitchen}{{4.13}{109.29}{44.85}{20.76}{4.75}}      {{14}{117}{48}{4}{14}}
                            & \tableRow{TUM Room}{{7.76}{146.36}{64.28}{\xmark}{21.09}}      {{20}{175}{61}{\xmark}{20}}
                            & \tableRow{TUM Desk}{{3.75}{62.42}{22.00}{\xmark}{7.01}}      {{13}{121}{51}{\xmark}{13}}
                            & \tableRow{TUM Comp-R}{{5.10}{81.77}{30.57}{\xmark}{9.01}}      {{18}{127}{48}{\xmark}{18}}
                            & \tableRow{TUM Comp-T}{{8.75}{212.38}{59.58}{\xmark}{16.56}}      {{18}{172}{53}{\xmark}{18}}
            \bottomrule
        \end{tabular}
    }
    \vspace{-1.2em}
\end{table*}
\endgroup

\begingroup
\renewcommand{\arraystretch}{0.9}%
\setlength{\tabcolsep}{3pt}%
\begin{table*}[t]
    \centering
    \caption{
        \textbf{GPU Runtime and Iterations Results:}
        The median runtime (in seconds) and the median number of iterations each solver required for the proposed method (Ours), Original, Original + VarPro, and Dense on a GPU implementation, evaluated from random initializations over 5 trials. The runtime includes our method's required precompute and the formations of \emph{Dense's} Schur Complement formulation.
        {The runtime improvement factor is defined as $\frac{\text{Baseline}}{\text{Ours}}$.
        Iteration improvement factor is $\frac{\text{Baseline iters}}{\text{Ours iters}}$.
        Values $>1$ indicate our method required less time/iterations than the baseline.
        Each row's fastest solve time is bolded.
        For runtime improvement factors, values $>1$ are green and values $\leq1$ are red.
         We use (\mredx) to indicate a method failed to converge and (\xmark) to indicated that the 24GB of RAM allocated was exceeded. \textbf{Bold} is the fastest time while \underline{underline} is the second fastest.}}
    
    \label{tab:gpu_results}
    \resizebox{\linewidth}{!}{%
        \begin{tabular}{ll cccc cccc cc}
            \toprule
                            &         & \multicolumn{4}{c}{\textbf{Runtime (s)} $\downarrow$} & \multicolumn{4}{c}{\textbf{Solver Iterations} $\downarrow$} & \multicolumn{2}{c}{\textbf{Improvement Factor} $\uparrow$} \\
            \cmidrule(lr){3-6} \cmidrule(lr){7-10} \cmidrule(lr){11-12}
                            & Dataset & Ours & Original & Orig. + VP & Dense & Ours & Original & Orig. + VP & Dense & \makecell[c]{Runtime \\{\scriptsize(Orig.\,/Orig.+VP/\,Dense)}} & \makecell[c]{Iterations \\{\scriptsize(Orig.\,/Orig.+VP/\,Dense)}} \\
            \midrule
\pgoMultiRow    & \tableRowGPU{Intel}{0.10}{0.41}{0.20}{0.21}{20}{65}{22}{20}
                & \tableRowGPU{Garage}{5.62}{\mredx}{\mredx}{22.37}{231}{\mredx}{\mredx}{236}
                & \tableRowGPU{Grid3D}{0.82}{1.76}{1.28}{7.67}{15}{32}{19}{15}
                & \tableRowGPU{MIT}{0.07}{0.18}{0.09}{0.09}{18}{40}{23}{18}
                & \tableRowGPU{M3500}{0.33}{3.86}{1.52}{1.09}{33}{156}{47}{33}
                & \tableRowGPU{City10000}{0.81}{3.70}{2.00}{14.84}{29}{87}{38}{29}
                & \tableRowGPU{Torus}{0.30}{0.51}{0.49}{1.41}{12}{21}{21}{12}
                & \tableRowGPU{Sphere}{0.26}{0.52}{0.41}{0.57}{18}{34}{23}{18}
\midrule
\raslamMultiRow & \tableRowGPU{TIERS}{2.82}{4.57}{3.94}{170.70}{68}{126}{75}{68}
                & \tableRowGPU{Single Drone}{0.31}{1.56}{0.69}{1.11}{36}{138}{36}{37}
                & \tableRowGPU{Plaza2}{0.67}{1.01}{0.65}{7.02}{35}{70}{34}{35}
                & \tableRowGPU{Plaza1}{2.17}{3.64}{2.20}{86.19}{54}{85}{55}{55}
                & \tableRowGPU{MR.CLAM2}{1.37}{6.22}{2.23}{\xmark}{28}{112}{26}{\xmark}
                & \tableRowGPU{MR.CLAM4}{0.93}{3.78}{1.59}{\xmark}{24}{64}{23}{\xmark}
                & \tableRowGPU{MR.CLAM6}{0.65}{2.81}{1.26}{20.56}{33}{91}{33}{33}
                & \tableRowGPU{MR.CLAM7}{0.71}{2.94}{1.21}{29.30}{31}{93}{31}{31}
                & \tableRowGPU{Outfinite}{2.80}{\mredx}{\mredx}{197.70}{73}{\mredx}{\mredx}{75}
\midrule
\snlMultiRow    & \tableRowGPU{Intel}{0.42}{\mredx}{1.14}{0.52}{34}{\mredx}{83}{34}
                & \tableRowGPU{Garage}{6.62}{\mredx}{\mredx}{\mredx}{249}{\mredx}{\mredx}{\mredx}
                & \tableRowGPU{Grid3D}{1.86}{\mredx}{2.42}{46.67}{58}{\mredx}{74}{58}
                & \tableRowGPU{MIT}{0.08}{\mredx}{0.37}{0.09}{14}{\mredx}{39}{14}
                & \tableRowGPU{M3500}{0.83}{\mredx}{3.30}{3.98}{53}{\mredx}{214}{53}
                & \tableRowGPU{City10000}{9.14}{\mredx}{\mredx}{\mredx}{201}{\mredx}{\mredx}{\mredx}
                & \tableRowGPU{Torus}{0.59}{\mredx}{2.15}{4.56}{41}{\mredx}{120}{41}
                & \tableRowGPU{Sphere}{1.62}{\mredx}{\mredx}{4.46}{54}{\mredx}{\mredx}{54}
\midrule
\sfmMultiRow    & \tableRowGPU{BAL-93}{0.11}{2.35}{0.52}{0.09}{18}{213}{24}{18}
                & \tableRowGPU{BAL-392}{1.07}{\mredx}{29.16}{0.90}{19}{\mredx}{126}{19}
                & \tableRowGPU{BAL-1934}{13.32}{\mredx}{43.04}{15.71}{23}{\mredx}{28}{23}
                & \tableRowGPU{IMC Gate}{6.97}{\mredx}{184.98}{8.35}{30}{\mredx}{154}{31}
                & \tableRowGPU{IMC Temple}{2.96}{21.10}{6.79}{3.35}{16}{94}{17}{16}
                & \tableRowGPU{IMC Rome}{11.24}{\mredx}{139.49}{13.47}{25}{\mredx}{73}{25}
                & \tableRowGPU{Rep. Office0-100}{0.09}{0.80}{0.34}{0.11}{10}{41}{13}{10}
                & \tableRowGPU{Rep. Office1-100}{0.05}{0.53}{0.21}{0.07}{12}{63}{14}{12}
                & \tableRowGPU{Rep. Room0-100}{0.13}{1.02}{0.49}{0.16}{11}{40}{14}{11}
                & \tableRowGPU{Rep. Room1-100}{0.12}{1.83}{0.65}{0.15}{12}{59}{14}{12}
                & \tableRowGPU{Mip-NeRF Garden}{0.29}{1.99}{1.05}{0.29}{14}{53}{17}{14}
                & \tableRowGPU{Mip-NeRF Room}{2.03}{13.29}{4.30}{1.42}{21}{140}{22}{22}
                & \tableRowGPU{Mip-NeRF Kitchen}{1.35}{21.77}{7.26}{1.36}{16}{112}{22}{16}
                & \tableRowGPU{TUM Room}{2.00}{34.50}{9.45}{2.36}{23}{170}{22}{23}
                & \tableRowGPU{TUM Desk}{1.02}{11.79}{4.38}{1.00}{14}{100}{19}{14}
                & \tableRowGPU{TUM Comp-R}{1.13}{10.97}{4.03}{1.10}{15}{113}{17}{15}
                & \tableRowGPU{TUM Comp-T}{1.97}{33.67}{9.12}{2.10}{18}{157}{21}{18}
            \bottomrule
        \end{tabular}
    }
    \vspace{-1.2em}
\end{table*}
\endgroup

%% file: tab/robustcpu.tex
\definecolor{ImpGood}{HTML}{1B7F3B}
\definecolor{ImpBad}{HTML}{C0392B}

\ExplSyntaxOn

\cs_new_protected:Npn \tbl_min_update:n #1
  {
    \tl_if_eq:nnF {#1} {\mredx}
      {
        \fp_compare:nT { #1 < \l_tmpa_fp }
          { \fp_set:Nn \l_tmpa_fp {#1} }
      }
  }

\cs_new_protected:Npn \tbl_min2_update:n #1
  {
    \tl_if_eq:nnF {#1} {\mredx}
      {
        \fp_compare:nTF { #1 < \l_tmpa_fp }
          {
            \fp_set_eq:NN \l_tmpb_fp \l_tmpa_fp
            \fp_set:Nn \l_tmpa_fp {#1}
          }
          {
            \fp_compare:nT { #1 < \l_tmpb_fp }
              { \fp_set:Nn \l_tmpb_fp {#1} }
          }
      }
  }

\NewDocumentCommand \impval { m m }
  {
    \tl_if_eq:nnTF {#1} {\mredx}
      { \mredx }
      {
        \tl_if_eq:nnTF {#2} {\mredx}
          { \mredx }
          {
            \fp_set:Nn \l_tmpa_fp { round(#1/#2,1) }
            \fp_compare:nTF { \l_tmpa_fp > 1 }
              { \textcolor{ImpGood}{ \fp_to_decimal:N \l_tmpa_fp } }
              {
                \fp_compare:nTF { \l_tmpa_fp < 1 }
                  { \textcolor{ImpBad}{ \fp_to_decimal:N \l_tmpa_fp } }
                  { \fp_to_decimal:N \l_tmpa_fp }
              }
          }
      }
  }

\NewDocumentCommand \plainratio { m m }
  {
    \tl_if_eq:nnTF {#1} {\mredx}
      { \mredx }
      {
        \tl_if_eq:nnTF {#2} {\mredx}
          { \mredx }
          { \fp_eval:n { round(#1/#2,1) } }
      }
  }

\NewDocumentCommand \boldminIV { m m m m m }
  {
    \tl_if_eq:nnTF {#1} {\mredx}
      { \mredx }
      {
        \fp_set:Nn \l_tmpa_fp { 1e99 }
        \fp_set:Nn \l_tmpb_fp { 1e99 }

        \tbl_min2_update:n {#2}
        \tbl_min2_update:n {#3}
        \tbl_min2_update:n {#4}
        \tbl_min2_update:n {#5}

        \fp_compare:nTF { #1 = \l_tmpa_fp }
          { \textbf{#1} }
          {
            \fp_compare:nTF { #1 = \l_tmpb_fp }
              { \underline{#1} }
              { #1 }
          }
      }
  }

\ExplSyntaxOff
\newcommand{\rowCB}[3]{#1 \rowRun #2 \rowIt #3 \rowRunImp #2 \rowItImp #3 \\}
\newcommand{\rowRun}[4]{%
  & \boldminIV{#1}{#1}{#2}{#3}{#4}
  & \boldminIV{#2}{#1}{#2}{#3}{#4}
  & \boldminIV{#3}{#1}{#2}{#3}{#4}
  & \boldminIV{#4}{#1}{#2}{#3}{#4}}
\newcommand{\rowIt}[4]{& #1 & #2 & #3 & #4}
\newcommand{\rowRunImp}[4]{%
  & \impval{#2}{#1}\,/\,\impval{#3}{#1}\,/\,\impval{#4}{#1}}
\newcommand{\rowItImp}[4]{%
  & \plainratio{#2}{#1}\,/\,\plainratio{#3}{#1}\,/\,\plainratio{#4}{#1}}
\def\wifiMultiRow{\multirow[c]{10}{*}{\rotatebox[origin=c]{90}{\textbf{CosmoBench Wi-Fi}}}}
\def\proradioMultiRow{\multirow[c]{10}{*}{\rotatebox[origin=c]{90}{\textbf{CosmoBench Radio}}}}
\def\nebulaMultiRow{\multirow[c]{4}{*}{\rotatebox[origin=c]{90}{\textbf{Nebula}}}}
\begingroup
\renewcommand{\arraystretch}{0.9}%
\setlength{\tabcolsep}{3pt}%
\begin{table*}[t]
    \centering
    \caption{%
        \textbf{Robust IRLS-GNC (Geman--McClure) on CosmoBench
        \citep{mcgann2025cosmobenchbenchmarkcollaborativeslam} and Nebula
        \citep{chang2022lamp} multi-robot pose-graph datasets.} The runtime (in seconds) and number of
        iterations each solver required for the proposed method (Ours),
        Original, Original + VarPro, and GTSAM. Each entry is the median over
        five odometry-chained initializations with added noise, with every solver given the
        identical initialization on a given trial.
        Runtime improvement factor is
        $\frac{\text{Baseline}}{\text{Ours}}$; iteration improvement is
        $\frac{\text{Baseline iters}}{\text{Ours iters}}$. Values $>1$ indicate our method required less time/iterations than the baseline. \textbf{Bold} is the fastest time while \underline{underline} is the second fastest.%
    }
    \label{tab:cosmobench_irls_gnc}
    \resizebox{\linewidth}{!}{%
        \begin{tabular}{ll cccc cccc cc}
            \toprule
                & & \multicolumn{4}{c}{\textbf{Runtime (s)} $\downarrow$}
                  & \multicolumn{4}{c}{\textbf{Solver iters} $\downarrow$}
                  & \multicolumn{2}{c}{\textbf{Improvement} $\uparrow$} \\
            \cmidrule(lr){3-6} \cmidrule(lr){7-10} \cmidrule(lr){11-12}
                & Dataset
                & Ours & Original. & Orig.\,+\,VP & GTSAM
                & Ours & Original & Orig.\,+\,VP & GTSAM
                & \makecell[c]{Runtime \\{\scriptsize(Orig.\,/\,Orig.+VP\,/\,GTSAM)}}
                & \makecell[c]{Iterations \\{\scriptsize(Orig.\,/\,Orig.+VP\,/\,GTSAM)}} \\
            \midrule
\wifiMultiRow
  & \rowCB{Kittredge Loop} {{5.59}{34.54}{27.53}{6.02}}        {{99}{203}{140}{86}}
  & \rowCB{Main Campus}    {{3.64}{54.84}{32.37}{4.10}}        {{86}{266}{170}{81}}
  & \rowCB{KTH R3-00}      {{1.80}{49.69}{45.78}{2.07}}        {{98}{876}{726}{138}}
  & \rowCB{KTH R3-01}      {{2.24}{106.46}{81.58}{4.43}}       {{141}{1717}{1598}{302}}
  & \rowCB{KTH R4-00}      {{3.17}{132.59}{101.12}{4.05}}      {{115}{1253}{1032}{203}}
  & \rowCB{NTU R3-00}      {{4.80}{\mredx}{\mredx}{5.75}}      {{102}{\mredx}{\mredx}{194}}
  & \rowCB{NTU R3-01}      {{2.74}{\mredx}{276.53}{6.07}}      {{107}{\mredx}{2919}{298}}
  & \rowCB{NTU R3-02}      {{1.26}{\mredx}{\mredx}{2.82}}      {{88}{\mredx}{\mredx}{208}}
  & \rowCB{NTU R4-00}      {{3.36}{\mredx}{176.38}{5.72}}      {{102}{\mredx}{1412}{160}}
  & \rowCB{NTU R5-00}      {{5.32}{\mredx}{\mredx}{16.72}}     {{123}{\mredx}{\mredx}{448}}
\midrule
\proradioMultiRow
  & \rowCB{Kittredge Loop} {{4.87}{40.39}{22.90}{7.18}}        {{97}{221}{140}{82}}
  & \rowCB{Main Campus}    {{4.31}{34.29}{23.37}{4.99}}        {{89}{168}{119}{69}}
  & \rowCB{KTH R3-00}      {{1.92}{60.63}{55.31}{3.05}}        {{98}{737}{661}{181}}
  & \rowCB{KTH R3-01}      {{2.67}{100.45}{97.77}{3.22}}       {{129}{1225}{1171}{191}}
  & \rowCB{KTH R4-00}      {{2.66}{126.21}{80.99}{3.39}}       {{103}{1096}{753}{158}}
  & \rowCB{NTU R3-00}      {{2.79}{\mredx}{186.25}{6.58}}      {{100}{\mredx}{1668}{208}}
  & \rowCB{NTU R3-01}      {{1.89}{\mredx}{\mredx}{5.84}}      {{97}{\mredx}{\mredx}{277}}
  & \rowCB{NTU R3-02}      {{1.08}{\mredx}{58.50}{2.46}}       {{96}{\mredx}{1315}{175}}
  & \rowCB{NTU R4-00}      {{4.10}{\mredx}{254.29}{7.24}}      {{111}{\mredx}{2009}{202}}
  & \rowCB{NTU R5-00}      {{4.19}{\mredx}{244.09}{9.42}}      {{110}{\mredx}{1808}{253}}
\midrule
\nebulaMultiRow
  & \rowCB{Finals}         {{0.60}{3.52}{3.45}{1.04}}          {{101}{187}{133}{95}}
  & \rowCB{Kentucky UG}    {{5.39}{\mredx}{\mredx}{\mredx}}    {{105}{\mredx}{\mredx}{\mredx}}
  & \rowCB{Tunnel}         {{1.50}{10.32}{8.50}{2.41}}         {{77}{165}{109}{69}}
  & \rowCB{Urban}          {{0.59}{2.28}{1.93}{1.00}}          {{99}{148}{99}{85}}

            \bottomrule
        \end{tabular}
    }
\end{table*}
\endgroup

%% file: sec/6_conclusion.tex
\section{Conclusion}
\label{sec:conclusion}

This paper introduced SPARSER: a novel method to accelerate the solution of
large-scale, separable nonlinear least-squares problems prevalent in
robotic perception. Our approach leverages a sparsity-preserving VarPro
scheme that can be applied as an efficient, one-time preprocessing step.
The scheme represents the reduced problem with an efficient matrix-free
Schur complement operator that readily integrates with standard iterative
optimization frameworks.

\textbf{Results, takeaways, and contributions.}
Experiments on diverse problems demonstrated the practical benefits of
our approach. Across 42 benchmark datasets spanning pose-graph optimization,
range-aided SLAM, sensor network localization, and structure from motion, our
method was the fastest on 41 on CPU and 35 on GPU against state-of-the-art
baselines, including a prior VarPro technique~\citep{khosoussi2016Sparse},
and was the only method to solve all 42. Our method also proved more
scalable and memory-efficient than the widely deployed GTSAM solver,
most notably on eight large-scale SfM instances where GTSAM exceeded
64\,GB of RAM. Comparing against our most direct counterpart, the
matrix-free operator provides a far more scalable alternative, matching
the iteration counts of the dense Schur complement while remaining
tractable on the two RA-SLAM datasets where the dense complement exceeded
24\,GB of GPU memory. We further showed that the framework
extends gracefully to robust costs handled via iteratively reweighted
least squares, in which the symbolic preprocessing is reused across
outer reweightings and each iteration of graduated non-convexity
requires only a numerical refresh. On 24 real-world multi-robot SLAM
datasets with naturally outlier-corrupted loop closures (CosmoBench and
Nebula), our robust variant matches the trajectory accuracy of
GTSAM-GNC while being the fastest method on all 24, running $1.1\times$
to $3.1\times$ faster than GTSAM-GNC on every dataset tested.

The core of our method's
success lies in its ability to exploit both problem separability and
sparsity, overcoming the challenges posed by gauge symmetries that have
limited previous approaches. We
also provided a clear characterization of the class of problems for
which our method is applicable, offering a practical guide for its
adoption on new problems. 

\textbf{Limitations and future work.} SPARSER performs well due to the specific structure enabled by the example problems shown and thus still remains limited.
Our method requires a set of unconstrained variables to eliminate, so
problems such as pure rotation averaging~\citep{hartley2013rotation} fall
outside its scope. The largest efficiency gains additionally rely on
linear residuals and an unconstrained Jacobian $\matAf$ with
incidence-like structure. While these
conditions are surprisingly common in robotic perception, the most
prevalent violation is nonlinear residuals, most notably the
reprojection errors in full bundle adjustment. In such cases the general
VarPro scheme still applies, but the preprocessing must be repeated at
each linearization and the per-iteration efficiency gain is reduced.
Characterizing how much of the benefit can be retained in the
nonlinear-residual situations and extending the framework to problems lacking
incidence-like structure via approximate column-space
bases~\citep{coleman1987null} are natural directions of future work.

To support further research and application,
our work is released as an open-source C++ library, including all
datasets and baselines from our experiments.

%% file: sec/9_appendix.tex
\section*{Appendix}
\begin{appendices}

\section{Inhomogeneous Linear Residuals}
\label{sec:appendix:quadratic-derivation}

We walk through the derivation of the quadratic cost form
(\cref{sec:problem-formulation}) for the inhomogeneous case. The residuals
of interest are of the form
\begin{equation}
    r_i(X) = A_i X + B_i, \quad i = 1, \dots, m.
\end{equation}
By introducing a homogenized variable
\begin{equation}
    \tilde{\matX} \triangleq \begin{bmatrix} \matX \\ I_d \end{bmatrix}
    \in \R^{(n+d) \times d}
\end{equation}
that absorbs the constant terms, each residual can be written as
\begin{equation}
    \tilde{H}_i\, \tilde{\matX}
    = \begin{bmatrix} A_i & B_i \end{bmatrix}
      \begin{bmatrix} \matX \\ I_d \end{bmatrix}
    = A_i \matX + B_i,
\end{equation}
and the stacked cost takes the homogeneous quadratic form
\begin{equation}
    \sum_i \big\|\tilde{r}_i(\matX)\big\|_{\matOmega_i}^2
    \,=\, \big\|\tilde{\matH}\, \tilde{\matX}\big\|_{\tilde{\matOmega}}^2
    \,=\, \tr\!\left(\tilde{\matX}^\top \tilde{\matH}^\top \tilde{\matOmega}\,
    \tilde{\matH}\, \tilde{\matX}\right),
\end{equation}
with
\begin{align}
    \tilde{\matH}     & \triangleq \begin{bmatrix} \matA \mid \tilde{\matB} \end{bmatrix},                                                            \\
    \matA             & \triangleq \begin{bmatrix} \matA_1^\top \cdots \matA_m^\top \end{bmatrix}^\top,                                               \\
    \tilde{\matB}     & \triangleq \begin{bmatrix} B_1^\top \cdots B_m^\top \end{bmatrix}^\top,                                                       \\
    \tilde{\matOmega} & \triangleq \mathrm{blkdiag}(\matOmega_1, \ldots, \matOmega_m).
\end{align}

The result is structurally identical to the homogeneous form derived in
\cref{sec:problem-formulation}, but over the augmented variable
$\tilde{\matX}$ whose trailing block is fixed to $I_d$. Optimization is
therefore carried out only over the top block $\matX$, with the constraint
set and the partition into constrained and unconstrained variables
inherited unchanged from \cref{prob:nlls}. All subsequent development, the
separable-structure analysis of \cref{sec:prelim:separable-structure}, the
matrix-free Schur reformulation of
\cref{sec:prelim:focus-schur-complement-products}, and the graph-theoretic
sparsity conditions of
\cref{sec:prelim:efficient-schur-complement-products}, applies verbatim
with $\tilde{\matH}$ playing the role of $\matA$ and $\tilde{\matOmega}$
the role of $\matOmega$.

\section{Simplifying the Reduced Cost}
\label{sec:appendix:reduced-cost-derivation}

We can write the cost of our original quadratic cost problem
(\cref{prob:general-qcqp}) as
\begin{equation}
    \begin{aligned}
        \tr(X^\top Q X) =\;
        & \underbrace{\tr\!\left(\matXc^\top \matQcc \matXc\right)}_{A} \\
        & + \underbrace{\tr\!\left(\matXf^\top \matQff \matXf\right)}_{B} \\
        & + \underbrace{2 \tr\!\left(\matXc^\top \matQcf \matXf\right)}_{C}.
    \end{aligned}
\end{equation}
In \cref{sec:prelim:separable-structure} we showed that for a fixed
$\matXc$, the optimal value of $\matXf$ is
\begin{equation}
    \matXfstar = -\matQff^{\dagger} \matQfc \matXc,
\end{equation}
where $\matQff^{\dagger}$ is the Moore-Penrose pseudoinverse of $\matQff$.
Substituting $\matXfstar$ into each piece,
\begin{align*}
    A & = \tr\!\left(\matXc^\top \matQcc \matXc\right),                                                                            \\
    B & = \tr\!\left((\matXfstar)^\top \matQff \matXfstar\right)                                                                   \\
      & = \tr\!\left((-\matQff^{\dagger} \matQfc \matXc)^\top \matQff (-\matQff^{\dagger} \matQfc \matXc)\right)                   \\
      & = \tr\!\left(\matXc^\top \matQcf \matQff^{\dagger} \matQff \matQff^{\dagger} \matQfc \matXc\right)                         \\
      & = \tr\!\left(\matXc^\top \matQcf \matQff^{\dagger} \matQfc \matXc\right),                                                  \\
    C & = 2 \tr\!\left(\matXc^\top \matQcf \matXfstar\right)                                                                       \\
      & = -2 \tr\!\left(\matXc^\top \matQcf \matQff^{\dagger} \matQfc \matXc\right)                                                \\
      & = -2B,
\end{align*}
where the simplification of $B$ uses the Moore-Penrose identity
$\matQff^{\dagger} \matQff \matQff^{\dagger} = \matQff^{\dagger}$.
Combining the three terms,
\begin{equation}
    \tr(X^\top Q X) \;=\; A + B + C \;=\; A + B - 2B \;=\; A - B,
\end{equation}
yielding the reduced cost
\begin{equation}
\begin{aligned}
    \tr(X^\top Q X)
    &= \tr\!\left(\matXc^\top \left(\matQcc - \matQcf \matQff^{\dagger} \matQfc\right) \matXc\right) \\
    &= \tr\!\left(\matXc^\top \matQmarg\, \matXc\right).
\end{aligned}
\end{equation}
which recovers the Schur complement $\matQmarg$ of \cref{eq:reduced-cost}.

\section{Anchoring}
An anchored measurement is not a pairwise difference between unconstrained
variables, so, like any residual that is not incidence-like
(\cref{sec:prelim:efficient-schur-complement-products}), it departs from the
incidence pattern that our efficient construction exploits. It does so, however, in
the most benign way possible. Rather than invoking the basis-construction machinery
of \cref{sec:prelim:efficient-schur-complement-products}, an anchor \emph{removes}
the gauge symmetry that makes $\matAf$ rank-deficient, leaving a full-rank reduced
problem and requiring no change to the algorithm.

\textbf{Structure.}
A range to a known anchor $a_k$ has residual $t_j - a_k - u_{jk}\,\tilde d_{jk}$,
and a direct position fix has residual $t_j - \tilde a_j$, and both are linear, with
the constant absorbed by the homogenization of Appendix~A. In contrast to the
relative residual $t_j - t_i - u_{ij}\,\tilde d_{ij}$ of Table~1, which is a $\pm I$
pair, the Jacobian of an anchored residual with respect to the unconstrained
translations is a lone $+I$ block, a \emph{grounding} row rather than an edge. Hence
$\matAf$ is not an incidence matrix.

\textbf{Full rank.}
The incidence null space is the all-ones translation gauge (Remark~2), since relative
measurements see only differences, so a global shift leaves the cost unchanged. An
absolute measurement breaks exactly this invariance, as shifting all translations
changes $t_j - a_k$, so the all-ones vector leaves $\ker \matAf$. With at least one
anchor per connected component, $\matAf$ has full column rank and
\begin{equation}
\label{eq:grounded-laplacian}
\matAf^\top \matOmega \matAf = L_{\mathrm{rel}} + D_{\mathrm{anc}} \;\succ\; 0,
\end{equation}
the \emph{grounded} (Dirichlet) Laplacian, which is the weighted Laplacian
$L_{\mathrm{rel}}$ of the relative-measurement graph plus a positive diagonal
$D_{\mathrm{anc}}$ contributed by the anchors. It is sparse (inheriting the graph)
and positive definite.

\textbf{Operator.}
Full column rank is precisely the classical variable-projection setting noted in
\cref{sec:prelim:focus-schur-complement-products}, where the pseudoinverse in
$\matQmarg$ becomes an ordinary inverse and the CR decomposition is unnecessary. One
takes $\matC = \matAf$ directly, with no column deletion and no basis construction,
so the reduced matrix $\matM = \matAf^\top \matOmega \matAf$ is the grounded Laplacian
of \cref{eq:grounded-laplacian}, and a single sparse Cholesky yields the operator.
\cref{eq:Qmarg-product-final} and \cref{alg:schur-complement-product} then run
verbatim.

\textbf{Positioning.}
Grounding an anchored node and deleting a reference node
(\cref{sec:prelim:efficient-schur-complement-products}) are two routes to the same
full-rank reduced Laplacian. Anchoring supplies the datum that the gauge otherwise
leaves free, so the reference-node deletion can simply be skipped. Anchoring
therefore removes the very rank deficiency that motivated the column-deletion
machinery.

\section{Further Details on the Outfinite Dataset}
\label{app:outfinite}

The \emph{Outfinite} dataset (Fig.~\ref{fig:outfinite}) is the largest and most
challenging problem in our evaluation. It is a three-agent, range-aided SLAM
sequence that we collected on the MIT campus to address the scarcity of
large-scale, multi-agent, range-aided benchmarks. Here we describe the collection
procedure, the salient characteristics of the raw data, and the preprocessing used
to prepare the data for state estimation.

\begin{figure}[t]
  \centering
  \includegraphics[width=0.45\textwidth]{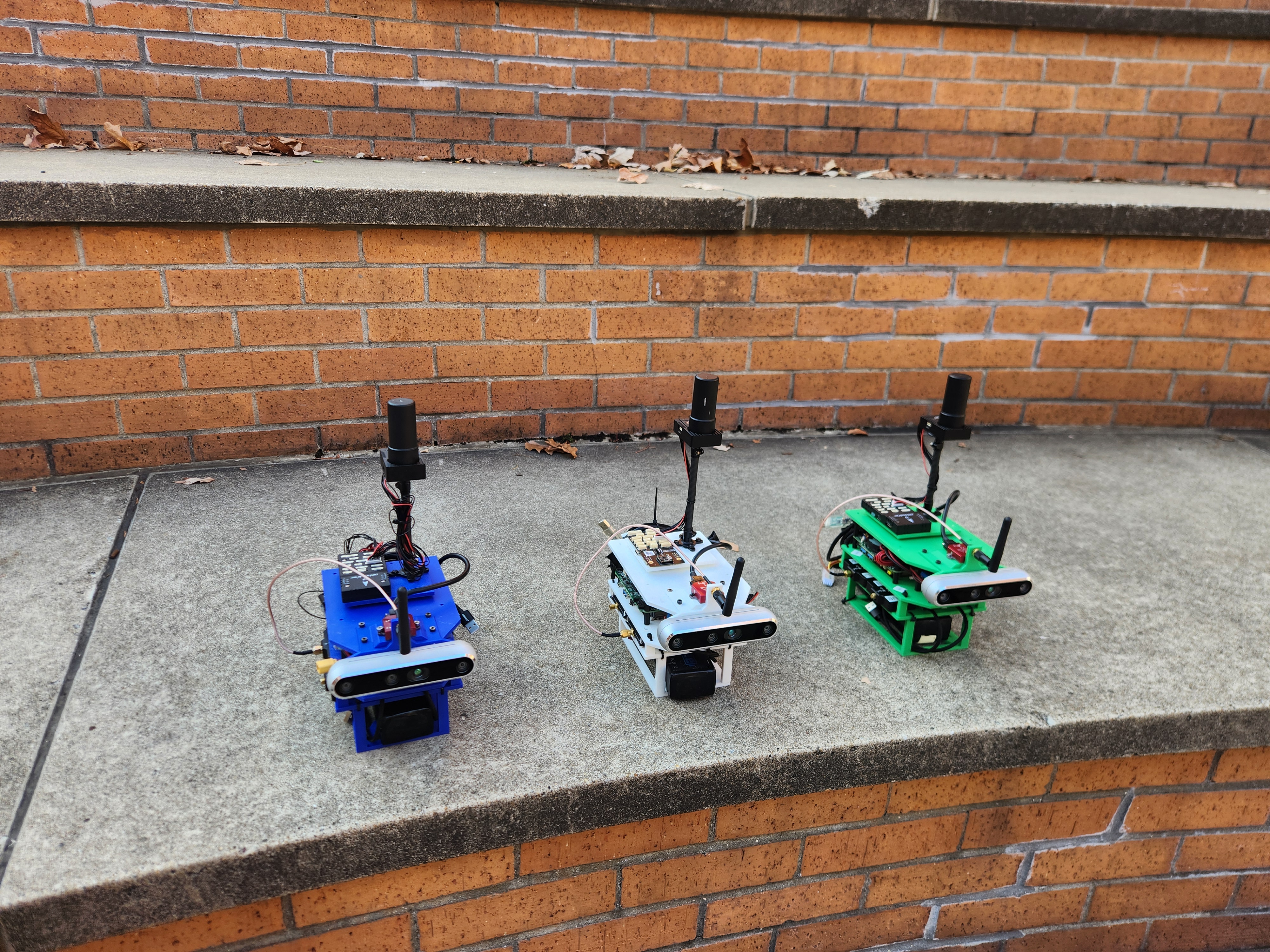}
  \caption{The three sensor payloads used to collect the \emph{Outfinite}
  dataset (\cref{fig:outfinite}), each comprising a Nooploop LinkTrack P-B UWB
  radio, a Vectornav VN-100 AHRS, an Intel RealSense D455 RGB-D camera, and a
  GPS receiver. The GPS is logged only as a positional reference for ground truth comparison and is never
  used in state estimation.}
  \label{fig:payload}
\end{figure}

\textbf{Data collection.} The data were collected mid-day along a major outdoor
walkway on the MIT campus, colloquially the ``Outfinite Corridor'' as it runs
along the exterior of MIT's Infinite Corridor. Three human operators each carried
a sensor payload (Fig.~\ref{fig:payload}) along the corridor, together traversing
roughly $6$\,km (about $2$\,km per payload) and generating approximately $13.5$k
poses. The environment contained numerous dynamic elements such as pedestrians and
moving vehicles, as well as structural occlusions from surrounding buildings and
infrastructure. Each payload comprised a Nooploop LinkTrack P-B ultra-wideband
(UWB) radio, a Vectornav VN-100 attitude-and-heading reference system (AHRS), an
Intel RealSense D455 RGB-D camera, and a GPS receiver logged as a positional
reference. Six static UWB beacons were placed throughout the environment, and the
UWB radios provided both inter-agent ranges and ranges to these beacons, yielding
roughly $7.6$k range measurements in total. Images were recorded at $30$\,Hz,
AHRS data at $200$\,Hz, UWB ranges at $50$\,Hz, and GPS at $10$\,Hz.

\textbf{Imagery and lighting.} The RealSense D455 provides a single global-shutter
RGB image together with a depth image computed from a stereo infrared pair and an
infrared projector. As is typical for outdoor vision, lighting strongly affected
image quality. The mid-day collection spanned a wide range of conditions,
including large saturated regions and a characteristic purple tint that the
RealSense exhibits under strong illumination. This
variability motivates fusing range and vision, since vision provides reliable
frame-to-frame tracking while range corrects the accumulated drift. Unlike visual
SLAM approaches that depend on visual place recognition for loop closure,
range-aided navigation does not require a scene to be visually recognizable across
viewpoints.

\begin{figure}[t]
  \centering
  \includegraphics[width=\linewidth]{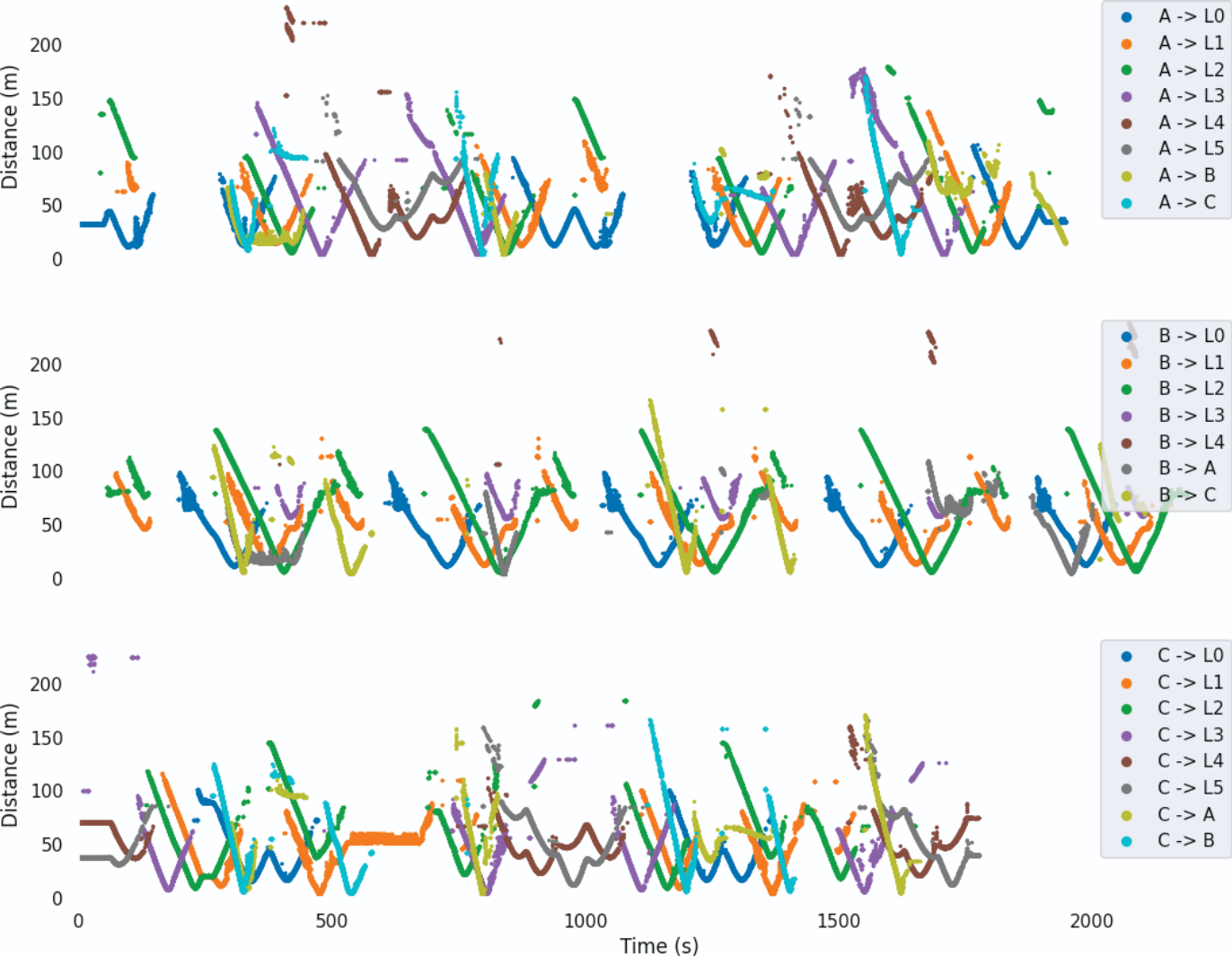}
  \caption{Raw UWB ranges over time for the three agents (A, B, C). Legends give
  the measurement association (e.g., \texttt{A->L1} is agent A to beacon 1;
  \texttt{A->B} is agent A to agent B). Note the isolated spurious returns and the
  intervals of dropout.}
  \label{fig:ranges}
\end{figure}

\textbf{Ultra-wideband ranges.} For each measurement the LinkTrack P-B reports the
range (two-way time-of-flight), the first-path received signal strength indicator
(FP RSSI), and the total received signal strength indicator (RX RSSI). Only the
range enters state estimation, but a low FP RSSI signals that the first-arriving
return was likely reflected or obstructed, which commonly accompanies a spurious
range. The collected ranges are largely self-consistent over time yet contain many
such spurious measurements from multipath and non-line-of-sight (NLOS)
propagation, along with periodic dropouts during which no ranges are available
(Fig.~\ref{fig:ranges}).

\textbf{Odometry.} We estimate per-agent odometry from the aligned RGB and depth
images using RTAB-Map~\citep{Labb__2018} with default parameters (the OpenCV~\citep{opencv_library} implementation of the
GFTT detector with ORB descriptors), and when tracking is lost the odometry is
discarded until it is reacquired. Overlaying the dead-reckoned trajectories on GPS shows that the odometry is locally accurate, in that short
segments align closely with GPS, but accumulates substantial drift over the full
traverse, confirming that the range measurements serve to provide global
correction of this drift.

\begin{figure}[t]
  \centering
  \includegraphics[width=\linewidth]{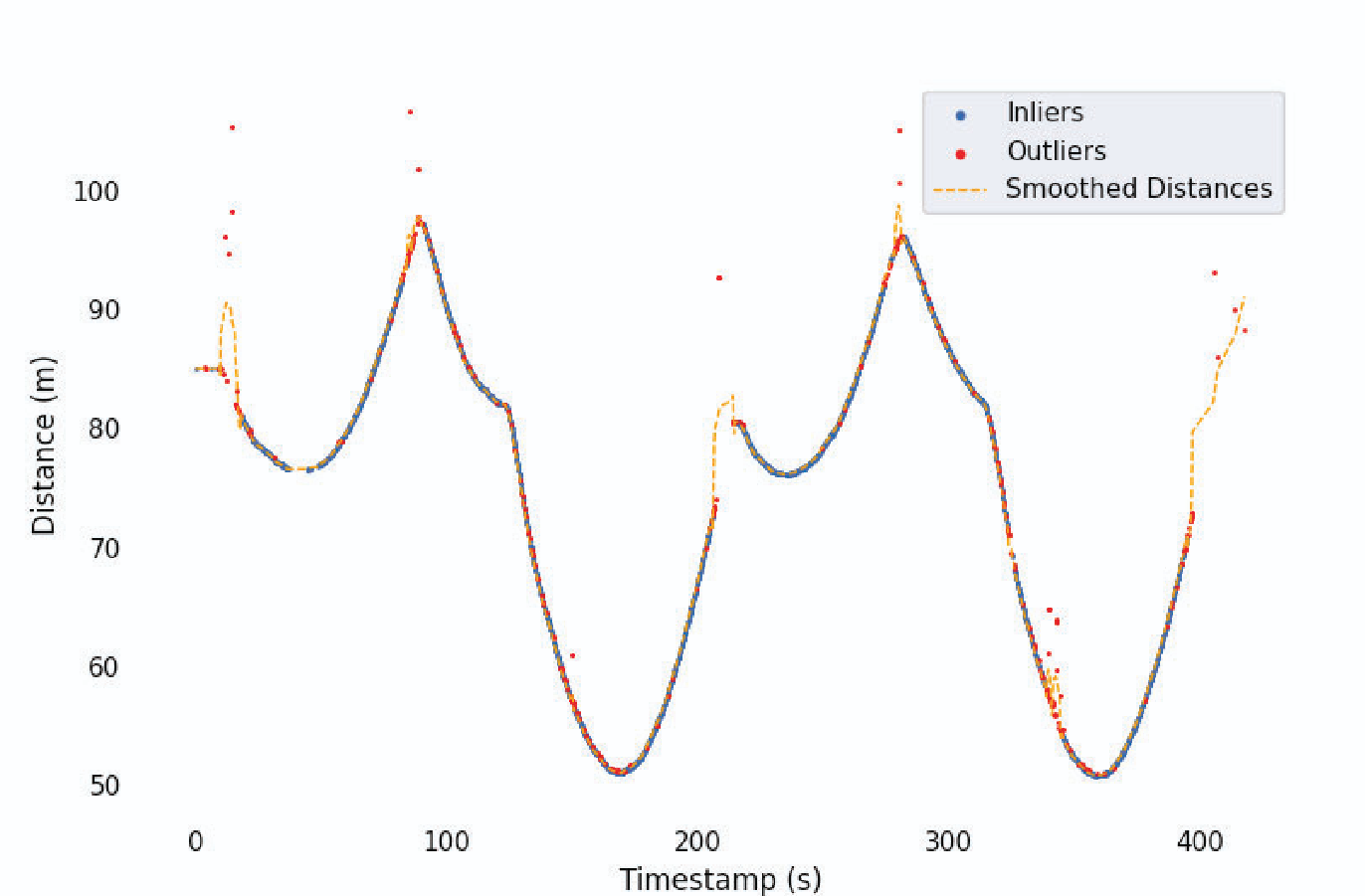}
  \caption{Temporal-filter outlier rejection on a single range stream. Inliers
  (blue) and rejected outliers (red) are shown against the Savitzky--Golay
  polynomial fit (dashed).}
  \label{fig:outlier}
\end{figure}

\textbf{Preparing the data for state estimation.} Because the raw ranges contain
apparent outliers and our formulation does not model them explicitly, we reject
outliers with a lightweight temporal filter applied independently to each range
stream. We fit a second-order polynomial to a sliding $2$\,s window using a
Savitzky--Golay filter~\citep{savitzky1964smoothing} and take each measurement's deviation from this fit as its
residual. Using the median absolute deviation (MAD) of these residuals as a robust
noise estimate, we discard any measurement whose residual exceeds five times the
MAD (Fig.~\ref{fig:outlier}). The filter is deliberately conservative, since it
removes some genuine inliers, but it reliably eliminates the apparent outliers, a
favorable trade-off given the high $50$\,Hz range rate. The released problem
instances apply this outlier rejection together with UWB range calibration and a
single start-to-end loop closure per trajectory.

\end{appendices}

%% file: bib/references.bib
@string{mit   = "The MIT Press, Cambridge, MA"}

@string{fusion = "Intl. Conf. on Information Fusion (FUSION)"}

@string{cvpr  = "Proc. {IEEE} Int. Conf. Computer Vision and Pattern Recognition"}

@string{eccv  = "Eur. Conf. on Computer Vision (ECCV)"}

@string{rss   = "Robotics: Science and Systems (RSS)"}

@string{icra  = "IEEE Intl. Conf. on Robotics and Automation (ICRA)"}

@string{ieee  = "Proc. of the IEEE"}

@misc{weber2024Power,
  title     = {Power Variable Projection for Initialization-Free Large-Scale Bundle Adjustment},
  url       = {http://arxiv.org/abs/2405.05079},
  doi       = {10.48550/arXiv.2405.05079},
  language  = {en},
  urldate   = {2025-08-21},
  publisher = {arXiv},
  author    = {Weber, Simon and Hong, Je Hyeong and Cremers, Daniel},
  month     = aug,
  year      = {2024},
  note      = {arXiv:2405.05079 [cs]}
}

@inproceedings{hong2017Revisiting,
  address   = {Honolulu, HI},
  title     = {Revisiting the Variable Projection Method for Separable Nonlinear Least Squares Problems},
  isbn      = {978-1-5386-0457-1},
  url       = {http://ieeexplore.ieee.org/document/8100112/},
  doi       = {10.1109/CVPR.2017.629},
  language  = {en},
  urldate   = {2025-08-21},
  booktitle = {2017 {IEEE} {Conference} on {Computer} {Vision} and {Pattern} {Recognition} ({CVPR})},
  publisher = {IEEE},
  author    = {Hong, Je Hyeong and Zach, Christopher and Fitzgibbon, Andrew},
  month     = jul,
  year      = {2017},
  pages     = {5939--5947}
}

@incollection{hong2016Projective,
  address   = {Cham},
  title     = {Projective Bundle Adjustment from Arbitrary Initialization Using the Variable Projection Method},
  volume    = {9905},
  isbn      = {978-3-319-46447-3 978-3-319-46448-0},
  url       = {http://link.springer.com/10.1007/978-3-319-46448-0_29},
  language  = {en},
  urldate   = {2025-08-21},
  booktitle = {Computer {Vision} – {ECCV} 2016},
  publisher = {Springer International Publishing},
  author    = {Hong, Je Hyeong and Zach, Christopher and Fitzgibbon, Andrew and Cipolla, Roberto},
  editor    = {Leibe, Bastian and Matas, Jiri and Sebe, Nicu and Welling, Max},
  year      = {2016},
  doi       = {10.1007/978-3-319-46448-0_29},
  note      = {Series Title: Lecture Notes in Computer Science},
  pages     = {477--493}
}

@article{golub2003Separable,
  title      = {Separable nonlinear least squares: the variable projection method and its applications},
  volume     = {19},
  issn       = {0266-5611, 1361-6420},
  shorttitle = {Separable nonlinear least squares},
  url        = {https://iopscience.iop.org/article/10.1088/0266-5611/19/2/201},
  doi        = {10.1088/0266-5611/19/2/201},
  language   = {en},
  number     = {2},
  urldate    = {2025-08-21},
  journal    = {Inverse Problems},
  author     = {Golub, Gene and Pereyra, Victor},
  month      = apr,
  year       = {2003},
  pages      = {R1--R26}
}

@article{matthews2004Active,
  title     = {Active Appearance Models Revisited},
  volume    = {60},
  copyright = {https://www.springernature.com/gp/researchers/text-and-data-mining},
  issn      = {0920-5691, 1573-1405},
  url       = {https://link.springer.com/10.1023/B:VISI.0000029666.37597.d3},
  doi       = {10.1023/B:VISI.0000029666.37597.d3},
  language  = {en},
  number    = {2},
  urldate   = {2025-08-22},
  journal   = {International Journal of Computer Vision},
  author    = {Matthews, Iain and Baker, Simon},
  month     = nov,
  year      = {2004},
  pages     = {135--164}
}

@inproceedings{carlone2015Initialization,
  address    = {Seattle, WA, USA},
  title      = {Initialization techniques for {3D} {SLAM}: {A} survey on rotation estimation and its use in pose graph optimization},
  isbn       = {978-1-4799-6923-4},
  shorttitle = {Initialization techniques for {3D} {SLAM}},
  url        = {http://ieeexplore.ieee.org/document/7139836/},
  doi        = {10.1109/ICRA.2015.7139836},
  language   = {en},
  urldate    = {2025-08-22},
  booktitle  = {2015 {IEEE} {International} {Conference} on {Robotics} and {Automation} ({ICRA})},
  publisher  = {IEEE},
  author     = {Carlone, Luca and Tron, Roberto and Daniilidis, Kostas and Dellaert, Frank},
  month      = may,
  year       = {2015},
  pages      = {4597--4604}
}

@article{kaess2012ISAM2,
  title      = {{iSAM2}: {Incremental} smoothing and mapping using the {Bayes} tree},
  volume     = {31},
  issn       = {0278-3649, 1741-3176},
  shorttitle = {{iSAM2}},
  url        = {https://journals.sagepub.com/doi/10.1177/0278364911430419},
  doi        = {10.1177/0278364911430419},
  language   = {en},
  number     = {2},
  urldate    = {2025-08-22},
  journal    = {The International Journal of Robotics Research},
  author     = {Kaess, Michael and Johannsson, Hordur and Roberts, Richard and Ila, Viorela and Leonard, John J and Dellaert, Frank},
  month      = feb,
  year       = {2012},
  pages      = {216--235}
}

@inproceedings{kummerle2011G2o,
  address    = {Shanghai, China},
  title      = {G$^{\textrm{2}}$o: {A} general framework for graph optimization},
  isbn       = {978-1-61284-386-5},
  shorttitle = {G$^{\textrm{2}}$o},
  url        = {http://ieeexplore.ieee.org/document/5979949/},
  doi        = {10.1109/ICRA.2011.5979949},
  language   = {en},
  urldate    = {2025-08-22},
  booktitle  = {2011 {IEEE} {International} {Conference} on {Robotics} and {Automation}},
  publisher  = {IEEE},
  author     = {Kummerle, Rainer and Grisetti, Giorgio and Strasdat, Hauke and Konolige, Kurt and Burgard, Wolfram},
  month      = may,
  year       = {2011},
  pages      = {3607--3613}
}

@inproceedings{okatani2011Efficient,
  address   = {Barcelona, Spain},
  title     = {Efficient algorithm for low-rank matrix factorization with missing components and performance comparison of latest algorithms},
  isbn      = {978-1-4577-1102-2 978-1-4577-1101-5 978-1-4577-1100-8},
  url       = {http://ieeexplore.ieee.org/document/6126324/},
  doi       = {10.1109/ICCV.2011.6126324},
  language  = {en},
  urldate   = {2025-08-22},
  booktitle = {2011 {International} {Conference} on {Computer} {Vision}},
  publisher = {IEEE},
  author    = {Okatani, Takayuki and Yoshida, Takahiro and Deguchi, Koichiro},
  month     = nov,
  year      = {2011},
  pages     = {842--849}
}

@inproceedings{woodford2020Large,
  title     = {Large Scale Photometric Bundle Adjustment},
  url       = {http://arxiv.org/abs/2008.11762},
  doi       = {10.48550/arXiv.2008.11762},
  language  = {en},
  booktitle = {2020 {British} {Machine} {Vision} {Conference}},
  urldate   = {2025-09-06},
  publisher = {arXiv},
  author    = {Woodford, Oliver J. and Rosten, Edward},
  month     = sep,
  year      = {2020},
  note      = {arXiv:2008.11762 [cs]}
}

@article{khosoussi2016Sparse,
  title = {A Sparse Separable {SLAM} Back-End},
  volume    = {32},
  copyright = {https://ieeexplore.ieee.org/Xplorehelp/downloads/license-information/IEEE.html},
  issn      = {1552-3098, 1941-0468},
  url       = {http://ieeexplore.ieee.org/document/7592861/},
  doi       = {10.1109/TRO.2016.2609394},
  language  = {en},
  number    = {6},
  urldate   = {2025-08-22},
  journal   = {IEEE Transactions on Robotics},
  author    = {Khosoussi, Kasra and Huang, Shoudong and Dissanayake, Gamini},
  month     = dec,
  year      = {2016},
  pages     = {1536--1549}
}

@incollection{agarwal2010Bundle,
  address   = {Berlin, Heidelberg},
  title     = {Bundle Adjustment in the Large},
  volume    = {6312},
  copyright = {http://www.springer.com/tdm},
  isbn      = {978-3-642-15551-2 978-3-642-15552-9},
  url       = {http://link.springer.com/10.1007/978-3-642-15552-9_3},
  language  = {en},
  urldate   = {2025-09-06},
  booktitle = {Computer {Vision} – {ECCV} 2010},
  publisher = {Springer Berlin Heidelberg},
  author    = {Agarwal, Sameer and Snavely, Noah and Seitz, Steven M. and Szeliski, Richard},
  year      = {2010},
  doi       = {10.1007/978-3-642-15552-9_3},
  note      = {Series Title: Lecture Notes in Computer Science},
  pages     = {29--42}
}

@software{agarwal2022ceres,
  author  = {Agarwal, Sameer and Mierle, Keir and The Ceres Solver Team},
  title   = {{Ceres Solver}},
  license = {Apache-2.0},
  url     = {https://github.com/ceres-solver/ceres-solver},
  version = {2.2},
  year    = {2023},
  month   = {10}
}

@article{dellaert2012factor,
  title   = {Factor graphs and {GTSAM}: A hands-on introduction},
  author  = {Dellaert, Frank},
  journal = {Georgia Institute of Technology, Tech. Rep},
  volume  = {2},
  number  = {4},
  year    = {2012}
}

@article{rosen2019SESync,
  title      = {{SE}-{Sync}: {A} certifiably correct algorithm for synchronization over the special {Euclidean} group},
  volume     = {38},
  issn       = {0278-3649, 1741-3176},
  shorttitle = {{SE}-{Sync}},
  url        = {https://journals.sagepub.com/doi/10.1177/0278364918784361},
  doi        = {10.1177/0278364918784361},
  language   = {en},
  number     = {2-3},
  urldate    = {2025-08-29},
  journal    = {The International Journal of Robotics Research},
  author     = {Rosen, David M and Carlone, Luca and Bandeira, Afonso S and Leonard, John J},
  month      = mar,
  year       = {2019},
  pages      = {95--125}
}

@article{papalia2024Certifiably,
  title     = {Certifiably Correct Range-Aided {SLAM}},
  volume    = {40},
  copyright = {https://ieeexplore.ieee.org/Xplorehelp/downloads/license-information/IEEE.html},
  issn      = {1552-3098, 1941-0468},
  url       = {https://ieeexplore.ieee.org/document/10665918/},
  doi       = {10.1109/TRO.2024.3454430},
  language  = {en},
  urldate   = {2025-08-29},
  journal   = {IEEE Transactions on Robotics},
  author    = {Papalia, Alan and Fishberg, Andrew and O'Neill, Brendan W. and How, Jonathan P. and Rosen, David M. and Leonard, John J.},
  year      = {2024},
  pages     = {4265--4283}
}

@article{halsted22arxiv,
  author  = {Halsted, Trevor and Schwager, Mac},
  title   = {The {Riemannian} Elevator for certifiable distance-based localization},
  year    = {2022},
  journal = {Preprint},
  note    = {{Accessed:} Jan. 20, 2023. [Online] Available \url{https://msl.stanford.edu/papers/halsted_riemannian_2022.pdf}}
}

@article{han2025Building,
  title     = {Building {Rome} with Convex Optimization},
  url       = {http://arxiv.org/abs/2502.04640},
  doi       = {10.48550/arXiv.2502.04640},
  urldate   = {2025-08-29},
  publisher = {arXiv},
  author    = {Han, Haoyu and Yang, Heng},
  journal   = rss,
  month     = jul,
  year      = {2025},
  note      = {arXiv:2502.04640 [cs]}
}

@article{strang2024Elimination,
  title    = {Elimination and Factorization},
  volume   = {97},
  issn     = {0025-570X, 1930-0980},
  url      = {https://www.tandfonline.com/doi/full/10.1080/0025570X.2024.2401295},
  doi      = {10.1080/0025570X.2024.2401295},
  language = {en},
  number   = {5},
  urldate  = {2025-09-02},
  journal  = {Mathematics Magazine},
  author   = {Strang, Gilbert},
  month    = oct,
  year     = {2024},
  pages    = {484--487}
}

@article{strang2022Three,
  title    = {Three matrix factorizations from the steps of elimination},
  volume   = {20},
  issn     = {0219-5305, 1793-6861},
  url      = {https://www.worldscientific.com/doi/10.1142/S0219530522400061},
  doi      = {10.1142/S0219530522400061},
  language = {en},
  number   = {06},
  urldate  = {2025-09-02},
  journal  = {Analysis and Applications},
  author   = {Strang, Gilbert and Drucker, Daniel},
  month    = nov,
  year     = {2022},
  pages    = {1147--1157}
}

@book{zhang2006schur,
  title     = {The {Schur} complement and its applications},
  author    = {Zhang, Fuzhen},
  volume    = {4},
  year      = {2006},
  publisher = {Springer Science \& Business Media}
}

@article{ruhe1980algorithms,
  title     = {Algorithms for separable nonlinear least squares problems},
  author    = {Ruhe, Axel and Wedin, Per {\AA}ke},
  journal   = {SIAM review},
  volume    = {22},
  number    = {3},
  pages     = {318--337},
  year      = {1980},
  publisher = {SIAM}
}

@book{saad2003iterative,
  title     = {Iterative methods for sparse linear systems},
  author    = {Saad, Yousef},
  year      = {2003},
  publisher = {SIAM}
}

@article{yang2020Graduated,
  title      = {Graduated Non-Convexity for Robust Spatial Perception: From Non-Minimal Solvers to Global Outlier Rejection},
  volume     = {5},
  copyright  = {https://ieeexplore.ieee.org/Xplorehelp/downloads/license-information/IEEE.html},
  issn       = {2377-3766, 2377-3774},
  shorttitle = {Graduated {Non}-{Convexity} for {Robust} {Spatial} {Perception}},
  url        = {https://ieeexplore.ieee.org/document/8957085/},
  doi        = {10.1109/LRA.2020.2965893},
  language   = {en},
  number     = {2},
  urldate    = {2025-09-05},
  journal    = {IEEE Robotics and Automation Letters},
  author     = {Yang, Heng and Antonante, Pasquale and Tzoumas, Vasileios and Carlone, Luca},
  month      = apr,
  year       = {2020},
  pages      = {1127--1134}
}

@article{tian2020asynchronous,
  title     = {Asynchronous and parallel distributed pose graph optimization},
  author    = {Tian, Yulun and Koppel, Alec and Bedi, Amrit Singh and How, Jonathan P},
  journal   = {IEEE Robotics and Automation Letters},
  volume    = {5},
  number    = {4},
  pages     = {5819--5826},
  year      = {2020},
  publisher = {IEEE}
}

@inproceedings{mcgann2024asynchronous,
  title        = {Asynchronous distributed smoothing and mapping via on-manifold consensus {ADMM}},
  author       = {McGann, Daniel and Lassak, Kyle and Kaess, Michael},
  booktitle    = {2024 IEEE International Conference on Robotics and Automation (ICRA)},
  pages        = {4577--4583},
  year         = {2024},
  organization = {IEEE}
}

@article{fan2023majorization,
  title     = {Majorization minimization methods for distributed pose graph optimization},
  author    = {Fan, Taosha and Murphey, Todd D.},
  journal   = {IEEE Transactions on Robotics},
  volume    = {40},
  pages     = {22--42},
  year      = {2024},
  publisher = {IEEE},
  doi       = {10.1109/TRO.2023.3324818}
}

@inproceedings{papalia2023score,
  author    = {Papalia, Alan and Morales, Joseph and Doherty, Kevin J. and Rosen, David M. and Leonard, John J.},
  booktitle = {2023 IEEE International Conference on Robotics and Automation (ICRA)},
  title     = {{SCORE}: A Second-Order Conic Initialization for Range-Aided {SLAM}},
  year      = {2023},
  volume    = {},
  number    = {},
  pages     = {10637-10644},
  doi       = {10.1109/ICRA48891.2023.10160787}
}

@incollection{golub1976differentiation,
  title     = {Differentiation of pseudoinverses, separable nonlinear least square problems and other tales},
  author    = {Golub, GH and Pereyra, V},
  booktitle = {Generalized inverses and applications},
  pages     = {303--324},
  year      = {1976},
  publisher = {Elsevier}
}

@incollection{triggs2000Bundle,
  address   = {Berlin, Heidelberg},
  title     = {Bundle Adjustment—A Modern Synthesis},
  volume    = {1883},
  isbn      = {978-3-540-67973-8 978-3-540-44480-0},
  url       = {https://link.springer.com/10.1007/3-540-44480-7_21},
  language  = {en},
  urldate   = {2025-09-06},
  booktitle = {Vision {Algorithms}: {Theory} and {Practice}},
  publisher = {Springer Berlin Heidelberg},
  author    = {Triggs, Bill and McLauchlan, Philip F. and Hartley, Richard I. and Fitzgibbon, Andrew W.},
  editor    = {Goos, Gerhard and Hartmanis, Juris and Van Leeuwen, Jan and Triggs, Bill and Zisserman, Andrew and Szeliski, Richard},
  year      = {2000},
  doi       = {10.1007/3-540-44480-7_21},
  note      = {Series Title: Lecture Notes in Computer Science},
  pages     = {298--372}
}

@article{dellaert2017factor,
  title     = {Factor graphs for robot perception},
  author    = {Dellaert, Frank and Kaess, Michael and others},
  journal   = {Foundations and Trends{\textregistered} in Robotics},
  volume    = {6},
  number    = {1-2},
  pages     = {1--139},
  year      = {2017},
  publisher = {Now Publishers, Inc.}
}

@article{kaess2008isam,
  title     = {{iSAM}: Incremental smoothing and mapping},
  author    = {Kaess, Michael and Ranganathan, Ananth and Dellaert, Frank},
  journal   = {IEEE Transactions on Robotics},
  volume    = {24},
  number    = {6},
  pages     = {1365--1378},
  year      = {2008},
  publisher = {IEEE}
}

@article{ebadi2023present,
  title     = {Present and future of {SLAM} in extreme environments: The {DARPA SUBT} challenge},
  author    = {Ebadi, Kamak and Bernreiter, Lukas and Biggie, Harel and Catt, Gavin and Chang, Yun and Chatterjee, Arghya and Denniston, Christopher E and Desch{\^e}nes, Simon-Pierre and Harlow, Kyle and Khattak, Shehryar and others},
  journal   = {IEEE Transactions on Robotics},
  volume    = {40},
  pages     = {936--959},
  year      = {2023},
  publisher = {IEEE}
}

@article{kunze2018artificial,
  title     = {Artificial intelligence for long-term robot autonomy: A survey},
  author    = {Kunze, Lars and Hawes, Nick and Duckett, Tom and Hanheide, Marc and Krajn{\'\i}k, Tom{\'a}{\v{s}}},
  journal   = {IEEE Robotics and Automation Letters},
  volume    = {3},
  number    = {4},
  pages     = {4023--4030},
  year      = {2018},
  publisher = {IEEE}
}

@article{tranzatto2022cerberus,
  title     = {Cerberus in the {DARPA} subterranean challenge},
  author    = {Tranzatto, Marco and Miki, Takahiro and Dharmadhikari, Mihir and Bernreiter, Lukas and Kulkarni, Mihir and Mascarich, Frank and Andersson, Olov and Khattak, Shehryar and Hutter, Marco and Siegwart, Roland and others},
  journal   = {Science Robotics},
  volume    = {7},
  number    = {66},
  pages     = {eabp9742},
  year      = {2022},
  publisher = {American Association for the Advancement of Science}
}

@article{cadena2017past,
  title     = {Past, present, and future of simultaneous localization and mapping: Toward the robust-perception age},
  author    = {Cadena, Cesar and Carlone, Luca and Carrillo, Henry and Latif, Yasir and Scaramuzza, Davide and Neira, Jos{\'e} and Reid, Ian and Leonard, John J},
  journal   = {IEEE Transactions on robotics},
  volume    = {32},
  number    = {6},
  pages     = {1309--1332},
  year      = {2017},
  publisher = {IEEE}
}

@INPROCEEDINGS{schonberger2016structure,
  author={Schönberger, Johannes L. and Frahm, Jan-Michael},
  booktitle={2016 IEEE Conference on Computer Vision and Pattern Recognition (CVPR)}, 
  title={Structure-from-Motion Revisited}, 
  year={2016},
  volume={},
  number={},
  pages={4104-4113},
  doi={10.1109/CVPR.2016.445}}

@article{mao2007wireless,
  title     = {Wireless sensor network localization techniques},
  author    = {Mao, Guoqiang and Fidan, Bar{\i}{\c{s}} and Anderson, Brian DO},
  journal   = {Computer networks},
  volume    = {51},
  number    = {10},
  pages     = {2529--2553},
  year      = {2007},
  publisher = {Elsevier}
}

@book{golub2013matrix,
  title     = {Matrix computations},
  author    = {Golub, Gene H. and Van Loan, Charles F.},
  year      = {2013},
  publisher = {JHU press}
}

@article{absil2007trust,
  title     = {Trust-region methods on {Riemannian} manifolds},
  author    = {Absil, Pierre-Antoine and Baker, Christopher G.. and Gallivan, Kyle A.},
  journal   = {Foundations of Computational Mathematics},
  volume    = {7},
  number    = {3},
  pages     = {303--330},
  year      = {2007},
  publisher = {Springer}
}

@book{chung1997spectral,
  title     = {Spectral graph theory},
  author    = {Chung, Fan RK},
  volume    = {92},
  year      = {1997},
  publisher = {American Mathematical Soc.}
}

@book{nocedal2006numerical,
  title     = {Numerical optimization},
  author    = {Nocedal, Jorge and Wright, Stephen J.},
  year      = {2006},
  publisher = {Springer}
}

@article{barham1972algorithm,
  title     = {An algorithm for least squares estimation of nonlinear parameters when some of the parameters are linear},
  author    = {Barham, Richard H. and Drane, Wanzer},
  journal   = {Technometrics},
  volume    = {14},
  number    = {3},
  pages     = {757--766},
  year      = {1972},
  publisher = {Taylor \& Francis}
}

@article{luo2010semidefinite,
  title     = {Semidefinite relaxation of quadratic optimization problems},
  author    = {Luo, Zhi-Quan and Ma, Wing-Kin and So, Anthony Man-Cho and Ye, Yinyu and Zhang, Shuzhong},
  journal   = {IEEE Signal Processing Magazine},
  volume    = {27},
  number    = {3},
  pages     = {20--34},
  year      = {2010},
  publisher = {IEEE}
}

@article{coleman1987null,
  title     = {The null space problem {II}. Algorithms},
  author    = {Coleman, Thomas F. and Pothen, Alex},
  journal   = {{SIAM} Journal on Algebraic Discrete Methods},
  volume    = {8},
  number    = {4},
  pages     = {544--563},
  year      = {1987},
  publisher = {SIAM}
}

@book{boumal2023introduction,
  title     = {An introduction to optimization on smooth manifolds},
  author    = {Boumal, Nicolas},
  year      = {2023},
  publisher = {Cambridge University Press}
}

@article{criscitiello2025sensor,
  title={Sensor network localization has a benign landscape after low-dimensional relaxation},
  author={Criscitiello, Christopher and McRae, Andrew D. and Rebjock, Quentin and Boumal, Nicolas},
  journal={arXiv preprint arXiv:2507.15662},
  year={2025}
}

@article{mcrae2024benign,
  title={Benign landscapes of low-dimensional relaxations for orthogonal synchronization on general graphs},
  author={McRae, Andrew D. and Boumal, Nicolas},
  journal={SIAM Journal on Optimization},
  volume={34},
  number={2},
  pages={1427--1454},
  year={2024},
  publisher={SIAM}
}

@misc{xu2026certifiableestimationfactorgraphs,
      title={Certifiable Estimation with Factor Graphs}, 
      author={Zhexin Xu and Nikolas R. Sanderson and Hanna Jiamei Zhang and David M. Rosen},
      year={2026},
      eprint={2603.01267},
      archivePrefix={arXiv},
      primaryClass={cs.RO},
      url={https://arxiv.org/abs/2603.01267}, 
}

@INPROCEEDINGS{sunderhauf2012Switchable,
  author={Sünderhauf, Niko and Protzel, Peter},
  booktitle={2012 IEEE/RSJ International Conference on Intelligent Robots and Systems}, 
  title={Switchable constraints for robust pose graph SLAM}, 
  year={2012},
  volume={},
  number={},
  pages={1879-1884},
  doi={10.1109/IROS.2012.6385590}}

@article{green1984iteratively,
author = {Green, P. J.},
title = {Iteratively Reweighted Least Squares for Maximum Likelihood Estimation, and Some Robust and Resistant Alternatives},
journal = {Journal of the Royal Statistical Society: Series B (Methodological)},
volume = {46},
number = {2},
pages = {149-170},
doi = {https://doi.org/10.1111/j.2517-6161.1984.tb01288.x},
url = {https://rss.onlinelibrary.wiley.com/doi/abs/10.1111/j.2517-6161.1984.tb01288.x},
eprint = {https://rss.onlinelibrary.wiley.com/doi/pdf/10.1111/j.2517-6161.1984.tb01288.x},
year = {1984}
}

@INPROCEEDINGS{jiang2024Robust,
  author={Jiang, Fan and Caruso, David and Dhekne, Ashutosh and Qu, Qi and Engel, Jakob Julian and Dong, Jing},
  booktitle={2024 IEEE International Conference on Robotics and Automation (ICRA)}, 
  title={Robust Indoor Localization with Ranging-IMU Fusion}, 
  year={2024},
  volume={},
  number={},
  pages={11963-11969},
  doi={10.1109/ICRA57147.2024.10611274}}

@incollection{sh-ch3-outlier,
  title        = {Robustness to Incorrect Data Association and Outliers},
  author       = {Heng Yang and Josh Mangelson and Yun Chang and Jingnan Shi and Luca Carlone},
  booktitle    = {{SLAM Handbook.} From Localization and Mapping to Spatial Intelligence},
  publisher    = {Cambridge University Press},
  editor       = {Luca Carlone and Ayoung Kim and Timothy Barfoot and Daniel Cremers and Frank Dellaert},
  year         = {2026}
}

@article{chen2008algorithm,
 author = {Chen, Yanqing and Davis, Timothy A and Hager, William W and Rajamanickam, Sivasankaran},
 journal = {ACM Transactions on Mathematical Software (TOMS)},
 number = {3},
 pages = {22},
 publisher = {ACM},
 title = {Algorithm 887: CHOLMOD, supernodal sparse Cholesky factorization and update/downdate},
 volume = {35},
 year = {2008}
}

@ARTICLE{holmes2023landmark,
  author={Holmes, Connor and Barfoot, Timothy D.},
  journal={IEEE Robotics and Automation Letters}, 
  title={An Efficient Global Optimality Certificate for Landmark-Based SLAM}, 
  year={2023},
  volume={8},
  number={3},
  pages={1539-1546},
  doi={10.1109/LRA.2023.3238173}}

@INPROCEEDINGS{Demmel2021square,
  author={Demmel, Nikolaus and Sommer, Christiane and Cremers, Daniel and Usenko, Vladyslav},
  booktitle={2021 IEEE/CVF Conference on Computer Vision and Pattern Recognition (CVPR)}, 
  title={Square Root Bundle Adjustment for Large-Scale Reconstruction}, 
  year={2021},
  volume={},
  number={},
  pages={11718-11727},
  doi={10.1109/CVPR46437.2021.01155}}

@INPROCEEDINGS{weber2023Power,
  author={Weber, Simon and Demmel, Nikolaus and Chan, Tin Chon and Cremers, Daniel},
  booktitle={2023 IEEE/CVF Conference on Computer Vision and Pattern Recognition (CVPR)}, 
  title={Power Bundle Adjustment for Large-Scale 3D Reconstruction}, 
  year={2023},
  volume={},
  number={},
  pages={281-289},
  doi={10.1109/CVPR52729.2023.00035}}

@article{Wise_2026,
   title={A certifiably correct algorithm for generalized robot-world and hand-eye calibration},
   ISSN={1741-3176},
   url={http://dx.doi.org/10.1177/02783649261420308},
   DOI={10.1177/02783649261420308},
   journal={The International Journal of Robotics Research},
   publisher={SAGE Publications},
   author={Wise, Emmett and Kaveti, Pushyami and Cheng, Qilong and Wang, Wenhao and Singh, Hanumant and Kelly, Jonathan and Rosen, David M. and Giamou, Matthew},
   year={2026},
   month=Mar }

@ARTICLE{olson2006Robust,
  author={Olson, Edwin and Leonard, John J. and Teller, Seth},
  journal={IEEE Journal of Oceanic Engineering}, 
  title={Robust Range-Only Beacon Localization}, 
  year={2006},
  volume={31},
  number={4},
  pages={949-958},
  doi={10.1109/JOE.2006.880386}}

@book{barfoot2017state,
author = {Barfoot, Timothy D.},
title = {State Estimation for Robotics},
year = {2017},
isbn = {1107159393},
publisher = {Cambridge University Press},
address = {USA},
edition = {1st}
}

@article{hartley2013rotation,
  author  = {Hartley, Richard and Trumpf, Jochen and Dai, Yuchao and Li, Hongdong},
  title   = {Rotation Averaging},
  journal = {International Journal of Computer Vision},
  volume  = {103},
  number  = {3},
  pages   = {267--305},
  year    = {2013},
  doi     = {10.1007/s11263-012-0601-0}
}

@misc{mcgann2025cosmobenchbenchmarkcollaborativeslam,
      title={COSMO-Bench: A Benchmark for Collaborative SLAM Optimization}, 
      author={Daniel McGann and Easton R. Potokar and Michael Kaess},
      year={2025},
      eprint={2508.16731},
      archivePrefix={arXiv},
      primaryClass={cs.RO},
      url={https://arxiv.org/abs/2508.16731}, 
}

@ARTICLE{chang2022lamp,
  author={Chang, Yun and Ebadi, Kamak and Denniston, Christopher E. and Ginting, Muhammad Fadhil and Rosinol, Antoni and Reinke, Andrzej and Palieri, Matteo and Shi, Jingnan and Chatterjee, Arghya and Morrell, Benjamin and Agha-mohammadi, Ali-akbar and Carlone, Luca},
  journal={IEEE Robotics and Automation Letters}, 
  title={LAMP 2.0: A Robust Multi-Robot SLAM System for Operation in Challenging Large-Scale Underground Environments}, 
  year={2022},
  volume={7},
  number={4},
  pages={9175-9182},
  doi={10.1109/LRA.2022.3191204}}

@book{koller2009probabilistic,
author = {Koller, Daphne and Friedman, Nir},
title = {Probabilistic Graphical Models: Principles and Techniques - Adaptive Computation and Machine Learning},
year = {2009},
isbn = {0262013193},
publisher = {The MIT Press}
}

@ARTICLE{Umeyama1991least,
  author={Umeyama, S.},
  journal={IEEE Transactions on Pattern Analysis and Machine Intelligence}, 
  title={Least-squares estimation of transformation parameters between two point patterns}, 
  year={1991},
  volume={13},
  number={4},
  pages={376-380},
  doi={10.1109/34.88573}}

@misc{xu2026implementingrobustmestimatorscertifiable,
      title={Implementing Robust M-Estimators with Certifiable Factor Graph Optimization}, 
      author={Zhexin Xu and Hanna Jiamei Zhang and Helena Calatrava and Pau Closas and David M. Rosen},
      year={2026},
      eprint={2603.20932},
      archivePrefix={arXiv},
      primaryClass={cs.RO},
      url={https://arxiv.org/abs/2603.20932}, 
}

@INPROCEEDINGS{papalia2025sparsevariableprojectionrobotic,
  author={Papalia, Alan and Sanderson, Nikolas and Han, Haoyu and Yang, Heng and Singh, Hanumant and Everett, Michael},
  booktitle={2026 IEEE International Conference on Robotics and Automation (ICRA)}, 
  title={Sparse Variable Projection in Robotic Perception: Exploiting Separable Structure for Efficient Nonlinear Optimization}, 
  year={2026},
  volume={},
  number={},
  pages={11826-11833},
  doi={10.1109/ICRA57385.2026.11697424}}

@inproceedings{Martiros-RSS-22,
    author    = {Hayk Martiros AND Aaron Miller AND Nathan Bucki AND Bradley Solliday AND Ryan Kennedy AND Jack Zhu AND Tung Dang AND Dominic Pattison AND Harrison Zheng AND Teo Tomic AND Peter Henry AND Gareth Cross AND Josiah VanderMey AND Alvin Sun AND Samuel Wang AND Kristen Holtz},
    title     = {{SymForce: Symbolic Computation and Code Generation for Robotics}},
    booktitle = {Proceedings of Robotics: Science and Systems},
    year      = {2022},
    doi       = {10.15607/RSS.2022.XVIII.041}
}

@misc{martens2026casparcudaacceleratorsymbolic,
      title={Caspar: CUDA Accelerator for Symbolic Programming with Adaptive Reordering}, 
      author={Emil Martens and Aaron Miller and Matias Varnum and Annette Stahl},
      year={2026},
      eprint={2605.30583},
      archivePrefix={arXiv},
      primaryClass={cs.RO},
      url={https://arxiv.org/abs/2605.30583}, 
}

@inproceedings{dellaert2020shonan,
 title = {Shonan Rotation Averaging: Global Optimality by Surfing ${SO}(p)^n$},
author = {Dellaert, Frank and Rosen, David M. and Wu, Jing and Mahony, Robert and Carlone, Luca},
booktitle = {European Conference on Computer Vision ({ECCV})},
month = aug,
year = {2020},
}

@inproceedings{mcgann_imesa_2024, 
    title = {{iMESA}: Incremental Distributed Optimization for Collaborative Simultaneous Localization and Mapping},
    author = {D. McGann and M. Kaess},
    fullauthor = {Daniel McGann and Michael Kaess},
    booktitle = {Proc. Robotics: Science and Systems (RSS)},
    year = 2024,
    address = {Delft, {NL}}
}

@ARTICLE{mcgann2026rimesa,
  author={McGann, Daniel and Kaess, Michael},
  journal={IEEE Transactions on Robotics}, 
  title={riMESA: Consensus ADMM for Real-World Collaborative SLAM}, 
  year={2026},
  volume={},
  number={},
  pages={1-20},
  doi={10.1109/TRO.2026.3706569}}

@INPROCEEDINGS{Mcgann2023riSAM,
  author={McGann, Daniel and Rogers, John G. and Kaess, Michael},
  booktitle={2023 IEEE International Conference on Robotics and Automation (ICRA)}, 
  title={Robust Incremental Smoothing and Mapping (riSAM)}, 
  year={2023},
  volume={},
  number={},
  pages={4157-4163},
  doi={10.1109/ICRA48891.2023.10161438}}

@ARTICLE{holmes2026fastsync,
  author={Holmes, Shane and Luo, Yiran and Taxpulat, Firat and Rosen, David M. and Dellaert, Frank},
  journal={IEEE Robotics and Automation Letters}, 
  title={FAST-Sync: Fast Group Synchronization for Any Matrix Lie Group}, 
  year={2026},
  volume={11},
  number={9},
  pages={10377-10384},
  doi={10.1109/LRA.2026.3710327}}

@misc{dumbgen2025exploiting,
  title         = {Exploiting Chordal Sparsity for Fast Global Optimality with Application to Localization},
  author        = {D{\"u}mbgen, Frederike and Holmes, Connor and Barfoot, Timothy D.},
  year          = {2025},
  eprint        = {2406.02365},
  archivePrefix = {arXiv},
  primaryClass  = {cs.RO},
  url           = {https://arxiv.org/abs/2406.02365}
}

@misc{subramanian2026exploitingchordalsparsityglobally,
      title={Exploiting Chordal Sparsity for Globally Optimal Estimation with Factor Graphs}, 
      author={Avinash Subramanian and Connor Holmes and Timothy D. Barfoot and Frank Dellaert and Frederike Dümbgen},
      year={2026},
      eprint={2605.30617},
      archivePrefix={arXiv},
      primaryClass={cs.RO},
      url={https://arxiv.org/abs/2605.30617}, 
}

@article{fan2023daba,
    title={Decentralization and Acceleration Enables Large-Scale Bundle Adjustment}, 
    author={Fan, Taosha and Ortiz, Joseph and Hsiao, Ming and Monge, Maurizio and Dong, Jing and Murphey, Todd and  Mukadam, Mustafa},
    journal={arXiv:2305.07026},
    year={2023},
}

@article{leung2011utias,
  author    = {Leung, Keith Y. K. and Halpern, Yoni and Barfoot, Timothy D. and Liu, Hugh H. T.},
  title     = {The {UTIAS} Multi-Robot Cooperative Localization and Mapping Dataset},
  journal   = {The International Journal of Robotics Research},
  volume    = {30},
  number    = {8},
  pages     = {969--974},
  year      = {2011},
  month     = jul,
  publisher = {SAGE Publications},
  doi       = {10.1177/0278364911398404}
}

@misc{yu2023fusingodometryuwbranging,
      title={Fusing Odometry, UWB Ranging, and Spatial Detections for Relative Multi-Robot Localization}, 
      author={Xianjia Yu and Iacopo Catalano and Paola Torrico Morón and Sahar Salimpour and Tomi Westerlund and Jorge Peña Queralta},
      year={2023},
      eprint={2304.06264},
      archivePrefix={arXiv},
      primaryClass={cs.RO},
      url={https://arxiv.org/abs/2304.06264}, 
}

@article{djugash2009plaza,
author = {Djugash, Joseph and Hamner, Bradley and Roth, Stephan},
title = {Navigating with ranging radios: Five data sets with ground truth},
journal = {Journal of Field Robotics},
volume = {26},
number = {9},
pages = {689-695},
doi = {https://doi.org/10.1002/rob.20311},
url = {https://onlinelibrary.wiley.com/doi/abs/10.1002/rob.20311},
eprint = {https://onlinelibrary.wiley.com/doi/pdf/10.1002/rob.20311},
year = {2009}
}

@misc{gopinath2026graphitegpuacceleratedmixedprecisiongraph,
      title={Graphite: A GPU-Accelerated Mixed-Precision Graph Optimization Framework}, 
      author={Shishir Gopinath and Karthik Dantu and Steven Y. Ko},
      year={2026},
      eprint={2509.26581},
      archivePrefix={arXiv},
      primaryClass={cs.RO},
      url={https://arxiv.org/abs/2509.26581}, 
}

@inproceedings{ren2022megba,
author = {Ren, Jie and Liang, Wenteng and Yan, Ran and Mai, Luo and Liu, Shiwen and Liu, Xiao},
title = {MegBA: A GPU-Based Distributed Library for Large-Scale Bundle Adjustment},
year = {2022},
isbn = {978-3-031-19835-9},
publisher = {Springer-Verlag},
address = {Berlin, Heidelberg},
url = {https://doi.org/10.1007/978-3-031-19836-6_40},
doi = {10.1007/978-3-031-19836-6_40},
booktitle = {Computer Vision – ECCV 2022: 17th European Conference, Tel Aviv, Israel, October 23–27, 2022, Proceedings, Part XXXVII},
pages = {715–731},
numpages = {17},
location = {Tel Aviv, Israel}
}

@misc{sonawalla2026overlappingdomaindecompositiondistributed,
      title={Overlapping Domain Decomposition for Distributed Pose Graph Optimization}, 
      author={Aneesa Sonawalla and Yulun Tian and Jonathan P. How},
      year={2026},
      eprint={2603.03499},
      archivePrefix={arXiv},
      primaryClass={cs.RO},
      url={https://arxiv.org/abs/2603.03499}, 
}

@article{Labb__2018,
   title={RTAB‐Map as an open‐source lidar and visual simultaneous localization and mapping library for large‐scale and long‐term online operation},
   volume={36},
   ISSN={1556-4967},
   url={http://dx.doi.org/10.1002/rob.21831},
   DOI={10.1002/rob.21831},
   number={2},
   journal={Journal of Field Robotics},
   publisher={Wiley},
   author={Labbé, Mathieu and Michaud, François},
   year={2018},
   month=Oct, pages={416–446} }

@article{opencv_library,
    author = {Bradski, G.},
    journal = {Dr. Dobb's Journal of Software Tools},
    title = {{The OpenCV Library}},
    year = {2000}
}

@article{savitzky1964smoothing,
  author  = {Savitzky, Abraham and Golay, Marcel J. E.},
  title   = {Smoothing and differentiation of data by simplified least squares procedures},
  journal = {Analytical Chemistry},
  volume  = {36},
  number  = {8},
  pages   = {1627--1639},
  year    = {1964}
}

@article{cis/1241018500,
author = {Alessandro Chiuso and Giorgio Picci and Stefano Soatto},
title = {{Wide-Sense Estimation on the Special Orthogonal Group}},
volume = {8},
journal = {Communications in Information\& Systems},
number = {3},
publisher = {International Press of Boston},
pages = {185 -- 200},
year = {2008},
}

@ARTICLE{930934,
  author={Kanatani, K. and Morris, D.D.},
  journal={IEEE Transactions on Information Theory}, 
  title={Gauges and gauge transformations for uncertainty description of geometric structure with indeterminacy}, 
  year={2001},
  volume={47},
  number={5},
  pages={2017-2028},
  doi={10.1109/18.930934}}

@ARTICLE{8354808,
  author={Zhang, Zichao and Gallego, Guillermo and Scaramuzza, Davide},
  journal={IEEE Robotics and Automation Letters}, 
  title={On the Comparison of Gauge Freedom Handling in Optimization-Based Visual-Inertial State Estimation}, 
  year={2018},
  volume={3},
  number={3},
  pages={2710-2717},
  doi={10.1109/LRA.2018.2833152}}

@ARTICLE{6898876,
  author={Carlevaris-Bianco, Nicholas and Kaess, Michael and Eustice, Ryan M.},
  journal={IEEE Transactions on Robotics}, 
  title={Generic Node Removal for Factor-Graph SLAM}, 
  year={2014},
  volume={30},
  number={6},
  pages={1371-1385},
  doi={10.1109/TRO.2014.2347571}}

@article{black1996unification,
  author  = {Black, Michael J. and Rangarajan, Anand},
  title   = {On the unification of line processes, outlier rejection, and
             robust statistics with applications in early vision},
  journal = {International Journal of Computer Vision},
  volume  = {19}, number = {1}, pages = {57--91}, year = {1996}
}

@book{doi:10.1137/1.9780898718881,
author = {Davis, Timothy A.},
title = {Direct Methods for Sparse Linear Systems},
publisher = {Society for Industrial and Applied Mathematics},
year = {2006},
doi = {10.1137/1.9780898718881},
address = {},
edition   = {},
URL = {https://epubs.siam.org/doi/abs/10.1137/1.9780898718881},
eprint = {https://epubs.siam.org/doi/pdf/10.1137/1.9780898718881}
}

@ARTICLE{goudar2024optimal,
  author={Goudar, Abhishek and Dümbgen, Frederike and Barfoot, Timothy D. and Schoellig, Angela P.},
  journal={IEEE Robotics and Automation Letters}, 
  title={Optimal Initialization Strategies for Range-Only Trajectory Estimation}, 
  year={2024},
  volume={9},
  number={3},
  pages={2160-2167},
  doi={10.1109/LRA.2024.3354623}}
